\documentclass{article}

\usepackage{microtype}
\usepackage{graphicx}
\usepackage{subcaption}
\usepackage{placeins}
\usepackage{stfloats}
\usepackage{multicol}
\usepackage{balance}
\usepackage{booktabs} %

\usepackage{hyperref}

\usepackage[preprint]{icml2026}

\usepackage{amsmath}
\usepackage{amssymb}
\usepackage{mathtools}
\usepackage{amsthm}

\usepackage{url}
\usepackage[utf8]{inputenc} %
\usepackage[T1]{fontenc}    %
\usepackage{booktabs}
\usepackage{url}            %
\usepackage{amsfonts}       %
\usepackage{nicefrac}       %
\usepackage{microtype}      %
\usepackage{natbib}
\usepackage[export]{adjustbox}
\usepackage{graphicx} %
\usepackage{xurl}
\usepackage{enumitem}
\usepackage{makecell}
\usepackage{xspace}
\usepackage{graphbox}
\usepackage{enumitem}
\usepackage{colortbl}
\usepackage{multirow}
\usepackage{color}
\usepackage[capitalize,noabbrev]{cleveref}
\usepackage[dvipsnames]{xcolor}

\usepackage{booktabs}
\usepackage{multirow}
\usepackage{algorithm}
\usepackage{listings}
\usepackage[most]{tcolorbox}

\theoremstyle{plain}

\theoremstyle{definition}

\theoremstyle{remark}

\usepackage[textsize=tiny]{todonotes}

\newcommand{\treasure}{treasure\xspace}

\newcolumntype{C}[1]{>{\centering\arraybackslash}m{#1}}
\newcolumntype{L}[1]{>{\raggedright\arraybackslash}m{#1}}
\newcolumntype{R}[1]{>{\raggedleft\arraybackslash}m{#1}}
\newlength\newl
\newlength\newlc

\newlength\colwidth
\newlength\figwidth

\newlength\myindent
\newcommand{\frozenlake}{\textsc{FrozenLake}\xspace}
\newcommand{\dataalchemy}{\textsc{DataAlchemy}\xspace}

\definecolor{brightpurple}{RGB}{160,0,255}

\definecolor{lightorange}{RGB}{255, 229, 180}
\definecolor{lightblue}{RGB}{155, 155, 255}
\definecolor{lightgreen}{RGB}{182, 235, 185}

\icmltitlerunning{
In- and Out-of-Distribution Generalization of Reasoning in Multimodal LLMs for Simple Visual Planning Tasks
}

\begin{document}

\twocolumn[
  \icmltitle{%
    In- and Out-of-Distribution Generalization of Reasoning in Multimodal LLMs for Simple Visual Planning Tasks
    }

  \icmlsetsymbol{equal}{*}

  \begin{icmlauthorlist}
    \icmlauthor{Yannic Neuhaus}{tuebingen}
    \icmlauthor{Chanoknan Tanchotikul}{aalto}
    \icmlauthor{Nicolas Flammarion}{epfl}
    \icmlauthor{Matthias Hein}{tuebingen}
    \icmlauthor{Francesco Croce}{ellis,aalto}
  \end{icmlauthorlist}

  \icmlaffiliation{tuebingen}{Tübingen AI Center -- University of Tübingen}
  \icmlaffiliation{epfl}{EPFL}
  \icmlaffiliation{ellis}{ELLIS Institute Finland}
  \icmlaffiliation{aalto}{Aalto University}
  
  \icmlcorrespondingauthor{Yannic Neuhaus}{yannic.neuhaus@uni-tuebingen.de}

  \icmlkeywords{Machine Learning, ICML}

  \vskip 0.3in
]

\makeatletter
\renewcommand{\Notice@String}{\textit{Published as a paper at the
  $\mathit{1}^{st}$ FoGen workshop, ICML 2026},
Seoul, South Korea, 2026.
Copyright 2026 by the author(s).}
\makeatother
\printAffiliationsAndNotice{}

\begin{abstract}
Integrating reasoning in (multimodal) large language models has recently led to significant improvement of their capabilities.
However, generalization in reasoning models is still vaguely defined and poorly understood.
In this work, we present an evaluation framework to rigorously examine how well chain-of-thought (CoT) approaches generalize in simple visual planning, specifically on a grid-based navigation task.
The versatility of the task and its data allows us to fine-tune model variants using different input representations (visual and textual) and CoT reasoning strategies, and systematically evaluate them under both in-distribution (ID) and out-of-distribution (OOD) test conditions.
Our experiments show that the out-of-distribution generalization (e.g., to larger maps) is largely impacted by the format used for input maps and CoT chains.
Surprisingly, we find that coordinate-based reasoning traces which combine text descriptions yield the best OOD generalization.
Moreover, CoT reproducing the steps of the A* algorithm yields the state-of-the-art ID accuracy, and simple augmentation of the map solutions seen during training greatly boosts OOD results.
Repeating the experiments with maps represented by coordinates yields similar results for another model family. Tests on structurally complex maps reveal a sharp performance drop when shortest solutions require many movement directions or turns. Additional fine-tuning on examples of directionally complex maps substantially recovers this performance.
Finally, purely text-based models consistently outperform those utilizing image-based inputs, including a recently proposed approach relying on latent space reasoning.
\end{abstract}

\section{Introduction}
\label{sec:intro}

\begin{figure*}
    \centering
    \small
    \includegraphics[width=.95\textwidth]{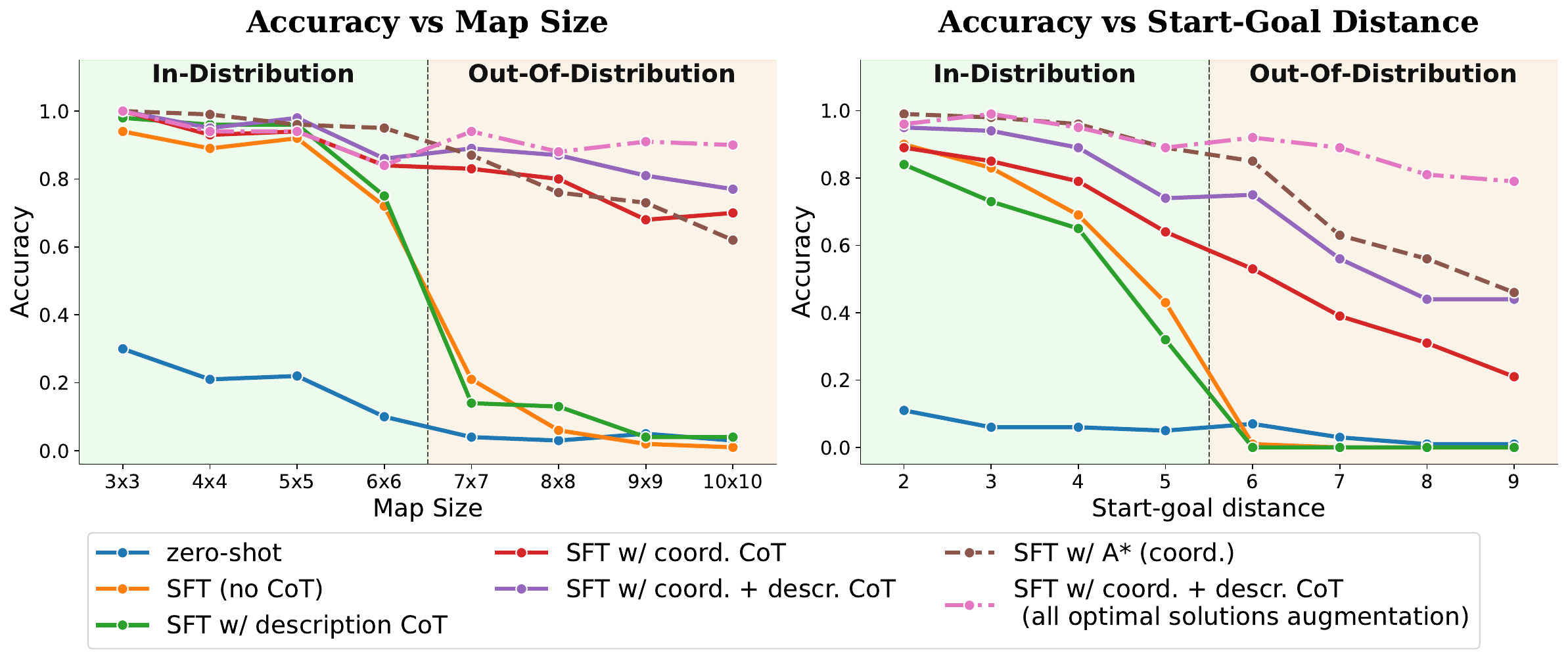}
    \caption{\textbf{Chain-of-thought (CoT) format and augmentations impact in- (ID) and out-of-distribution (OOD) generalization.}
    We fine-tune Qwen2.5-VL-7B-Instruct on \frozenlake with maps in \textit{coordinates} format.
    Supervised fine-tuning (SFT) without CoT or with CoT in \textit{description} format lead to no OOD generalization to larger map sizes or start-goal distances.
    Conversely, combining \textit{coordinates} and \textit{description} reasoning steps yields good accuracy under severe shifts. 
    CoT reproducing the steps of the A* algorithm achieves overall the best ID and competitive OOD results.
    This shows that generalization of reasoning is influenced by the format of the CoT traces.
    Finally, augmenting the training data with multiple solutions for each map further boosts OOD accuracy. 
    }
    \label{fig:teaser}
\end{figure*}

Chain-of-thought (CoT) reasoning has been shown to be a powerful tool to improve the ability of large language models (LLMs) to solve complex tasks \cite{wei2022chain}.
In practice, both test-time prompting (\textit{``Let's think step-by-step''}) \cite{kojima2022large} or specifically designed fine-tuning schemes \cite{ho2023large} can make LLMs generate intermediate reasoning steps which aid arriving to the correct solution. 
This approach may also benefit interpretability, as the user can inspect the reasoning process behind the response and potential failure points.
Recently, the chain-of-thought paradigm has been extended to multimodal models to enhance their capabilities to reason about visual inputs \cite{huang2025vision, yang2025machine}.
Despite its widespread success and adoption, we still lack a fundamental understanding of why chain-of-thought is effective and what kinds of reasoning capabilities LLMs actually acquire. Recent work further suggests that the apparent reasoning abilities of multimodal models may largely reflect statistical regularities in the training data rather than genuine algorithmic learning, as evidenced by performance degradation when inputs diverge from the training distribution \cite{stechly2024chain, zhao2025chain}.

\begin{figure*}[t]
    \centering
    \resizebox{.9\textwidth}{!}{\begingroup
\newcommand{\RepFont}{\fontfamily{pcr}\fontseries{m}\selectfont\tiny}
\newcommand{\RepHeight}{3.20cm}

\newcommand{\RepBox}[2]{%
\begin{tcolorbox}[
    enhanced,
    colback=gray!12,
    colframe=gray!12,
    boxrule=0pt,
    arc=4pt,
    boxsep=0pt,
    left=4pt,
    right=4pt,
    top=3pt,
    bottom=3pt,
    width=#1,
    height=\RepHeight,
    valign=center,
]
{\RepFont #2}
\end{tcolorbox}%
}

\resizebox{\linewidth}{!}{%
\begin{tabular}{@{}c@{\hspace{4pt}}c@{\hspace{4pt}}c@{\hspace{4pt}}c@{\hspace{4pt}}c@{}}
\multicolumn{1}{c}{\normalsize Image} &
\multicolumn{1}{c}{\normalsize Description} &
\multicolumn{1}{c}{\normalsize Grid} &
\multicolumn{1}{c}{\normalsize Coordinates}
\\[-1pt]

\begin{tcolorbox}[
    enhanced,
    colback=white,
    colframe=white,
    boxrule=0pt,
    arc=0pt,
    boxsep=0pt,
    left=0pt,
    right=0pt,
    top=0pt,
    bottom=0pt,
    width=3.25cm,
    height=\RepHeight,
    valign=center,
]
\centering
\includegraphics[height=\RepHeight,keepaspectratio]{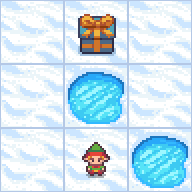}
\end{tcolorbox}
&
\RepBox{5.00cm}{%
This is a 3x3 map.\\
The player is at: row 3, column 2;\\
The hole(s) are at: Row 2, Column 2;\\
Row 3, Column 3;\\
The goal is at: Row 1, Column 2.%
}
&
\RepBox{0.95cm}{%
Grid:\\
.~G~.\\
.~X~.\\
.~P~X%
}
&

\RepBox{5.75cm}{%
Coordinates are zero-indexed as (row, col),\\
with (0,0) at the top-left corner.\\
The map is a 3 by 3 grid.\\
The player starts at (2,1) and the goal is at (0,1).\\
Holes are located at \{(1,1),(2,2)\}.\\
Every coordinate not listed as a hole is a safe cell.%
}
\end{tabular}%
}
\endgroup
}
    \caption{\textbf{Maze representations.} 
    %
    Besides the \textit{image} and \textit{description} formats used by \citet{wu2024vsp}, we introduce \textit{grid} and \textit{coordinates} as text-based map representations.
    }
    \label{fig:maze-repr}
\end{figure*}

In this work, we want to systematically study both in-distribution (ID) and, especially, out-of-distribution (OOD) generalization behavior of LLMs in a controlled setup.
Such a comprehensive evaluation is not straightforward with common benchmarks \citep{lu2022learn, yue2024mmmu}, as there is no clear distinction between ID and OOD tasks, or an evident ground-truth algorithm to be learned. 
Therefore, we first build a rich evaluation environment upon the \frozenlake dataset \cite{wu2024vsp}, which requires visual planning to find the correct path to reach a treasure without falling into lakes (see 
Fig.~\ref{fig:maze-repr}).
The fact that even small maps are challenging for state-of-the-art LLMs \cite{wu2024vsp} makes it a relevant testbed to study generalization of reasoning models.
Moreover, 
both maps and reasoning steps can be represented as images or in text-only formats (Fig.~\ref{fig:maze-repr}).
Crucially, task complexity can be easily controlled through factors such as map size, start-goal distance, and the number of lakes, allowing us to distinguish between in- and out-of-distribution data, and to create multiple distribution shifts for evaluating generalization to unseen data types.
We further control the minimum number of movement directions and turns among the shortest solutions.

We train both models without CoT supervision and reasoning models, using the same training maps but encoding them in different formats.
We consider four main input formats: \textit{image}, \textit{description}, \textit{grid}, and \textit{coordinates} (the latter three being text-based).
Reasoning traces are expressed either in one of these text-based formats or in a combination of them (Fig.~\ref{fig:reasoning-traces}).
Moreover, the \textit{coordinates} format naturally supports A*-like reasoning traces, which provide an algorithmic strategy guaranteed to find a correct path.
As most maps admit multiple valid paths, our framework also allows us to vary the number of solutions (and reasoning traces) per map while keeping the underlying training maps fixed.
This lets us study solution-set augmentation as a controlled complement to input and CoT format.
Across all models, performance drops on OOD data, especially when the start and goal positions are farther apart than in the training data. 
Surprisingly, however, we find that the choice of input and CoT formats strongly influences how sharply accuracy degrades. 
As shown in Fig.~\ref{fig:teaser}, reasoning traces based on \textit{coordinates + description} and A*-like formats yield substantially better OOD performance than no CoT, \textit{description}-only CoT, or even \textit{coordinates}-only CoT on two challenging distribution shifts.
Even on ID test maps, these models achieve state-of-the-art results, outperforming recent methods based on continuous-space reasoning \citep{yang2025machine} and specialized vision-only models \citep{xu2025visual}, with our A*-like CoT model reaching 98\% accuracy compared to 92\% for prior methods (Table~\ref{tab:other_baselines_main}).
Repeating the experiments with coordinate-based inputs gives similar results for models from another family.
%
Finally, we show that data augmentation is crucial for the strongest OOD results: augmenting the training maps with multiple solutions further improves OOD accuracy, reaching 85\% on the most challenging fixed-size distance shift and 91\% on larger maps with $d_\infty \geq 6$ (Fig.~\ref{fig:teaser}).
Tests on structurally complex maps reveal a remaining limitation: accuracy drops sharply when shortest solutions require many movement directions or turns.
Continued fine-tuning on directionally complex maps substantially improves these results.

Overall, our study shows that both ID and OOD generalization in visual planning can be largely improved through appropriate multimodal data formats and simple augmentations.

\section{Related Work}
\label{sec:related}

Reasoning LLMs achieve notable results on a variety of tasks \citep{chen2025towards}, and understanding their functioning and limitations is an active area of research \citep{shojaee2025illusion, mirzadeh2025gsmsymbolic, wang2025theoretical}. 
In particular, \citet{stechly2024chain} study the performance of state-of-the-art LLMs in text-based planning domain and simple synthetic tasks.
In their case, CoT demonstrations are added in context rather than used for fine-tuning.
Across tasks, they observe limited improvements due to CoT prompts, restricted to when these are very similar to the target test examples.
Therefore, \citet{stechly2024chain} argue that the model does not learn general algorithmic procedures via such demonstrations.
\citet{zhao2025chain} introduce \dataalchemy, a controllable environment which abstracts language tasks with simple symbolic inputs and transformations.
In this way, they can control the OOD generalization along different distribution shifts, of both transformers trained from scratch and fine-tuned LLMs.
While CoT reasoning is effective on in-distribution data, it quickly degrades under moderate shifts, again suggesting that the models learn to recognize patterns from the training set rather than logical structures.
In our work, we expand this line of research of studying OOD generalization of CoT reasoning: our tasks are trivial for humans but challenging for SOTA LLMs, suitable for both training from scratch and fine-tuning, and can be extended to arbitrarily difficult levels.
Unlike previous benchmarks such as \dataalchemy, our dataset lets us study the interplay between input and CoT format, spanning both text and image representations, which has recently drawn attention in the context of multimodal reasoning models \citep{zhou2025visualizing}.
This reveals that in some cases significant OOD generalization can be achieved.

\section{A Controlled Environment to Benchmark Generalization of Reasoning}
\label{sec:method}

\subsection{
Input and chain-of-thought format}

\textbf{Task.}
We consider the spatial planning task introduced in \citet{wu2024vsp}, named \frozenlake.
In this task, the model has to navigate a player through a fully observable maze. The input consists of a textual description of the rules and goal of the game, and a grid-based map of the maze, where one cell hosts the player, another the \treasure, and the remaining cells are either empty or contain lakes.
The model's output is a sequence of moves (``UP'', ``DOWN'', ``LEFT'', ``RIGHT'') that should guide the player to the \treasure without falling into a lake. 
This setting offers several advantages for our analysis: the input map can be easily represented as either an image or text, and the solutions 
have a clear visual and descriptive structure.
Moreover, the difficulty of the task can be adjusted by, e.g., the size of the map or the distance between the start and the \treasure: these controllable difficulty factors enable us to systematically examine out-of-distribution generalization of the models.
Finally, despite its simplicity for humans, even powerful LLMs struggle to solve the task without fine-tuning \cite{wu2024vsp}, e.g. 46\% accuracy for GPT-4o \cite{hurst2024gpt} or 29\% for Gemini \cite{team2023gemini} (on average on small 3×3, 4×4, 5×5, and 6×6 maps), see Table~\ref{tab:other_baselines}. %


\textbf{Maze representations.}
In \citet{wu2024vsp}, the planning task is introduced with several representations of the maze: as an \textit{image}, an unstructured text \textit{description} and as an ASCII \textit{table} in Markdown-like format.
We additionally introduce \textit{(i)} an ASCII-based \textit{grid} representation, which has a compact table-like structure, which can be considered a text-version of the image grid 
and \textit{(ii)} a numerical \textit{coordinates} representation that specifies the location of player, goal and lakes on the 2D grid but does not have any explicit visual structure (we further test a variant with alphanumeric coordinates in App.~\ref{sec:additional_formats}).
We use the four representations illustrated in Fig.~\ref{fig:maze-repr} in the main part, and report results for additional formats in appendix.
This setup allows us to use the same data both with text-only and multimodal LLMs.

\begin{figure*}[t]
    \centering
    \resizebox{.9\textwidth}{!}{
    \input{tables/reasoning_traces_all_steps_table.tex}}
    \caption{\textbf{Reasoning traces in different formats.} We illustrate an example of the first two steps of reasoning traces in the various formats. %
    While \citet{xu2025visual} use the sequence of maze representations as \textit{images}, we generate %
    the corresponding steps as text-only \textit{descriptions}, \textit{grids}, \textit{coordinates} and A* (with \textit{coordinates}).
    In the \textit{description} format at each step we formulate a discussion of which the next step should be, while for the \textit{grid} and \textit{coordinates} formats we have the visual representation of the map after the next move.
    The A* trace explicitly tracks the open and closed sets, parent pointers, and the skipped invalid moves at each step.
    }
    \label{fig:reasoning-traces}
    \vspace{0mm}
\end{figure*}


\textbf{Reasoning traces representations.}
The simplicity of the task makes the reasoning process toward a correct solution sequence straightforward, since each step corresponds to a step along a correct path to the \treasure (more details on the construction of reasoning traces in App.~\ref{sec:exp_details}).
This, together with the versatility of the data, enables generating multiple variants of chain-of-thought reasoning traces with different formats (see Fig.~\ref{fig:reasoning-traces}), whose effect on ID and OOD performance is discussed in Sec.~\ref{sec:exp_format}.
\begin{itemize}[left=0pt, topsep=0pt, itemsep=4pt, parsep=0pt]
\item \textit{Image}: \citet{xu2025visual} use vision-only models, and their reasoning traces consist of images that show the current maze state after each move. %
While we do not directly employ this format in our experiments, as standard (multimodal) LLMs do not handle images in the reasoning process, it may be useful for future work on integrating visual elements in chain-of-thought.

\item \textit{Description}: %
This is a narration of the reasoning behind each step, discussing which moves would bring the player closer to the treasure, whether the path toward it is blocked by a lake, which are the feasible steps, and finally deciding which the next move should be.

\item \textit{Grid} and \textit{coordinates}: %
We represent the current map, after the next move, with our two text-based representations discussed above, for each step of a path to the treasure.

\item 
\textit{Grid + description}, \textit{coordinates + description}: %
We combine either \textit{grid} or \textit{coordinates} traces with \textit{description}. Thus, the chain-of-thought contains first a narrated description of the reasoning process about what the next step should be, and then the \textit{grid} or \textit{coordinates} representation of the map after applying the selected move.
These formats contain more information at the cost of being longer (and thus more expensive to generate at inference time, see Table~\ref{tab:response_length_success}).

\item \textit{A* with coordinates}: We also consider an algorithmic reasoning trace based on the exact trajectory of A* search in coordinate space. %
At each step, the trace writes the current search state as coordinate lists: the states waiting to be expanded, the states already expanded, and the parent links used to recover the final path. %
After the goal is reached, the trace follows the recorded parent links to reconstruct the path to the \treasure. %
Compared to the traces above, this format directly exposes the search procedure rather than example solutions, but it is also longer (Table~\ref{tab:response_length_success}).

\end{itemize}

\subsection{Solution multiplicity and training-set augmentation}
\label{sec:augmentation}

%
Beyond varying the representation of a map and the corresponding reasoning trace, our framework also allows us to control the amount of solution supervision per map.
While the standard training set pairs each map with a single solution trace (except A*-CoT models which take a different approach), \frozenlake maps often admit multiple correct paths to the goal.
We therefore construct augmented training data variants that keep the same (1000) training maps fixed, but expose the model to multiple valid solutions per map.
%
%
We consider three variants: up to three valid solutions per map, up to five valid solutions per map, and all optimal shortest solutions (total number of resulting training solutions in App.~\ref{sec:other_analyses}).
This factorizes the effect of solution diversity from the effect of seeing more maps: the environments are unchanged, while only the number and diversity of correct reasoning traces varies.
Unless stated otherwise, experiments use the default one-solution-per-map setting.
We study the effectiveness of these augmentation approaches in Sec.~\ref{sec:exp_augmentation}.

\subsection{Measuring OOD generalization via challenging distribution shifts}

We measure the out-of-distribution (OOD) generalization capabilities of the models along multiple axes, which control the difficulty of the task in different ways.
\begin{itemize}[left=0pt, topsep=0pt, itemsep=4pt, parsep=0pt]
    \item
\textbf{Map size:} %
First, by training on a range of different map sizes and testing on sizes not seen during training, we can evaluate whether the model generalizes to larger maps. Note that a larger map can contain the same maze as a smaller one, simply padded with empty cells or holes.

\item \textbf{Start-goal distance:} Next, we consider the $L_\infty$ distance between the start and the goal, i.e. for player coordinates $(s_1, s_2)$ and \treasure coordinates $(g_1, g_2)$ 
\[d_\infty = \max\{\vert s_1 - g_1\vert, \vert s_2 - g_2\vert\}.\]
For instance, in the 3×3 example in Fig.~\ref{fig:maze-repr}, this distance equals 2. In general, a map of size $n \times n$ %
has $d_\infty \leq n-1$. 
Therefore, we can test OOD generalization to maps with start-goal distance larger than what seen at training time.

\item \textbf{Direction and turn complexity:} Map size and start-goal distance do not fully determine the structure of a shortest solution. Let $\mathcal{P}^*(m)$ be the set of shortest valid paths for map $m$, and let $p=(a_1,\ldots,a_L)$ be one such path. We define its number of distinct movement directions and direction changes as
\[
\begin{aligned}
D(p) &= \left|\{a_i : 1 \leq i \leq L\}\right|, \\
T(p) &= \sum_{i=2}^{L}\mathbf{1}[a_i \neq a_{i-1}].
\end{aligned}
\]
Since a map can admit several shortest paths, we take the minimum of each measure over all shortest paths,
\[
\begin{aligned}
D^*(m) &= \min_{p\in\mathcal{P}^*(m)} D(p), \\
T^*(m) &= \min_{p\in\mathcal{P}^*(m)} T(p).
\end{aligned}
\]
We construct separate test sets for $D^*\in\{3,4\}$ and for $T^*\in\{5,6,7,8\}$, varying map size or start-goal distance at fixed $10\times10$ size. Using these minimum values avoids assigning a map to a harder set only because one of its shortest paths is unnecessarily complex.
\end{itemize}

\setlength{\intextsep}{10pt}
\makeatletter
\setlength{\@fptop}{0pt}
\setlength{\@fpbot}{0pt plus 1fil}
\setlength{\@dblfptop}{0pt}
\setlength{\@dblfpbot}{0pt plus 1fil}
\makeatother
\section{Experiments}
\label{sec:exp}

\subsection{Setup}

\textbf{Data.}
For training, we use the dataset introduced by \citet{yang2025machine}, which consists of 1,000 data points (100 3×3, 200 4×4, 300 5×5, and 400 6×6 maps). 
Thus, the training set only contains distances $d_\infty \leq 5$.
For each training example we generate the input and reasoning traces in all formats described in Sec.~\ref{sec:method}.
The in-distribution test sets (3×3, 4×4, 5×5, and 6×6 maps) are also identical to those in \citet{yang2025machine}, while we generated the out-of-distribution test sets via the Gym library \citep{brockman2016openaigym} as described in App.~\ref{sec:exp_details}. 
We generate 200 maps for each map size (or distance in the embedded maps case) in the corresponding OOD test sets. %
During evaluation, we extract the move sequence from the model output and simulate it in the environment. The sequence is correct if \textit{(i)} the player does not enter a cell with a hole and \textit{(ii)} the end position is on the \treasure.

\textbf{Models.}
As base model, we use Qwen2.5-VL-7B-Instruct \cite{bai2025qwen2}, following the setup of \citet{yang2025machine}, including their prompt and rule specification. All models are adapted via supervised fine-tuning for 10 epochs (details in App.~\ref{sec:exp_details}).
For each input format, we test the zero-shot performance of Qwen2.5-VL-7B-Instruct, and its versions fine-tuned without reasoning traces (no CoT).
Then, we train models 
on various combinations of input and CoT formats.

\subsection{Effect of input and CoT format}
\label{sec:exp_format}


\textbf{In-distribution performance.}
First, to test ID generalization, we measure the performance of the various models on maps of the same size as in the training set (3×3 to 6×6).
Table~\ref{tab:results_main} shows that zero-shot performance remains low across input formats, confirming that even small maps represent a challenging task for general-purpose LLMs.
\textit{Image} inputs (with and without CoT) perform consistently worse than text-only ones, highlighting the limitations of current multimodal LLMs in fully leveraging visual inputs.
CoT reasoning gives better accuracy than fine-tuning on the answers only, with the best results achieved by our \textit{coordinates} input with \textit{coordinates + description} (95\% accuracy) and A*-like CoT (98\%), which solves the task almost perfectly.

\begin{table*}[t!]
    \setlength{\tabcolsep}{2pt}
    \centering
    \footnotesize
    \extrarowheight=.9pt
    \caption{\textbf{Influence of data format on ID and OOD generalization.} %
    We report accuracy on both ID and OOD map sizes of zero-shot and fine-tuned LLMs using different combination of input and CoT traces (if any) formats.
    For OOD maps, we distinguish test sets where the distance between start and goal position is $d_\infty \geq 6$, i.e. not seen during training.
    \textit{Coordinates}-based representation substantially outperform other formats on OOD maps, especially when using 
    \textit{coordinates + description} CoT.
    A*-CoT over \textit{coordinates} input achieves the highest in-distribution accuracy.
    }
    \label{tab:results_main}
    \begin{adjustbox}{max width=\textwidth}
    \begin{tabular}{L{10mm} L{24mm} || *{4}{C{7.6mm}}|C{7.6mm}|| *{4}{C{7.6mm}}|C{7.6mm}|| *{4}{C{7.6mm}}|C{7.6mm}}
        \multirow{2}{*}{\makecell[l]{Input\\ format}} & \multirow{2}{*}{\makecell[l]{CoT\\ format}} & \multicolumn{5}{c||}{\textbf{ID test maps ($d_\infty \leq 5$)}} &\multicolumn{5}{c||}{\textbf{OOD maps (random $d_\infty$)}} & \multicolumn{5}{c}{\textbf{OOD maps ($d_\infty \geq 6$)}}\\
         &  & 3x3 & {4x4} & {5x5} & {6x6} & {Avg} & {7x7} & {8x8} & {9x9} & {10x10} & Avg & {7x7} & {8x8} & {9x9} & {10x10} & {Avg}\\
        \toprule
        
\multirow{3}{*}{Image} & zero-shot & 0.14 & 0.05 & 0.03 & 0.02 & 0.06 & 0.02 & 0.02 & 0.01 & 0.01 & 0.01 & 0.01 & 0.00 & 0.00 & 0.00 & 0.00\\
 & no CoT & 0.89 & 0.84 & 0.72 & 0.52 & 0.74 & 0.51 & 0.37 & 0.23 & 0.12 & 0.31 & 0.02 & 0.00 & 0.00 & 0.00 & 0.01\\
 & Descr. & 0.95 & 0.87 & 0.76 & 0.62 & 0.80 & 0.47 & 0.35 & 0.21 & 0.14 & 0.29 & 0.07 & 0.03 & 0.00 & 0.01 & 0.03\\
 \midrule

 
\multirow{5}{*}{Descr.} & zero-shot & 0.41 & 0.25 & 0.23 & 0.13 & 0.26 & 0.12 & 0.14 & 0.10 & 0.10 & 0.12 & 0.01 & 0.06 & 0.04 & 0.03 & 0.03\\
 & no CoT  & 0.92 & 0.91 & 0.86 & 0.64 & 0.83 & 0.65 & 0.55 & 0.49 & 0.41 & 0.52 & 0.19 & 0.08 & 0.04 & 0.02 & 0.08\\
 & Descr. & 0.94 & 0.92 & 0.90 & 0.67 & 0.86 & 0.67 & 0.63 & 0.55 & 0.46 & 0.58 & 0.07 & 0.02 & 0.01 & 0.01 & 0.02\\
 & Grid & 0.96 & 0.89 & 0.90 & 0.70 & 0.86 & 0.59 & 0.47 & 0.28 & 0.12 & 0.36 & 0.35 & 0.17 & 0.06 & 0.01 & 0.15\\
 & Grid + Descr. & 0.98 & 0.93 & 0.93 & 0.69 & 0.88 & 0.70 & 0.51 & 0.41 & 0.26 & 0.47 & 0.44 & 0.28 & 0.11 & 0.07 & 0.23\\
\midrule


\multirow{5}{*}{Grid} & zero-shot & 0.14 & 0.13 & 0.06 & 0.03 & 0.09 & 0.02 & 0.01 & 0.00 & 0.02 & 0.01 & 0.00 & 0.01 & 0.01 & 0.01 & 0.01\\
 & no CoT  & 0.95 & 0.89 & 0.83 & 0.62 & 0.82 & 0.55 & 0.39 & 0.34 & 0.26 & 0.38 & 0.04 & 0.02 & 0.00 & 0.00 & 0.01\\
 & Descr. & 0.94 & 0.91 & 0.86 & 0.62 & 0.83 & 0.53 & 0.41 & 0.35 & 0.25 & 0.39 & 0.10 & 0.07 & 0.01 & 0.01 & 0.05\\
& Grid & 0.98 & 0.92 & 0.97 & 0.71 & 0.90 & 0.74 & 0.51 & 0.39 & 0.24 & 0.47 & 0.43 & 0.22 & 0.07 & 0.04 & 0.19\\
 & Grid + Descr. & 0.99 & 0.93 & 0.92 & 0.79 & 0.91 & 0.83 & 0.68 & 0.49 & 0.35 & 0.59 & 0.65 & 0.55 & 0.23 & 0.20 & 0.41\\
\midrule


\multirow{5}{*}{Coord.} & zero-shot & 0.30 & 0.20 & 0.18 & 0.10 & 0.20 & 0.13 & 0.08 & 0.08 & 0.11 & 0.10 & 0.04 & 0.05 & 0.04 & 0.02 & 0.03 \\
 & no CoT  &  0.94 & 0.89 & 0.92 & 0.72 & 0.87 & 0.78 & 0.60 & 0.55 & 0.47 & 0.60 & 0.21 & 0.06 & 0.02 & 0.01 & 0.07 \\
 & Descr. & 0.98 &  0.96 & 0.96 & 0.75 & 0.91 & 0.72 & 0.63 & 0.55 & 0.49 & 0.59 & 0.14 & 0.13 & 0.04 & 0.04 & 0.09 \\
& Coord. & \textbf{1.00} & 0.93 & 0.94 & 0.84 & 0.93 & 0.93 & 0.88 & 0.88 & 0.83 & 0.88 & 0.83 & 0.80 & 0.68 & 0.70 & 0.75 \\
 & Coord. + Descr. & \textbf{1.00} & 0.95 & \textbf{0.98} & 0.86 & 0.95 & \textbf{0.97} & \textbf{0.93} & \textbf{0.90} & \textbf{0.88} & \textbf{0.92} & \textbf{0.89} & \textbf{0.87} & \textbf{0.81} & \textbf{0.77} & \textbf{0.83} \\
\midrule







Coord. & A* w/ Coord.
& \textbf{1.00} & \textbf{0.99} & 0.96 & \textbf{0.95} & \textbf{0.98} & 0.96 & 0.91 & 0.88 & 0.82 & 0.89 & 0.87 & 0.76 & 0.73 & 0.62 & 0.74 \\


\bottomrule
\end{tabular}
    \end{adjustbox}
\vspace{6mm}
\end{table*}


\textbf{OOD generalization: map size.} Our first OOD test set includes larger map sizes, not seen during training, i.e. 7×7, 8×8, 9×9, and 10×10.
Similar to ID data, these maps are generated with random sampling of the initial position of player and \treasure, without controlling for the start-goal distance (\textit{random} $d_\infty$ column
in Table~\ref{tab:results_main}).
For some maps, the start, goal and solution can fit into a map of size smaller than or equal to 6×6: these examples are OOD in terms of map size but not start-goal distance, as can be seen in Table~\ref{tab:avg_distance} (in appendix).
Similar to the ID case, models with text-based input significantly outperform the image-based ones. %
Again, models encoding maps in \textit{coordinates} format with \textit{coord.} (88\% accuracy), \textit{coord. + descr.} (92\%) and A* CoT (89\%) far outperform all others.

\textbf{OOD generalization: start-goal distance.}
To ensure that both map size and start-goal distance are out-of-distribution, we generate new OOD maps enforcing $d_\infty \geq 6$, i.e., the distance between player and \treasure is too large to fit in a 6×6 map.
As shown in \cref{tab:results_main} (right column), in this more challenging distribution shift, the accuracy of most models quickly drops near 0\% for larger maps.
%
%
While the best results are still obtained, with a large margin, by the \textit{coordinates} format (83\% average accuracy), all models combining a map encoding and the text description in the reasoning process (\textit{grid/coord. + descr.}) achieve the best performance among those with same input format.
We conjecture that generating a (more or less ``visual'') representation of the current map (as \textit{grid} or \textit{coordinates}) at each step helps tracking the progress of the map navigation, while the reasoning in the natural language (the \textit{description} CoT component) is better suited for the model to elaborate on the next move.
%
Interestingly, the accuracy of the model using \textit{description} format for both input and CoT sharply drops from 58\% on OOD maps with random $d_\infty$ to 7\% on 7×7 maps with $d_\infty \geq 6$: this demonstrates 
how our controlled setup can reveal severe limitations in generalization.

\newpage
\textbf{OOD generalization: start-goal distance with fixed map size.} 
To fully disentangle the effects of map size and $d_\infty$, we embedded all training images (of size up to 6×6) randomly into 10×10 maps, which does not change the correct path but fixes the map size.
Moreover, we generate a new test set of 10×10 maps with 200 samples for each value of $d_\infty\in\{2, \ldots, 9\}$.
This setup allows us to analyze generalization w.r.t. start-goal distance in isolation, without requiring the model to learn or generalize to maps of different size.
Table~\ref{tab:results_embedded_maps} shows the in-distribution ($d_\infty \leq 5$) and out-of-distribution ($d_\infty \geq 6$) accuracy for different data formats.\par 
Similarly to the previous setups, \textit{image} inputs yield worse results than text-based ones, and the combined CoT versions the best results.
Moreover, for both no CoT and \textit{description} CoT the accuracy drops to near 0\% already at $d_\infty=6$, i.e. the smallest distribution shift, indicating that the models do not typically learn an algorithmic solution to the task.
%
Instead, the A*-like reasoning significantly outperforms other models in both ID (95\%) and OOD (62\%).

\newpage
\textbf{OOD generalization: optimal solution length.} 
Next, we study generalization with regard to the length of optimal solution $l$.
Fig.~\ref{fig:sol-len-coord} (left plot) shows the distribution of the length of the shortest solution of each map for training and test set, aggregating the statistics over map sizes.
As expected, including the larger OOD maps in the test set introduces a distribution shift towards longer solutions, with $l\leq 14$ for the training set, up to $18$ for the test set.
Fig.~\ref{fig:sol-len-coord} (right) reports the accuracy of the models using \textit{coordinates} input representations vs the solution length of the test maps.
The accuracy of the LLMs with no CoT or \textit{description} CoT drops near zero for $l \geq 9$, where very few training examples are available.
Conversely, with \textit{coordinates}, \textit{coord. + descr.} and A* CoT the accuracy remains non-trivial until $l=15$, yet failing afterwards.
Therefore, generalization to maps with longer solutions is challenging, but even in this case our CoT representations greatly help achieving better accuracy.
The results for all input formats can be found in Fig.~\ref{fig:sol-len-all}.

\FloatBarrier
\begin{table*}[p]
\begingroup
\captionsetup{type=figure}
\begingroup
\renewenvironment{figure}[1][]{\begingroup}{\endgroup}%
\renewenvironment{figure*}[1][]{\begingroup}{\endgroup}%
\begin{figure*}[t]
    \centering
    \includegraphics[
    height=44mm
    ]{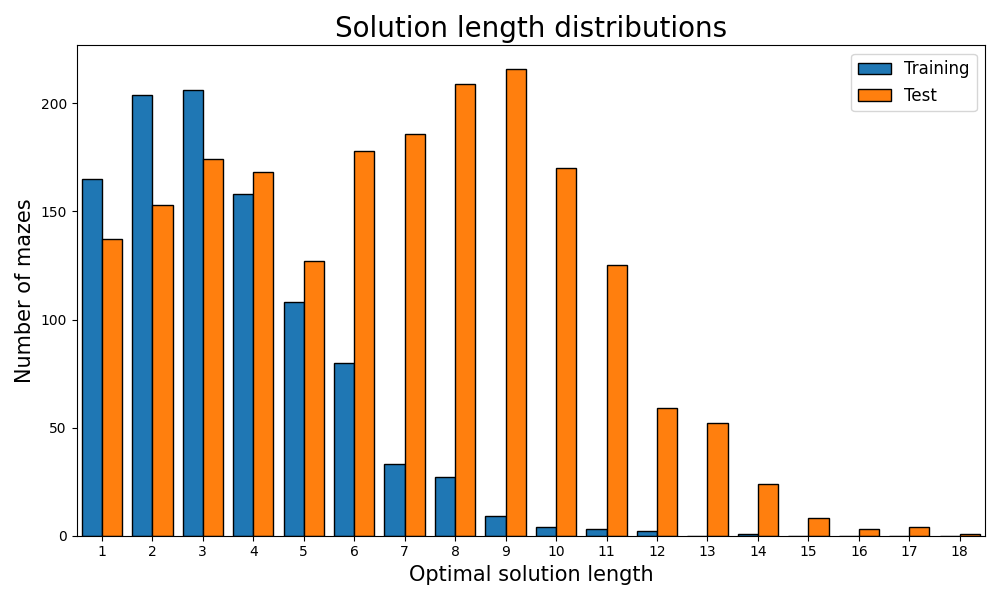}
    \includegraphics[
    height=46mm
    ]{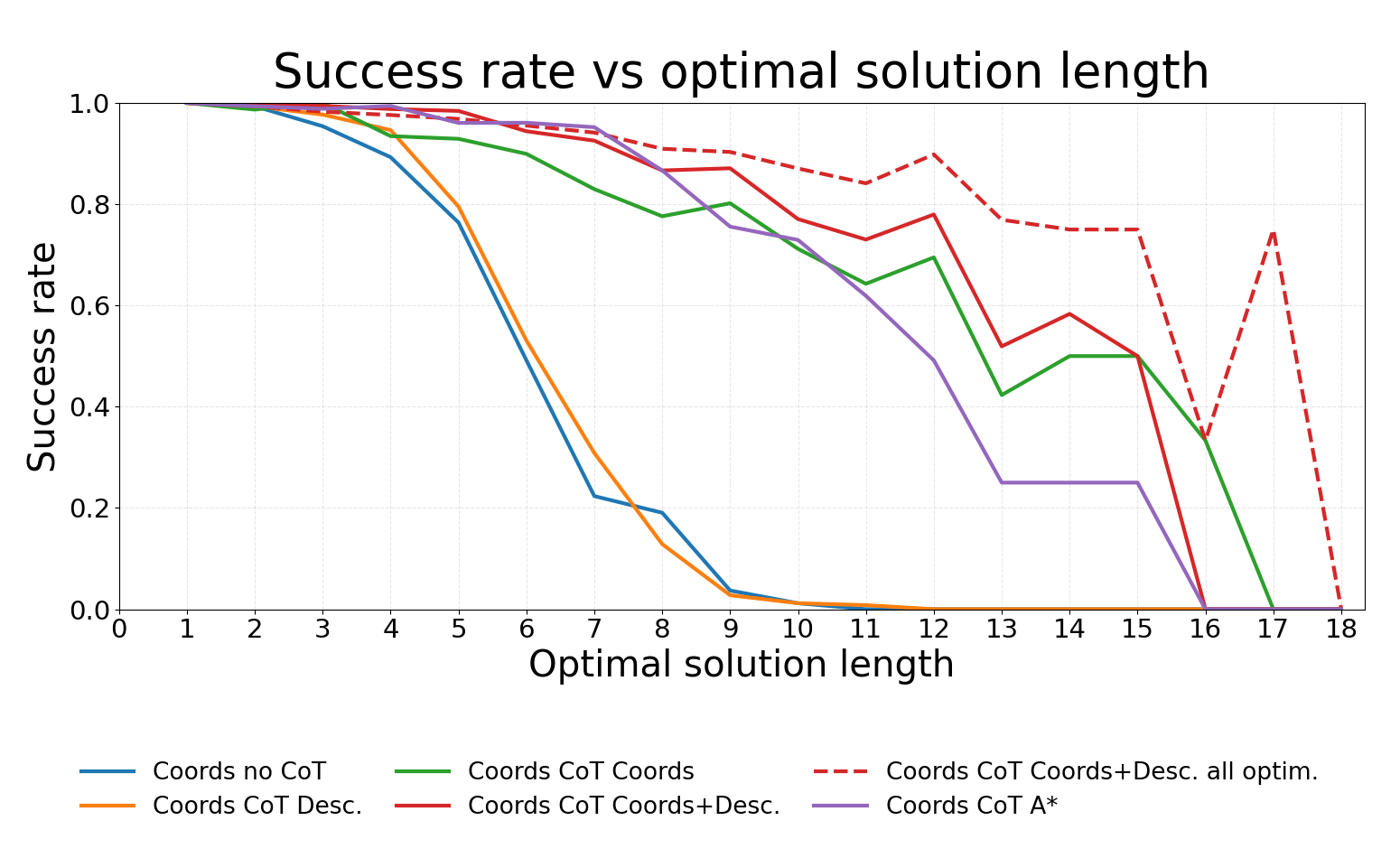}
    \vspace{-2mm}
    \caption{\textbf{OOD generalization w.r.t. optimal solution length.} \textbf{Left:} distribution of the length of the optimal (shortest) solution paths for both training and test maps (aggregated across map sizes).
    \textbf{Right:} success rate of models using \textit{coordinates} input vs solution length.
    }   
    \label{fig:sol-len-coord}
    \vspace{4mm}
\end{figure*}
\endgroup
\endgroup
\par\medskip
  \begingroup
  \renewenvironment{table}[1][]{\par\begingroup}{\par\endgroup}%
  \renewenvironment{table*}[1][]{\par\begingroup}{\par\endgroup}%
  \begin{table*}[t]
    \setlength{\tabcolsep}{6pt}
    \centering
    \small 
    \footnotesize
    \extrarowheight=.9pt
    \caption{\textbf{Generalization w.r.t. start-goal distance ($d_\infty$) with fixed 10×10 map size.}
    We compare different input and CoT format when the training maps have fixed size 10x10 and $d_\infty < 6$, test maps size 10×10 and $d_\infty \in \{2, \ldots, 5\}$ (ID) or $d_\infty \in \{6, \ldots,  9\}$ (OOD).
    Coordinate-based CoTs improve distance extrapolation substantially over grid-based traces, and coordinate + description traces remain accurate well beyond the training distance range.
    A* traces over coordinates perform best in this fixed-map-size setting, achieving the strongest ID and OOD accuracy across distances.
    }
    \label{tab:results_embedded_maps}
    \begin{tabular}{ll|| cccc|c|| cccc|c}
        \multirow{2}{*}{\makecell[l]{Input\\ format}} & \multirow{2}{*}{\makecell[l]{CoT\\ format}} & \multicolumn{5}{c||}{\textbf{ID test maps ($d_\infty \leq 5$)}}  & \multicolumn{5}{c}{\textbf{OOD maps ($d_\infty \geq 6$)}}\\
        & & 2 & 3 & 4 & 5 & Avg & 6 & 7 & 8 & 9 & Avg\\ 
        \toprule

\multirow{2}{*}{Image} & no CoT & 0.70 & 0.46 & 0.21 & 0.03 & 0.35 & 0.03 & 0.01 & 0.00 & 0.00 & 0.01\\
 & Descr. & 0.69 & 0.41 & 0.16 & 0.06 & 0.33 & 0.01 & 0.00 & 0.00 & 0.00 & 0.00\\
\midrule

\multirow{4}{*}{Descr.} & no CoT & 0.82 & 0.75 & 0.57 & 0.34 & 0.62 & 0.01 & 0.01 & 0.00 & 0.00 & 0.01\\
 & Descr. & 0.89 & 0.80 & 0.68 & 0.41 & 0.69 & 0.00 & 0.00 & 0.00 & 0.00 & 0.00\\
  & Grid & 0.82 & 0.76 & 0.67 & 0.49 & 0.69 & 0.34 & 0.20 & 0.10 & 0.04 & 0.17\\
 & Grid + Descr. & 0.82 & 0.73 & 0.61 & 0.47 & 0.66 & 0.36 & 0.25 & 0.17 & 0.14 & 0.23 \\
\midrule


\multirow{4}{*}{Grid} & no CoT & 0.86 & 0.81 & 0.55 & 0.12 & 0.58 & 0.00 & 0.01 & 0.00 & 0.00 & 0.00\\
 & Descr. & 0.84 & 0.78 & 0.45 & 0.10 & 0.54 & 0.00 & 0.01 & 0.00 & 0.00 & 0.00\\
  & Grid & 0.89 & 0.84 & 0.77 & 0.62 & 0.78 & 0.46 & 0.34 & 0.17 & 0.08 & 0.26\\
 & Grid + Descr. & 0.86 & 0.84 & 0.74 & 0.61 & 0.76 & 0.50 & 0.33 & 0.21 & 0.10 & 0.29 \\
 \midrule


\multirow{4}{*}{Coord.} 

 & no CoT 
& 0.90 & 0.83 & 0.69 & 0.43 & 0.71 & 0.01 & 0.00 & 0.00 & 0.00 & 0.00 \\

 & Descr. 
& 0.84 & 0.73 & 0.65 & 0.32 & 0.63 & 0.00 & 0.00 & 0.00 & 0.00 & 0.00 \\

 & Coord. 
& 0.89 & 0.85 & 0.79 & 0.64 & 0.79 & 0.53 & 0.39 & 0.31 & 0.21 & 0.36 \\

 & Coord. + Descr. 
& 0.95 & 0.94 & 0.89 & 0.74 & 0.88 & 0.75 & 0.56 & 0.44 & 0.44 & 0.55 \\

\midrule







Coord. & A* w/ Coord. 
& \textbf{0.99} & \textbf{0.98} & \textbf{0.96} & \textbf{0.89} & \textbf{0.95} & \textbf{0.85} & \textbf{0.63} & \textbf{0.56} & \textbf{0.46} & \textbf{0.62} \\


\bottomrule

\end{tabular}
\vspace{8mm}
    
\end{table*} 
  \endgroup

\par\vspace{10pt}
\noindent\begin{minipage}[t]{\columnwidth}\vspace{0pt}
\normalsize
\subsection{Effect of solution-set augmentation}
\label{sec:exp_augmentation}

We then study the effect of the training data variants with multiple solutions per map discussed in Sec.~\ref{sec:augmentation}.
We use \textit{coordinates} input with \textit{coordinates + description} CoT, which gives the strongest OOD results on larger maps, to fine-tune models on these three training sets.
We report the results of these models, together with the default one-solution-per-map setting, in Table~\ref{tab:other_training_data} (varying map size) and Table~\ref{tab:other_training_data_embedded_maps} (fixed map size).
\end{minipage}\hfill
\noindent\begin{minipage}[t]{\columnwidth}\vspace{0pt}
\normalsize
We see that the augmented training datasets yield in almost all cases notable improvements in both ID and, especially, OOD performance.
In particular, using all optimal solutions increases OOD accuracy on larger maps with $d_\infty \geq 6$ from 83\% to 91\% (Table~\ref{tab:other_training_data}), and from 55\% to 85\% on OOD maps with fixed 10x10 size (Table~\ref{tab:other_training_data_embedded_maps}).
This setting gives improvements even in generalization w.r.t. optimal solution length, see Fig.~\ref{fig:sol-len-coord} (right).
This demonstrates that augmenting training maps with additional correct solutions is a complementary approach to optimizing data format for improving generalization to distribution shifts.

\end{minipage}\par
\end{table*}

\FloatBarrier
\begin{table*}[!t]
\begingroup
\renewenvironment{table}[1][]{\par\begingroup}{\par\endgroup}%
\renewenvironment{table*}[1][]{\par\begingroup}{\par\endgroup}%
\begin{table*}[!tbp]
    \setlength{\tabcolsep}{1.7pt}
    \centering
    \footnotesize
    \extrarowheight=.9pt
    \caption{\textbf{Effect of solution-set augmentation on generalization w.r.t. map size.} %
    We report ID and OOD accuracy of LLMs fine-tuned on training sets with the same 1000 maps but different number of solutions per map. All models use \textit{coord.} input and \textit{coord. + descr.} CoT.
    }
    \label{tab:other_training_data}
    \begin{tabular}{L{27mm} 
    || *{4}{C{8mm}}|C{8mm}|| *{4}{C{8mm}}|C{8mm}|| *{4}{C{8mm}}|C{8mm}}
        \multirow{2}{*}{\makecell[l]{Training data \\ variant}} 
        & \multicolumn{5}{c||}{\textbf{ID test maps ($d_\infty \leq 5$)}} &\multicolumn{5}{c||}{\textbf{OOD maps (random $d_\infty$)}} & \multicolumn{5}{c}{\textbf{OOD maps ($d_\infty \geq 6$)}}\\
          & 3x3 & {4x4} & {5x5} & {6x6} & {Avg} & {7x7} & {8x8} & {9x9} & {10x10} & Avg & {7x7} & {8x8} & {9x9} & {10x10} & {Avg}\\
        \toprule

1 sol/map (default) & \textbf{1.00} & 0.95 & 0.98 & 0.86 & 0.95 & 0.97 & 0.93 & 0.90 & 0.88 & 0.92 & 0.89 & 0.87 & 0.81 & 0.77 & 0.83 \\

up to 3 sol/map & \textbf{1.00} & \textbf{0.98} & 0.98 & 0.87 & \textbf{0.96} & 0.98 & 0.96 & \textbf{0.95} & \textbf{0.92} & \textbf{0.95} & 0.92 & \textbf{0.90} & 0.87 & 0.85 & 0.89 \\

up to 5 sol/map &  \textbf{1.00} & 0.96 & \textbf{0.99} & \textbf{0.90} & \textbf{0.96} & \textbf{0.99} & 0.95 & 0.94 & \textbf{0.92} & \textbf{0.95} & 0.92 & 0.89 & 0.82 & 0.83 & 0.86 \\

all optimal solutions & \textbf{1.00} & 0.94 & 0.94 & 0.84 & 0.93 & 0.97 & \textbf{0.97} & \textbf{0.95} & \textbf{0.92} & \textbf{0.95} & \textbf{0.94} & 0.88 & \textbf{0.91} & \textbf{0.90} & \textbf{0.91} \\

\bottomrule

\end{tabular}
\vspace{4mm}

\end{table*}
\endgroup
\par\medskip
\begingroup
\renewenvironment{table}[1][]{\par\begingroup}{\par\endgroup}%
\renewenvironment{table*}[1][]{\par\begingroup}{\par\endgroup}%
\begin{table*}[!tbp]
    \setlength{\tabcolsep}{1.9pt}
    \centering
    \footnotesize
    \extrarowheight=.9pt
    \caption{\textbf{Effect of solution-set augmentation on generalization w.r.t. start-goal distance (10x10 maps).}
    We report ID and OOD accuracy of LLMs fine-tuned on the same 1000 maps but with different numbers of solutions per map. All models use \textit{coord.} input and \textit{coord. + descr.} CoT.
    }
    \label{tab:other_training_data_embedded_maps}
\begin{tabular}{L{30mm} || *{4}{C{8mm}}|C{8mm}|| *{4}{C{8mm}}|C{8mm}}
        \multirow{2}{*}{\makecell[l]{Training data  variant}} & \multicolumn{5}{c||}{\textbf{ID test maps ($d_\infty \leq 5$)}}  & \multicolumn{5}{c}{\textbf{OOD maps ($d_\infty \geq 6$)}}\\
        &  2 & 3 & 4 & 5 & Avg & 6 & 7 & 8 & 9 & Avg\\ 
        \toprule

1 sol/map (default) & 0.95 & 0.94 & 0.89 & 0.74 & 0.88 & 0.75 & 0.56 & 0.44 & 0.44 & 0.55 \\

up to 3 sol/map 
& 0.94 & 0.95 & 0.92 & 0.87 & 0.92 & 0.80 & 0.69 & 0.52 & 0.48 & 0.62 \\

up to 5 sol/map 
& 0.95 & 0.97 & \textbf{0.95} & \textbf{0.93} & \textbf{0.95} & \textbf{0.94} & 0.88 & \textbf{0.82} & 0.77 & \textbf{0.85} \\

all optimal solutions 
& \textbf{0.96} & \textbf{0.99} & \textbf{0.95} & 0.89 & \textbf{0.95} & 0.92 & \textbf{0.89} & 0.81 & \textbf{0.79} & \textbf{0.85} \\
\bottomrule

\end{tabular}
\end{table*}
\endgroup
\end{table*}

\subsection{Results with another model family}
\label{sec:qwen35_results}

We repeat the experiments with three Qwen3.5 models using \textit{coordinates} input \citep{qwen35}. Table~\ref{tab:qwen35_map} summarizes generalization across map sizes and start-goal distances~($d_\infty$). App.~\ref{sec:qwen35_full_results} gives the full results.

\par
\noindent\begin{minipage}{\columnwidth}
\captionsetup{type=table}
  \begingroup
  \renewenvironment{table}[1][]{\par\begingroup}{\par\endgroup}%
  \renewenvironment{table*}[1][]{\par\begingroup}{\par\endgroup}%
  \begin{table}[!bp]
\centering\footnotesize
\setlength{\tabcolsep}{2.35pt}
\extrarowheight=.9pt
\caption{\textbf{Accuracy of zero-shot and fine-tuned Qwen3.5 models by model size and CoT format.} We report average accuracy for ID sizes 3$\times$3--6$\times$6 and OOD sizes 7$\times$7--10$\times$10. In Tables~\mbox{\ref{tab:qwen35_map}--\ref{tab:hard_scaling_map}}, all models use coordinate input and distance tests use $10\times10$ maps. ID distances ($d_\infty$) are 2--5; OOD distances are 6--9.}
\label{tab:qwen35_map}
\label{tab:qwen35_distance}
\begin{tabular}{@{}ll||ccc||cc@{}}
\multirow{2}{*}[-6.2pt]{Model} & \multirow{2}{*}[-6.2pt]{CoT format} & \multicolumn{3}{c||}{By map size} & \multicolumn{2}{c}{By $d_\infty$}\\
\noalign{\vskip-\dimexpr\aboverulesep+.5\cmidrulewidth\relax}
\cmidrule(lr){3-5}\cmidrule(lr){6-7}
\noalign{\vskip-\dimexpr\belowrulesep+.5\cmidrulewidth\relax}
 &  & \rule{0pt}{13pt}ID & \makecell{OOD\\random} & \makecell{OOD\\$d_\infty\geq6$} & ID & OOD\\
\midrule
\multirow{7}{*}{0.8B} & zero-shot & 0.49 & 0.29 & 0.15 & 0.25 & 0.12\\
 & Coord. & 0.92 & 0.80 & 0.50 & 0.82 & 0.20\\
 & Coord. + Descr. & 0.96 & 0.89 & 0.66 & 0.88 & 0.20\\
 & A* w/ Coord. & \textbf{0.99} & 0.90 & \textbf{0.75} & 0.96 & 0.47\\
 & $K=3$ aug. & 0.97 & \textbf{0.91} & 0.69 & 0.96 & 0.52\\
 & $K=5$ aug. & 0.98 & 0.90 & 0.67 & \textbf{0.98} & 0.52\\
 & all-opt aug. & 0.96 & 0.90 & 0.70 & 0.97 & \textbf{0.53}\\
\midrule
\multirow{7}{*}{2B} & zero-shot & 0.98 & 0.89 & 0.78 & 0.92 & 0.64\\
 & Coord. & 0.95 & 0.88 & 0.61 & 0.88 & 0.26\\
 & Coord. + Descr. & 0.97 & 0.87 & 0.50 & 0.94 & 0.37\\
 & A* w/ Coord. & \textbf{0.99} & \textbf{0.95} & 0.83 & \textbf{0.98} & \textbf{0.76}\\
 & $K=3$ aug. & 0.97 & \textbf{0.95} & \textbf{0.84} & 0.97 & 0.59\\
 & $K=5$ aug. & \textbf{0.99} & 0.94 & 0.80 & \textbf{0.98} & 0.57\\
 & all-opt aug. & 0.96 & 0.93 & 0.79 & 0.96 & 0.59\\
\midrule
\multirow{7}{*}{4B} & zero-shot & 0.47 & 0.33 & 0.26 & 0.32 & 0.18\\
 & Coord. & 0.97 & 0.96 & 0.90 & 0.95 & 0.79\\
 & Coord. + Descr. & 0.97 & 0.95 & 0.87 & 0.95 & 0.68\\
 & A* w/ Coord. & \textbf{1.00} & \textbf{0.97} & 0.88 & \textbf{0.99} & \textbf{0.89}\\
 & $K=3$ aug. & 0.97 & \textbf{0.97} & \textbf{0.93} & 0.98 & 0.80\\
 & $K=5$ aug. & 0.98 & 0.96 & 0.88 & 0.98 & 0.80\\
 & all-opt aug. & 0.98 & 0.96 & 0.92 & 0.96 & 0.71\\
\bottomrule
\end{tabular}
\end{table}
  \endgroup

\end{minipage}\par
\newpage
Again, A* gives strong ID and OOD results for all three models, while solution-set augmentation improves OOD accuracy with \textit{coordinates + description}.

\subsection{Structural generalization}
\label{sec:structural_stress}

We then evaluate maps whose shortest solutions require direction and turn counts rarely observed during training (Table~\ref{tab:structural_counts} and Fig.~\ref{fig:structural-distributions} in the appendix). Tables~\ref{tab:direction_complexity_map} and~\ref{tab:turn_complexity_map} report average accuracy on the test sets defined in Sec.~\ref{sec:method}; App.~\ref{sec:structural_full_results} provides the full breakdowns.

\begin{table}[!htbp]
\centering\footnotesize
\setlength{\tabcolsep}{2.2pt}
\extrarowheight=.9pt
\caption{\textbf{Generalization to maps requiring three or four directions.} We report average ID and OOD accuracy on directionally complex maps, excluding unavailable test configurations.}
\label{tab:direction_complexity_map}
\label{tab:direction_complexity_distance}
\begin{tabular}{@{}ll||ccc||cc@{}}
\multirow{2}{*}[-6.2pt]{Dirs.} & \multirow{2}{*}[-6.2pt]{CoT format} & \multicolumn{3}{c||}{By map size} & \multicolumn{2}{c}{By $d_\infty$}\\
\noalign{\vskip-\dimexpr\aboverulesep+.5\cmidrulewidth\relax}
\cmidrule(lr){3-5}\cmidrule(lr){6-7}
\noalign{\vskip-\dimexpr\belowrulesep+.5\cmidrulewidth\relax}
 &  & \rule{0pt}{13pt}ID & \makecell{OOD\\random} & \makecell{OOD\\$d_\infty\geq6$} & ID & OOD\\
\midrule
\multirow{5}{*}{3} & Coord. + Descr. & 0.61 & 0.22 & 0.15 & 0.20 & 0.03\\
 & A* w/ coord. & \textbf{0.88} & \textbf{0.48} & \textbf{0.35} & \textbf{0.60} & 0.18\\
 & $K=3$ aug. & 0.76 & 0.37 & 0.32 & 0.39 & 0.16\\
 & $K=5$ aug. & 0.81 & 0.43 & \textbf{0.35} & 0.50 & \textbf{0.27}\\
 & all-opt aug. & 0.58 & 0.26 & 0.23 & 0.34 & 0.15\\
\midrule
\multirow{5}{*}{4} & Coord. + Descr. & 0.14 & 0.02 & 0.00 & 0.00 & 0.00\\
 & A* w/ coord. & \textbf{0.52} & \textbf{0.17} & \textbf{0.04} & \textbf{0.12} & \textbf{0.01}\\
 & $K=3$ aug. & 0.36 & 0.08 & 0.01 & 0.04 & 0.00\\
 & $K=5$ aug. & 0.45 & 0.13 & 0.03 & 0.06 & \textbf{0.01}\\
 & all-opt aug. & 0.13 & 0.03 & 0.00 & 0.02 & 0.00\\
\bottomrule
\end{tabular}
\end{table}

\FloatBarrier
\textbf{Number of directions.} Accuracy sharply decreases with the number of directions (Table~\ref{tab:direction_complexity_map}). A* gives the best results on larger maps, but its OOD accuracy drops from 48\% with three directions to 17\% with four. On fixed maps, all models are near zero with four directions.

{\clubpenalty=10000 \textbf{Number of turns.} Maps requiring more turns are even more challenging (Table~\ref{tab:turn_complexity_map}). With A*, OOD accuracy on larger maps decreases from 11\% with five turns to 1\% with eight and remains low on fixed maps. Thus, strong generalization to larger maps and start-goal distances does not ensure success on structurally complex maps.\par}

\begin{table}[!htbp]
\centering\footnotesize
\setlength{\tabcolsep}{2.2pt}
\extrarowheight=.9pt
\vspace{2mm}
\caption{\textbf{Generalization to maps requiring five to eight turns.} We report average accuracy for each turn count separately for the map size and distance tests, using only available configurations.}
\label{tab:turn_complexity_map}
\label{tab:turn_complexity_distance}
\begin{tabular}{@{}ll||ccc||cc@{}}
\multirow{2}{*}[-6.2pt]{Turns} & \multirow{2}{*}[-6.2pt]{CoT format} & \multicolumn{3}{c||}{By map size} & \multicolumn{2}{c}{By $d_\infty$}\\
\noalign{\vskip-\dimexpr\aboverulesep+.5\cmidrulewidth\relax}
\cmidrule(lr){3-5}\cmidrule(lr){6-7}
\noalign{\vskip-\dimexpr\belowrulesep+.5\cmidrulewidth\relax}
 &  & \rule{0pt}{13pt}ID & \makecell{OOD\\random} & \makecell{OOD\\$d_\infty\geq6$} & ID & OOD\\
\midrule
\multirow{3}{*}{5} & Coord. + Descr. & 0.06 & 0.02 & 0.04 & 0.00 & 0.00\\
 & A* w/ coord. & \textbf{0.46} & 0.11 & \textbf{0.13} & \textbf{0.10} & \textbf{0.09}\\
 & $K=3$ aug. & 0.24 & \textbf{0.12} & \textbf{0.13} & 0.04 & 0.06\\
\midrule
\multirow{3}{*}{6} & Coord. + Descr. & 0.02 & 0.01 & 0.02 & 0.00 & 0.00\\
 & A* w/ coord. & \textbf{0.23} & 0.06 & \textbf{0.08} & \textbf{0.04} & \textbf{0.03}\\
 & $K=3$ aug. & 0.14 & \textbf{0.07} & \textbf{0.08} & 0.02 & 0.02\\
\midrule
\multirow{3}{*}{7} & Coord. + Descr. & 0.01 & 0.01 & 0.01 & 0.00 & 0.00\\
 & A* w/ coord. & \textbf{0.20} & 0.02 & 0.04 & \textbf{0.02} & \textbf{0.01}\\
 & $K=3$ aug. & 0.07 & \textbf{0.03} & \textbf{0.05} & 0.01 & \textbf{0.01}\\
\midrule
\multirow{3}{*}{8} & Coord. + Descr. & 0.00 & 0.00 & 0.00 & \textbf{0.00} & 0.00\\
 & A* w/ coord. & \textbf{0.10} & \textbf{0.01} & \textbf{0.02} & \textbf{0.00} & \textbf{0.01}\\
 & $K=3$ aug. & 0.05 & \textbf{0.01} & \textbf{0.02} & \textbf{0.00} & \textbf{0.01}\\
\bottomrule
\end{tabular}
\end{table}

\FloatBarrier
\subsection{Continued fine-tuning on directionally complex maps}
\label{sec:hard_scaling}

\looseness=-1
We continue fine-tuning A* and \textit{coordinates + description} on 1,000 new maps with three- or four-direction shortest solutions. We train one model per setting and test on disjoint maps with both counts (Table~\ref{tab:hard_scaling_map}).

\par\noindent\begin{minipage}{\columnwidth}\captionsetup{type=table}
\centering\footnotesize
\setlength{\tabcolsep}{2.2pt}
\extrarowheight=.9pt
\vspace{4mm}
\caption{\textbf{Effect of continued fine-tuning on ID and OOD accuracy.} Average accuracy is shown for the original checkpoint (Base) and checkpoints fine-tuned on 1,000 additional maps. Train$\to$test gives minimum direction counts for training and evaluation.}
\label{tab:hard_scaling_map}
\label{tab:hard_scaling_distance}
\begin{tabular}{@{}ll||ccc||cc@{}}
\multirow{2}{*}[-6.2pt]{CoT} & \multirow{2}{*}[-6.2pt]{Train$\to$test} & \multicolumn{3}{c||}{By map size} & \multicolumn{2}{c}{By $d_\infty$}\\
\noalign{\vskip-\dimexpr\aboverulesep+.5\cmidrulewidth\relax}
\cmidrule(lr){3-5}\cmidrule(lr){6-7}
\noalign{\vskip-\dimexpr\belowrulesep+.5\cmidrulewidth\relax}
 &  & \rule{0pt}{13pt}ID & \makecell{OOD\\random} & \makecell{OOD\\$d_\infty\geq6$} & ID & OOD\\
\midrule
\multirow{6}{*}{A*} & base$\to$3 & 0.88 & 0.48 & 0.35 & 0.60 & 0.18\\
 & base$\to$4 & 0.52 & 0.17 & 0.04 & 0.12 & 0.01\\
 & $3\to3$ & 0.96 & 0.65 & 0.47 & 0.85 & 0.44\\
 & $3\to4$ & 0.84 & 0.33 & 0.10 & 0.30 & 0.04\\
 & $4\to3$ & \textbf{0.97} & \textbf{0.72} & \textbf{0.60} & \textbf{0.91} & \textbf{0.71}\\
 & $4\to4$ & \textbf{0.96} & \textbf{0.55} & \textbf{0.32} & \textbf{0.72} & \textbf{0.41}\\
\midrule
\multirow{6}{*}{\makecell[l]{Coord.\\+ Descr.}} & base$\to$3 & 0.61 & 0.22 & 0.15 & 0.20 & 0.03\\
 & base$\to$4 & 0.14 & 0.02 & 0.00 & 0.00 & 0.00\\
 & $3\to3$ & \textbf{0.90} & \textbf{0.50} & \textbf{0.40} & \textbf{0.68} & \textbf{0.36}\\
 & $3\to4$ & 0.52 & 0.12 & 0.06 & 0.16 & 0.04\\
 & $4\to3$ & 0.73 & 0.41 & 0.31 & 0.53 & 0.34\\
 & $4\to4$ & \textbf{0.85} & \textbf{0.36} & \textbf{0.18} & \textbf{0.45} & \textbf{0.19}\\
\bottomrule
\end{tabular}
\end{minipage}\par

\FloatBarrier
{\clubpenalty=10000 With A*, 1,000 additional examples improve accuracy on directionally complex maps. Using 1,000 four-direction maps increases OOD accuracy from 17\% to 55\% on larger maps and from 1\% to 41\% on fixed maps.\par}

{\clubpenalty=10000 Continued fine-tuning also improves the results with \textit{coordinates + description}, although by less than A*. The~A*~checkpoint trained on 1,000 four-direction maps also improves results on maps requiring more turns without turn-targeted fine-tuning (Tables~\ref{tab:direction_to_turn_transfer_map} and~\ref{tab:direction_to_turn_transfer_distance} in the appendix) and preserves or improves average accuracy on the original evaluations. In contrast, continued fine-tuning with \textit{coordinates + description} reduces accuracy on most of those evaluations (Tables~\ref{tab:hard_scaling_original_splits_map} and~\ref{tab:hard_scaling_original_splits_distance}), consistent with catastrophic forgetting \citep{kirkpatrick2017overcoming}.\par}

App.~\ref{sec:continued_scaling} reports the scaling study with 10 and 100 additional maps as well as the full results.



%
%

%

\FloatBarrier
\subsection{Additional analyses}
\label{sec:additional_analyses}

\textbf{Comparison to existing methods.}
Table~\ref{tab:other_baselines_main} 
reports the results of existing approaches on ID maps. %
\par
\begin{table}[!tbp]
    \setlength{\tabcolsep}{3.5pt}
    \centering
    \footnotesize
    \extrarowheight=.9pt
    \caption{\textbf{Other baselines.}
    We report the performance of other baseline models and visual reasoning methods on the ID tasks (results taken from previous work).
    Our best models with \textit{Coord.} input format achieve the best accuracy on ID maps.
    }
    \label{tab:other_baselines_main}
    \begin{adjustbox}{max width=\columnwidth}
    \begin{tabular}{l||cccc|c}
        \multirow{2}{*}{\makecell[l]{Model}} & \multicolumn{5}{c}{\textbf{ID test maps ($d_\infty \leq 5$)}} \\
         & 3x3 & {4x4} & {5x5} & {6x6} & {Avg} \\
        \toprule
        GPT-4o \cite{xu2025visual} & 0.68 & 0.58 & 0.35 & 0.24 & 0.46\\
        Anole Direct \cite{li2025imagine}& 0.83 &0.80& 0.75& 0.75 & 0.78 \\
        Anole CoT \cite{li2025imagine} & 0.94 & 0.72& 0.50 & 0.39 & 0.64\\
        MVoT \cite{li2025imagine} & 0.86 & 0.84 & 0.84 & 0.89 & 0.86\\
        VPFT \cite{xu2025visual} & 0.92 & 0.83 & 0.67 & 0.58 & 0.75 \\
        VPRL \cite{xu2025visual} & 0.98 & 0.96 & 0.91 & 0.82 & 0.92 \\
        \midrule
        \makecell{Coord., Coord. + Descr. CoT} & \textbf{1.00} & 0.95 & \textbf{0.98} & 0.86 & 0.95 \\
        Coord., A* CoT & \textbf{1.00} & \textbf{0.99} & 0.96 & \textbf{0.95} & \textbf{0.98}\\
        \bottomrule
    \end{tabular}
    \end{adjustbox}
\end{table}

\par
\citet{li2025imagine} fine-tune Anole \citep{chern2024anole} on a larger training set than ours for 40 epochs.
Their approach, MVoT, which also includes images of the map in the reasoning traces, results in 86\% accuracy.
\citet{xu2025visual} use the image-only LVM-7B \citep{bai2024sequential}
and report 75.4\% exact match with supervised fine-tuning (VPFT) and
91.6\% with reinforcement learning (VPRL).
\citet{jin2026latentum} introduce LatentUM, which plans from a map image by generating text tokens for each action, followed by
latent visual tokens for the resulting map state.
LatentUM achieves slightly higher ID accuracy than
our A* model but substantially lower accuracy than either of our
coordinate-based models in OOD splits
(Table~\ref{tab:comparison_latentum_short}).
Our models with \textit{coordinates} input significantly outperform all these previous approaches.
Since the models from \citet{xu2025visual} are not available, we could not test their OOD generalization.

\begin{table}[!htbp]
    \centering
    \footnotesize
    \setlength{\tabcolsep}{4pt}
    \extrarowheight=.9pt
    \caption{\textbf{Comparison to LatentUM on map-size generalization.}
    We report average ID and OOD accuracy for LatentUM and two Qwen2.5-VL
    models fine-tuned with coordinate input.
    Table~\ref{tab:comparison_latentum} gives the results by map size.}
    \label{tab:comparison_latentum_short}
    \begin{tabular}{l||ccc}
        Model / CoT format & ID & \makecell{OOD\\random $d_\infty$}
        & \makecell{OOD\\$d_\infty \geq 6$} \\
        \midrule
        LatentUM-Vis-Plan & \textbf{0.99} & 0.27 & 0.04 \\
        \midrule
        Coord. + Descr. & 0.95 & \textbf{0.92} & \textbf{0.83} \\
        A* w/ Coord.   & 0.98 & 0.89 & 0.74 \\
        \bottomrule
    \end{tabular}
\end{table}

\textbf{Comparison to Mirage.}
Mirage \citep{yang2025machine} integrates images into CoT through continuous tokens encoding helper images of a correct path, before giving the text-based solution.
\citet{yang2025machine} report that Mirage Direct, which relies on such continuous reasoning and supervised fine-tuning (SFT), outperforms SFT on the solution only (Direct SFT) and with text-based CoT (CoT SFT), all using \frozenlake maps as images.
Table~\ref{tab:comparison_mirage_short} compares the results reported in \citet{yang2025machine} to our models also trained on \textit{image} input format and no or \textit{descr.} CoTs.

\begin{table}[!tbp]
    \setlength{\tabcolsep}{2.5pt}
    \centering
    \footnotesize
    \extrarowheight=.9pt
    \caption{\textbf{Comparison to Mirage.} We compare the ID performance of models from \citet{yang2025machine} to ours (20 epochs training, \textsuperscript{$\dagger$}~trained for 10 epochs only).
    The continuous reasoning of Mirage does not provide benefit over our model trained without CoT data.
    }
    \label{tab:comparison_mirage_short}
    \begin{tabular}{l|| cccc|c}
        \multirow{2}{*}{\makecell[l]{Model}} & \multicolumn{5}{c}{\textbf{ID test maps ($d_\infty \leq 5$)}} \\
         &  3x3 & {4x4} & {5x5} & {6x6} & {Avg}%
         \\
        \toprule

        Direct SFT \citep{yang2025machine} & 0.88 & 0.81 & 0.73 & 0.47 & 0.72 \\
        CoT SFT \citep{yang2025machine} & 0.68 & 0.53 & 0.35 & 0.31 & 0.47 \\
        Mirage Direct \citep{yang2025machine}  & 0.93 & 0.83 & 0.76 & 0.51 & 0.76 \\
Mirage Direct (retrained)  & 0.91 & 0.82 & \textbf{0.79} & 0.51 & 0.76 \\
Mirage Direct shuffled& 0.92 & 0.85 & 0.82 & 0.50 & 0.77 \\
\midrule
Image, no CoT &  0.91 & 0.85 & 0.76 & 0.59 & 0.78 \\
Image, Descr. CoT \textsuperscript{$\dagger$} & \textbf{0.95} & \textbf{0.87} & 0.76 & \textbf{0.62} & \textbf{0.80} \\
\bottomrule
\end{tabular}
\vspace{2mm}
    
\end{table}

Surprisingly, our model without CoT already outperforms Mirage Direct and Direct SFT, which use the same data: we hypothesize this is due to our better choice of training hyperparameters. %
The model with our \textit{descrip.} CoT attains the best results (80\% average accuracy), while \citet{yang2025machine} report that this version, with their reasoning traces, largely fails (47\%): we conjecture our more concise CoT formulation 
yields this improvement.
Finally, we retrain Mirage Direct in the original setup and with randomly shuffled helper images. Both variants obtain similar performance to
the original Mirage Direct, suggesting that their continuous reasoning approach does not provide benefits on this task.
Table~\ref{tab:comparison_mirage} in appendix shows that the Mirage framework does not improve OOD accuracy.

\textbf{Other analyses.}
App.~\ref{sec:other_analyses} reports response lengths (Table~\ref{tab:response_length_success}) and the small effect of longer training (Fig.~\ref{fig:num-epochs}). Examples of generated reasoning traces are provided in App.~\ref{sec:additional_figures}.

\FloatBarrier
\section{Conclusion}
\label{sec:conc}

\textbf{Discussion.}
Our results confirm the observation of prior works \citep{stechly2024chain, zhao2025chain} that chain-of-thought reasoning, while helping LLMs to generalize to in-distribution test samples, often fails under small distribution shifts.
However, unlike existing works, we demonstrate that significant improvements can be achieved by carefully choosing the format for input and reasoning, and by augmenting the training maps with additional solutions.
In this way, we obtain state-of-the-art results on both in- and out-of-distribution data for the \frozenlake visual planning task, with similar trends for the Qwen3.5 models.
Even these models struggle on maps whose shortest solutions require many movement directions or turns, with their performance recovering after continuing the fine-tuning with directionally complex maps.
Overall, the effectiveness of A*-like CoT, solution-set augmentation, and continued fine-tuning underlines the importance of rich reasoning data, with multiple and diverse behaviors, in supervised fine-tuning of LLMs.

\textbf{Outlook.}
While the exact approaches we used may not be directly applicable to every planning or reasoning task, the rationale behind them and the takeaways from our experiments have the potential to benefit a large class of problems.
The flexibility of our environment, in terms of data formats and distribution shifts, also offers the possibility of constructing more complex OOD tasks and studying how techniques such as latent reasoning and reinforcement learning interact with different data formats.

\FloatBarrier
\section*{Impact Statement}

This paper presents work whose goal is to advance the field of Machine
Learning. There are many potential societal consequences of our work, none
which we feel must be specifically highlighted here.

\balance

\bibliographystyle{icml2026}

\clearpage
\nobalance

\appendix
\newlength{\appendixproseparskip}
\setlength{\appendixproseparskip}{\parskip}
\makeatletter
\setlength{\@dblfptop}{0pt}
\setlength{\@dblfpsep}{16pt}
\setlength{\@dblfpbot}{0pt plus 1fil}
\makeatother

\tcbset{
  gridbox/.style={
    colback=gray!12,
    colframe=teal!70!black,
    boxrule=0.8pt,
    arc=4pt,
    boxsep=6pt,
    left=2pt, right=2pt, top=0pt, bottom=0pt,
    enhanced
  }
}

\begin{table*}[p]
\begin{minipage}{\textwidth}
\normalsize
\begin{multicols}{2}
\setlength{\parskip}{\appendixproseparskip}
\section{Data and Experimental Details} \label{sec:exp_details}

\textbf{Data generation.}
For the OOD maps, we generate the maps randomly using a hole probability of $0.1$. We generated 200 maps per map size (statistics about start-goal distance are reported in Table~\ref{tab:avg_distance}). Player and goal are placed at random, using rejection sampling in cases where we have a constraint on the distance between start and goal. Similarly, to embed the small maps into 10×10 size, we generate a random 10×10 map, remove player and treasure, and then place the small map randomly within the large one. Direction-complexity and turn-complexity maps are generated by sampling and only accepting maps that meet their respective criteria.

\textbf{Construction of reasoning traces.}
To generate descriptive reasoning traces for each step of the solution sequence, we first identify the general direction from the player toward the \treasure, which can include one or two of the four possible directions.
In the 3×3 example, this direction is “UP.” Next, we check whether this direction is blocked by a hole. If it is not, this move is selected; otherwise, we iterate over the remaining directions: first the other direction pointing toward the \treasure (if it exists), and then the one not pointing towards the \treasure. We enforce that the chosen direction in the reasoning trace always corresponds to the next move in the target action sequence of the training point to ensure consistency between reasoning and solution.
The A* format always begins its chain-of-thought by stating its intention to solve the problem with the A* algorithm and writes down its choice of notation before beginning its simulation of the algorithm, as seen in Fig. \ref{fig:reasoning-traces-appendix}.

\textbf{Training setup.}
The statistics of the training data are reported in Table~\ref{tab:train-dist}.
For fine-tuning the models, we use the SFTTrainer from the trl \cite{vonwerra2020trl} package with the hyperparameters stated in Table~\ref{tab:hyperp}. All models were trained on two A100 GPUs.

\end{multicols}
\end{minipage}
\medskip

  \begingroup
  \renewenvironment{table}[1][]{\par\begingroup}{\par\endgroup}%
  \renewenvironment{table*}[1][]{\par\begingroup}{\par\endgroup}%
  \begin{table*}[t]
    \setlength{\tabcolsep}{5.7pt}
    \centering
    \small 
    \footnotesize
    \extrarowheight=.9pt
    \vspace{6mm}
    \caption{\textbf{Average start-goal distance.}
    We report the average distance $d_\infty$ between the start and goal positions for each map size.
    For the OOD maps with $d_\infty \geq 6$ we enforce that both map size and start-goal distance are out-of-distribution compared to the training data.
    }
    \label{tab:avg_distance}
    \begin{tabular}{l||cccc||cccc||cccc}
         & \multicolumn{4}{c||}{\textbf{ID test maps ($d_\infty \leq 5$)}} &\multicolumn{4}{c||}{\textbf{OOD maps (random $d_\infty$)}} & \multicolumn{4}{c}{\textbf{OOD maps ($d_\infty \geq 6$)}}\\
         & 3x3 & 4x4 & 5x5 & 6x6 & 7x7 & 8x8 & 9x9 & 10x10 & 7x7 & 8x8 & 9x9 & 10x10   \\
         \toprule
         Mean $d_\infty$ & 1.50 & 1.79 & 2.16 & 2.75 & 3.27 & 3.66 & 4.01 & 4.47 & 6.00 & 6.35 & 6.72 & 7.00 \\
         \bottomrule
    \end{tabular}
\vspace{6mm}
    
\end{table*}

  \endgroup

\medskip
\noindent\begin{minipage}[t]{\dimexpr(\textwidth-\columnsep)/2\relax}\vspace{0pt}

    \setlength{\tabcolsep}{4.5pt}
    \centering
    \footnotesize
    \extrarowheight=.9pt
    \caption{\textbf{Statistics of training data.} We report the mean $d_\infty$ as well as the mean length of the ground truth solution sequence on the training set.}
    \label{tab:train-dist}
    \begin{tabular}{l|cccc|c}
                             & 3x3 & 4x4 & 5x5 & 6x6 & Total  \\
        \toprule
         Number of samples& 100 & 200 & 300 & 400 & 1000\\
         Mean $d_\infty$ & 1.40 & 1.80& 2.23 & 2.72 & 2.26\\
         Mean solution length & 2.03 & 2.65 & 3.58 & 4.18 & 3.48\\
         \bottomrule
    \end{tabular}

\end{minipage}\hfill\begin{minipage}[t]{\dimexpr(\textwidth-\columnsep)/2\relax}\vspace{0pt}

    \setlength{\tabcolsep}{4.5pt}
    \centering
    
    \footnotesize
    \extrarowheight=.9pt
    \caption{\textbf{SFT hyperparameters.} These values were used for all text-based models. The image models use the same hyperparameters as \citet{yang2025machine}}
    \label{tab:hyperp}
    \begin{tabular}{l|c}
        \textbf{Hyperparameter} & \textbf{Value} \\
        \toprule
        \texttt{epochs} & 10 \\
        \texttt{per\_device\_train\_batch\_size} & 1 \\
        \texttt{gradient\_accumulation\_steps} & 1 \\
        \texttt{warmup\_steps} & 10 \\
        \texttt{learning rate} & 1e-5 \\
        \texttt{weight\_decay} & 0.01 \\
        \texttt{optim} & AdamW \\
        \texttt{bf16} & True \\
        \bottomrule
    \end{tabular}

\end{minipage}

\end{table*}

\begin{table*}[p]
\begin{minipage}{\textwidth}
\normalsize
\begin{multicols}{2}
\setlength{\parskip}{\appendixproseparskip}
\section{Additional experiments}
\label{sec:additional_exp}

\subsection{Additional data formats} \label{sec:additional_formats}

\textbf{Results of additional formats.}
Table~\ref{tab:results_main_all} and Table~\ref{tab:results_embedded_maps_all} report the results of all data formats on generalization to larger maps and larger start-goal distance with fixed map size respectively.
The \textit{table} format yields LLMs with behavior consistent with those using \textit{grid}, yet typically worse results and longer, thus more computationally expensive, answers.
The alphanumeric \textit{coordinates (sheet)}, inspired by spreadsheet conventions, perform similarly to or slightly worse than the numerical coordinates.
The same observation holds for A* reasoning in the \textit{coordinates (sheet)} format.
This confirms that the coordinate format, with the different CoT variants, is relatively robust to small variations in the exact formulation used. 

\end{multicols}
\end{minipage}
\medskip

  \begingroup
  \renewenvironment{table}[1][]{\par\begingroup}{\par\endgroup}%
  \renewenvironment{table*}[1][]{\par\begingroup}{\par\endgroup}%
  \begin{table*}
    \setlength{\tabcolsep}{3.8pt}
    \centering
    \footnotesize
    \extrarowheight=.9pt
    \caption{\textbf{Influence of data format on ID and OOD generalization.} %
    We report accuracy on both ID and OOD map sizes of zero-shot and fine-tuned LLMs using different combination of input and CoT traces (if any) formats.
    For OOD maps, we distinguish test sets where the distance between start and goal position is $d_\infty \geq 6$, i.e. not seen during training.
    Models provided with coordinate input and using coordinate CoT formats (\textit{coordinates/coordinates + description}) lead to the best out-of-distribution performance, while A* formats attain the highest in-distribution performance, with both categories achieving substantial accuracy until 10×10 maps while being trained on up to 6×6 maps.
    }
    \label{tab:results_main_all}
    \begin{tabular}{ll|| cccc|c|| cccc|c|| cccc|c}
        \multirow{2}{*}{\makecell[l]{Input\\ format}} & \multirow{2}{*}{\makecell[l]{CoT\\ format}} & \multicolumn{5}{c||}{\textbf{ID test maps ($d_\infty \leq 5$)}} &\multicolumn{5}{c||}{\textbf{OOD maps (random $d_\infty$)}} & \multicolumn{5}{c}{\textbf{OOD maps ($d_\infty \geq 6$)}}\\
         &  & 3x3 & {4x4} & {5x5} & {6x6} & {Avg} & {7x7} & {8x8} & {9x9} & {10x10} & Avg & {7x7} & {8x8} & {9x9} & {10x10} & {Avg}\\
        \toprule
        
\multirow{3}{*}{Image} & zero-shot & 0.14 & 0.05 & 0.03 & 0.02 & 0.06 & 0.02 & 0.02 & 0.01 & 0.01 & 0.01 & 0.01 & 0.00 & 0.00 & 0.00 & 0.00\\
 & no CoT & 0.89 & 0.84 & 0.72 & 0.52 & 0.74 & 0.51 & 0.37 & 0.23 & 0.12 & 0.31 & 0.02 & 0.00 & 0.00 & 0.00 & 0.01\\
 & Descr. & 0.95 & 0.87 & 0.76 & 0.62 & 0.80 & 0.47 & 0.35 & 0.21 & 0.14 & 0.29 & 0.07 & 0.03 & 0.00 & 0.01 & 0.03\\
 \midrule
 
\multirow{7}{*}{Descr.} & zero-shot & 0.41 & 0.25 & 0.23 & 0.13 & 0.26 & 0.12 & 0.14 & 0.10 & 0.10 & 0.12 & 0.01 & 0.06 & 0.04 & 0.03 & 0.03\\
 & no CoT  & 0.92 & 0.91 & 0.86 & 0.64 & 0.83 & 0.65 & 0.55 & 0.49 & 0.41 & 0.52 & 0.19 & 0.08 & 0.04 & 0.02 & 0.08\\
 & Descr. & 0.94 & 0.92 & 0.90 & 0.67 & 0.86 & 0.67 & 0.63 & 0.55 & 0.46 & 0.58 & 0.07 & 0.02 & 0.01 & 0.01 & 0.02\\
 & Grid & 0.96 & 0.89 & 0.90 & 0.70 & 0.86 & 0.59 & 0.47 & 0.28 & 0.12 & 0.36 & 0.35 & 0.17 & 0.06 & 0.01 & 0.15\\
 & Table & 0.99 & 0.90 & 0.93 & 0.69 & 0.88 & 0.64 & 0.45 & 0.33 & 0.16 & 0.40 & 0.28 & 0.16 & 0.07 & 0.02 & 0.13\\
 & Grid + Descr. & 0.98 & 0.93 & 0.93 & 0.69 & 0.88 & 0.70 & 0.51 & 0.41 & 0.26 & 0.47 & 0.44 & 0.28 & 0.11 & 0.07 & 0.23\\
 &  Table + Descr. & 0.93 & 0.93 & 0.93 & 0.71 & 0.88 & 0.66 & 0.52 & 0.40 & 0.32 & 0.47 & 0.31 & 0.21 & 0.12 & 0.09 & 0.18\\
\midrule

\multirow{5}{*}{Table} & zero-shot & 0.43 & 0.24 & 0.20 & 0.08 & 0.24 & 0.09 & 0.08 & 0.04 & 0.04 & 0.06 & 0.04 & 0.03 & 0.01 & 0.01 & 0.02 \\
 & no CoT  & 0.94 & 0.89 & 0.83 & 0.59 & 0.81 & 0.56 & 0.41 & 0.35 & 0.27 & 0.40 & 0.02 & 0.01 & 0.00 & 0.00 & 0.01\\
 & Descr. & 0.96 & 0.94 & 0.83 & 0.69 & 0.85 & 0.57 & 0.39 & 0.34 & 0.23 & 0.39 & 0.04 & 0.03 & 0.00 & 0.00 & 0.02\\
 & Table & 0.99 & 0.90 & 0.93 & 0.72 & 0.89 & 0.73 & 0.51 & 0.35 & 0.19 & 0.45 & 0.38 & 0.24 & 0.07 & 0.03 & 0.18\\
 & Table + Descr. & 0.97 & 0.93 & 0.95 & 0.78 & 0.91 & 0.81 & 0.57 & 0.41 & 0.27 & 0.51 & 0.48 & 0.32 & 0.12 & 0.06 & 0.25\\
\midrule

\multirow{5}{*}{Grid} & zero-shot & 0.14 & 0.13 & 0.06 & 0.03 & 0.09 & 0.02 & 0.01 & 0.00 & 0.02 & 0.01 & 0.00 & 0.01 & 0.01 & 0.01 & 0.01\\
 & no CoT  & 0.95 & 0.89 & 0.83 & 0.62 & 0.82 & 0.55 & 0.39 & 0.34 & 0.26 & 0.38 & 0.04 & 0.02 & 0.00 & 0.00 & 0.01\\
 & Descr. & 0.94 & 0.91 & 0.86 & 0.62 & 0.83 & 0.53 & 0.41 & 0.35 & 0.25 & 0.39 & 0.10 & 0.07 & 0.01 & 0.01 & 0.05\\
& Grid & 0.98 & 0.92 & 0.97 & 0.71 & 0.90 & 0.74 & 0.51 & 0.39 & 0.24 & 0.47 & 0.43 & 0.22 & 0.07 & 0.04 & 0.19\\
 & Grid + Descr. & 0.99 & 0.93 & 0.92 & 0.79 & 0.91 & 0.83 & 0.68 & 0.49 & 0.35 & 0.59 & 0.65 & 0.55 & 0.23 & 0.20 & 0.41\\

\midrule

\multirow{5}{*}{Coord.} & zero-shot & 0.30 & 0.20 & 0.18 & 0.10 & 0.20 & 0.13 & 0.08 & 0.08 & 0.11 & 0.10 & 0.04 & 0.05 & 0.04 & 0.02 & 0.03 \\
 & no CoT  &  0.94 & 0.89 & 0.92 & 0.72 & 0.87 & 0.78 & 0.60 & 0.55 & 0.47 & 0.60 & 0.21 & 0.06 & 0.02 & 0.01 & 0.07 \\
 & Descr. & 0.98 &  0.96 & 0.96 & 0.75 & 0.91 & 0.72 & 0.63 & 0.55 & 0.49 & 0.59 & 0.14 & 0.13 & 0.04 & 0.04 & 0.09 \\
& Coord. & \textbf{1.00} & 0.93 & 0.94 & 0.84 & 0.93 & 0.93 & 0.88 & 0.88 & 0.83 & 0.88 & 0.83 & 0.80 & 0.68 & 0.70 & 0.75 \\
 & Coord. + Descr. & \textbf{1.00} & 0.95 & \textbf{0.98} & 0.86 & 0.95 & \textbf{0.97} & 0.93 & \textbf{0.90} & \textbf{0.88} & \textbf{0.92} & \textbf{0.89} & \textbf{0.87} & \textbf{0.81} & \textbf{0.77} & \textbf{0.83} \\

\midrule
\multirow{5}{*}{\makecell[l]{Coord.\\ (sheet)}} & zero-shot 
& 0.18 & 0.14 & 0.13 & 0.06 & 0.13 & 0.11 & 0.10 & 0.06 & 0.11 & 0.09 & 0.07 & 0.05 & 0.05 & 0.05 & 0.05 \\

 & no CoT 
& 0.91 & 0.89 & 0.89 & 0.68 & 0.84 & 0.69 & 0.59 & 0.50 & 0.41 & 0.55 & 0.03 & 0.05 & 0.03 & 0.02 & 0.03 \\

 & Descr. 
& 0.95 & 0.92 & 0.90 & 0.73 & 0.88 & 0.71 & 0.59 & 0.49 & 0.43 & 0.55 & 0.09 & 0.05 & 0.01 & 0.01 & 0.04 \\

 & Coord. 
& \textbf{1.00} & 0.94 & 0.97 & 0.85 & 0.94 & 0.94 & 0.93 & 0.87 & 0.80 & 0.88 & 0.82 & 0.81 & 0.68 & 0.62 & 0.73 \\

 & Coord. + Descr. 
& 0.97 & 0.95 & 0.95 & 0.79 & 0.92 & 0.90 & 0.87 & 0.79 & 0.81 & 0.84 & 0.78 & 0.75 & 0.65 & 0.63 & 0.70 \\

\midrule
Coord. & A* w/ coord.
& \textbf{1.00} & \textbf{0.99} & 0.96 & \textbf{0.95} & \textbf{0.98} & 0.96 & 0.91 & 0.88 & 0.82 & 0.89 & 0.87 & 0.76 & 0.73 & 0.62 & 0.74 \\
\midrule

\makecell[l]{Coord.\\ (sheet)} & A* w/ coord. 
& \textbf{1.00} & 0.98 & 0.96 & 0.94 & 0.97 & 0.95 & \textbf{0.94} & 0.84 & 0.78 & 0.87 & 0.86 & 0.77 & 0.63 & 0.59 & 0.71 \\

\bottomrule
\end{tabular}
    
\end{table*}%
  \endgroup

\end{table*}

\begin{table*}[p]

  \begingroup
  \renewenvironment{table}[1][]{\par\begingroup}{\par\endgroup}%
  \renewenvironment{table*}[1][]{\par\begingroup}{\par\endgroup}%
  \begin{table*}[t]
    \setlength{\tabcolsep}{6.3pt}
    \centering
    \footnotesize
    \extrarowheight=.9pt
    \caption{\textbf{Generalization w.r.t. start-end distance ($d_\infty$) with fixed 10×10 map size.}
    We compare different input and CoT format when the training maps have fixed size 10x10 and $d_\infty < 6$, test maps size 10×10 and $d_\infty \in \{2, \ldots, 5\}$ (ID) or $d_\infty \in \{6, \ldots,  9\}$ (OOD).
    Using \textit{coordinates} input with \textit{coordinates}, \textit{coordinates + description}, or \textit{A*} CoT format yields the best results, with numerical coordinates outperforming their sheet-style counterparts; \textit{grid} input combined \textit{grid} or \textit{grid + description} CoT format yield lower yet nontrivial results, while the accuracy of models with no CoT or CoTs in other formats drop close to $0\%$ already at $d_\infty = 6$.
    }
    \label{tab:results_embedded_maps_all}
    \begin{tabular}{ll|| cccc|c|| cccc|c}
        \multirow{2}{*}{\makecell[l]{Input\\ format}} & \multirow{2}{*}{\makecell[l]{CoT\\ format}} & \multicolumn{5}{c||}{\textbf{ID test maps ($d_\infty \leq 5$)}}  & \multicolumn{5}{c}{\textbf{OOD maps ($d_\infty \geq 6$)}}\\
        & & 2 & 3 & 4 & 5 & Avg & 6 & 7 & 8 & 9 & Avg\\ 
        \toprule

\multirow{2}{*}{Image} & no CoT & 0.70 & 0.46 & 0.21 & 0.03 & 0.35 & 0.03 & 0.01 & 0.00 & 0.00 & 0.01\\
 & Descr. & 0.69 & 0.41 & 0.16 & 0.06 & 0.33 & 0.01 & 0.00 & 0.00 & 0.00 & 0.00\\
\midrule

\multirow{4}{*}{Descr.} & no CoT & 0.82 & 0.75 & 0.57 & 0.34 & 0.62 & 0.01 & 0.01 & 0.00 & 0.00 & 0.01\\
 & Descr. & 0.89 & 0.80 & 0.68 & 0.41 & 0.69 & 0.00 & 0.00 & 0.00 & 0.00 & 0.00\\
  & Grid & 0.82 & 0.76 & 0.67 & 0.49 & 0.69 & 0.34 & 0.20 & 0.10 & 0.04 & 0.17\\
 & Grid + Descr. & 0.82 & 0.73 & 0.61 & 0.47 & 0.66 & 0.36 & 0.25 & 0.17 & 0.14 & 0.23 \\
\midrule

\multirow{2}{*}{Table} & no CoT & 0.88 & 0.77 & 0.54 & 0.16 & 0.59 & 0.01 & 0.00 & 0.00 & 0.00 & 0.00\\
 & Descr. & 0.89 & 0.76 & 0.47 & 0.09 & 0.55 & 0.00 & 0.01 & 0.01 & 0.00 & 0.00\\
\midrule

\multirow{4}{*}{Grid} & no CoT & 0.86 & 0.81 & 0.55 & 0.12 & 0.58 & 0.00 & 0.01 & 0.00 & 0.00 & 0.00\\
 & Descr. & 0.84 & 0.78 & 0.45 & 0.10 & 0.54 & 0.00 & 0.01 & 0.00 & 0.00 & 0.00\\
  & Grid & 0.89 & 0.84 & 0.77 & 0.62 & 0.78 & 0.46 & 0.34 & 0.17 & 0.08 & 0.26\\
 & Grid + Descr. & 0.86 & 0.84 & 0.74 & 0.61 & 0.76 & 0.50 & 0.33 & 0.21 & 0.10 & 0.29 \\
 \midrule


\multirow{5}{*}{Coord.} & zero-shot 
& 0.11 & 0.06 & 0.06 & 0.05 & 0.07 & 0.07 & 0.03 & 0.01 & 0.01 & 0.03 \\

 & no CoT 
& 0.90 & 0.83 & 0.69 & 0.43 & 0.71 & 0.01 & 0.00 & 0.00 & 0.00 & 0.00 \\

 & Descr. 
& 0.84 & 0.73 & 0.65 & 0.32 & 0.63 & 0.00 & 0.00 & 0.00 & 0.00 & 0.00 \\

 & Coord. 
& 0.89 & 0.85 & 0.79 & 0.64 & 0.79 & 0.53 & 0.39 & 0.31 & 0.21 & 0.36 \\

 & Coord. + Descr. 
& 0.95 & 0.94 & 0.89 & 0.74 & 0.88 & 0.75 & 0.56 & 0.44 & 0.44 & 0.55 \\

\midrule
\multirow{5}{*}{\makecell[l]{Coord. (sheet)}} & zero-shot 
& 0.15 & 0.11 & 0.05 & 0.06 & 0.09 & 0.04 & 0.04 & 0.03 & 0.04 & 0.04 \\

 & no CoT 
& 0.83 & 0.71 & 0.64 & 0.35 & 0.63 & 0.00 & 0.00 & 0.00 & 0.00 & 0.00 \\

 & Descr. 
& 0.82 & 0.66 & 0.59 & 0.30 & 0.59 & 0.02 & 0.00 & 0.00 & 0.00 & 0.00 \\

 & Coord. 
& 0.97 & 0.94 & 0.92 & 0.85 & 0.92 & 0.77 & 0.61 & 0.41 & 0.40 & 0.54 \\

 & Coord. + Descr. 
& 0.84 & 0.74 & 0.65 & 0.50 & 0.68 & 0.47 & 0.36 & 0.26 & 0.27 & 0.34 \\

\midrule
Coord. & A* w/ coord. 
& \textbf{0.99} & 0.98 & \textbf{0.96} & \textbf{0.89} & \textbf{0.95} & \textbf{0.85} & 0.63 & \textbf{0.56} & \textbf{0.46} & \textbf{0.62} \\
\midrule

\makecell[l]{Coord. (sheet)} & A* w/ coord. 
& \textbf{0.99} & \textbf{0.99} & 0.95 & 0.88 & \textbf{0.95} & 0.77 & \textbf{0.67} & 0.50 & 0.40 & 0.58 \\

\bottomrule

\end{tabular}
    
\end{table*}%
  \endgroup

\end{table*}

\begin{figure*}[p]
\begin{minipage}{\textwidth}
\normalsize
\begin{multicols}{2}
\setlength{\parskip}{\appendixproseparskip}

\textbf{Generalization w.r.t. solution length for other formats.}
In Fig.~\ref{fig:sol-len-all} we report the success rate w.r.t. optimal solution length for different input formats, similar to what shown in Fig.~\ref{fig:sol-len-coord} for \textit{coordinates}, with similar conclusions.
As a further ablation, we re-train the model with \textit{grid} input and \textit{grid + description} CoT removing the examples with $l>10$ from the training set (10 maps in total): this model (the dashed curve in the Fig.~\ref{fig:sol-len-all}) performs similarly to the original one.

\end{multicols}
\end{minipage}
\medskip

    \centering
    \begin{tabular}{cc}
    \includegraphics[width=0.45\linewidth]{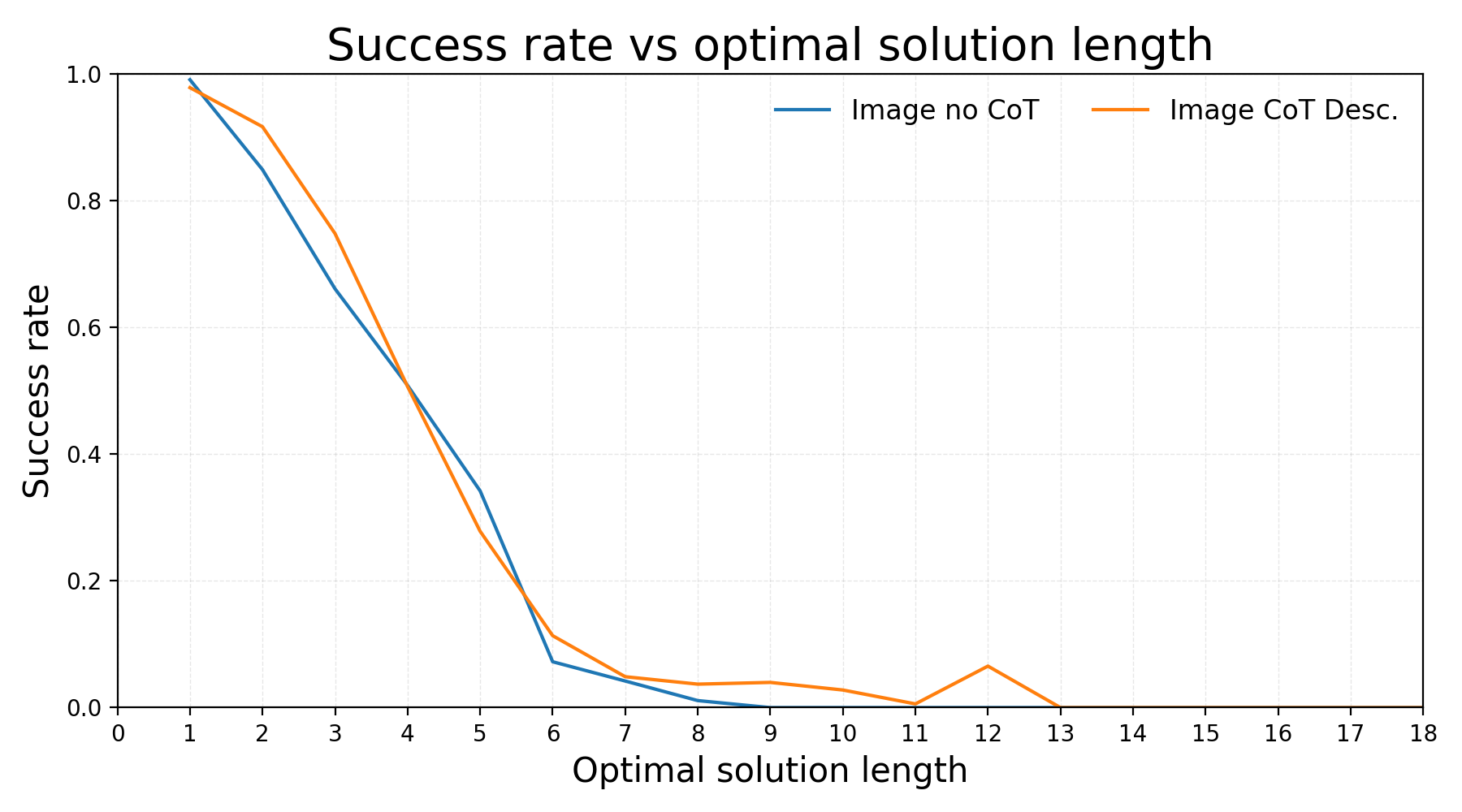}
    &
    \includegraphics[width=0.45\linewidth]{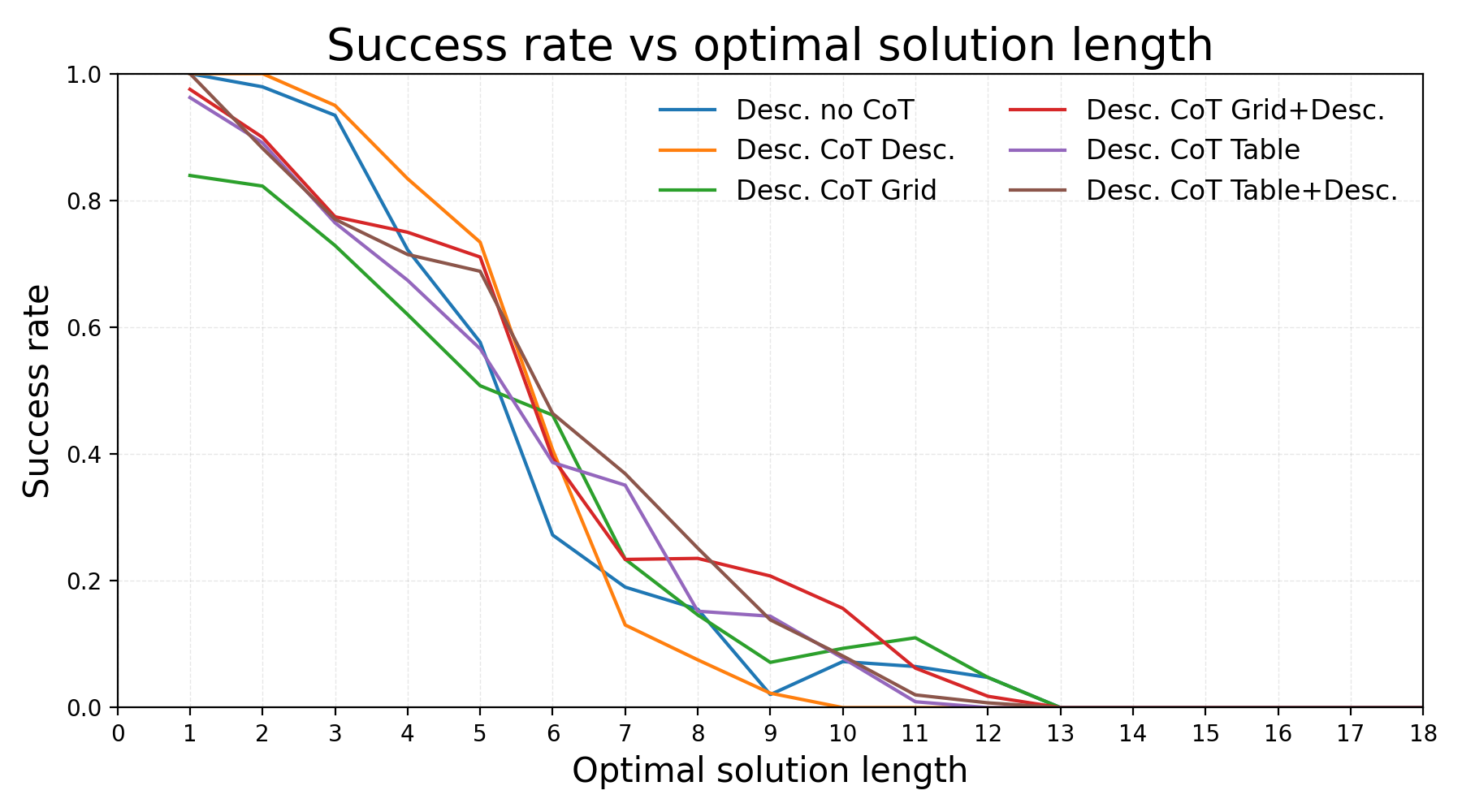}
    \\
    \includegraphics[width=0.45\linewidth]{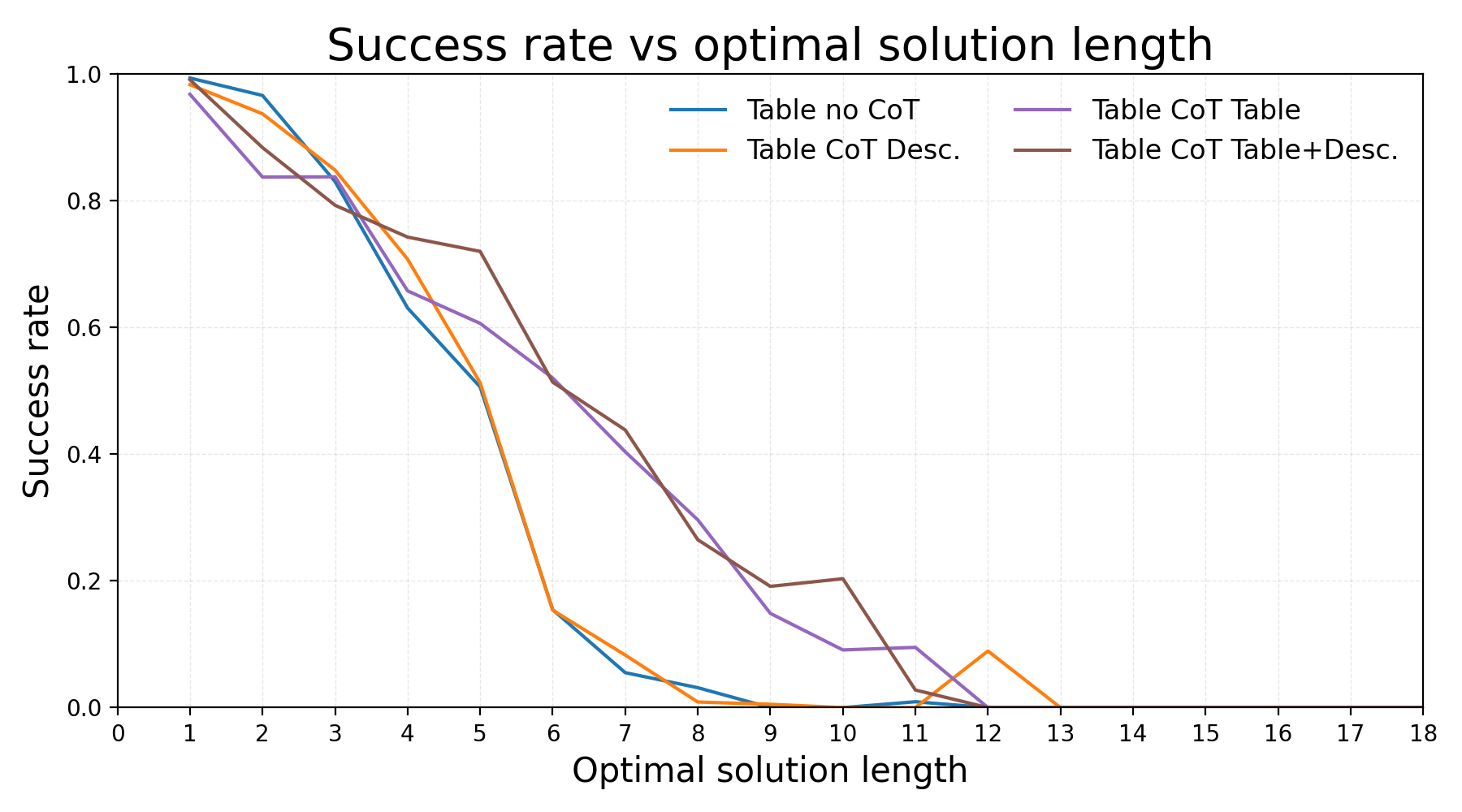}
    &
    \includegraphics[width=0.45\linewidth]{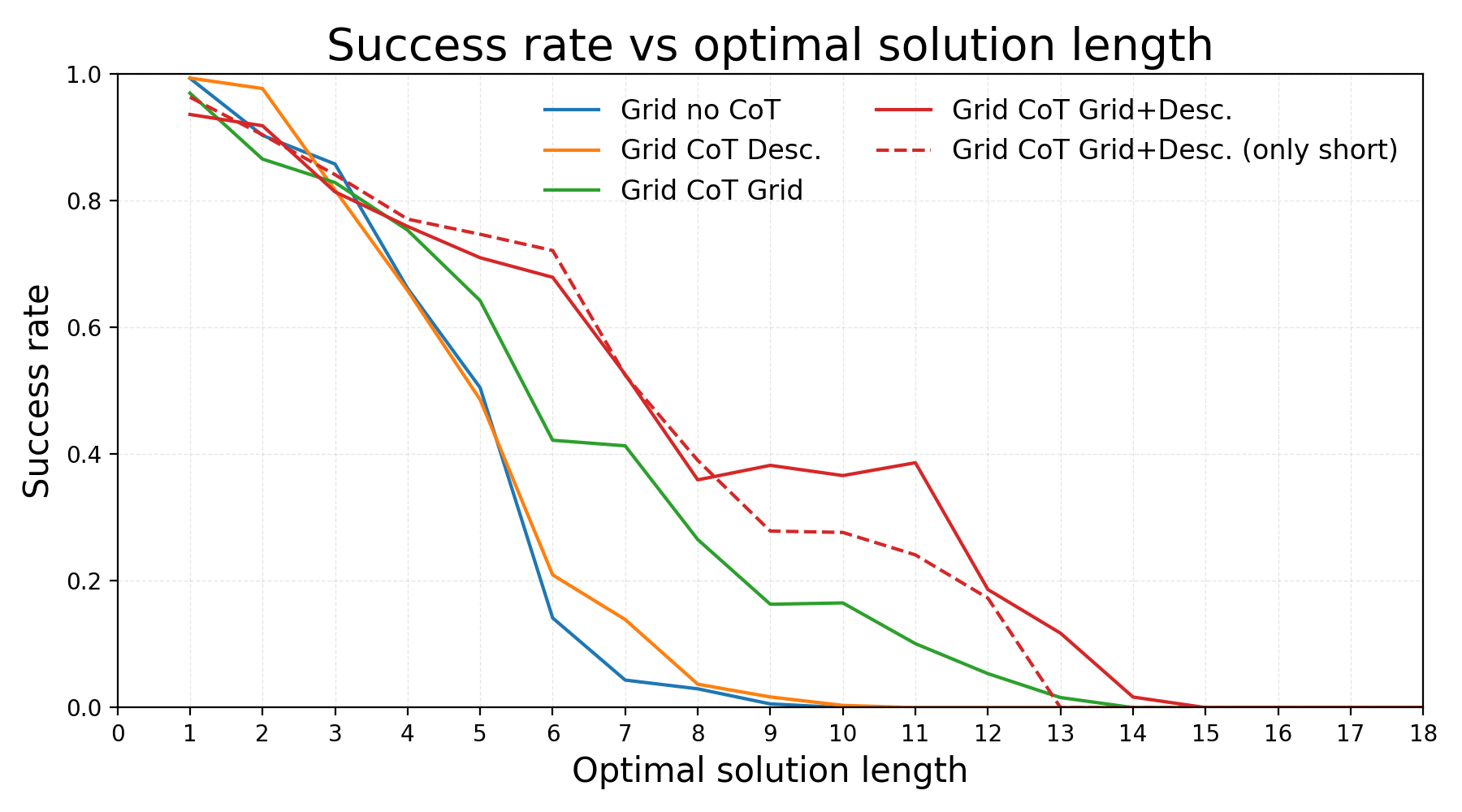}
    \\
    \includegraphics[width=0.45\linewidth]{img/success_rate_vs_solution_length_mapsize_all_coords.png}
    &
    \includegraphics[width=0.45\linewidth]{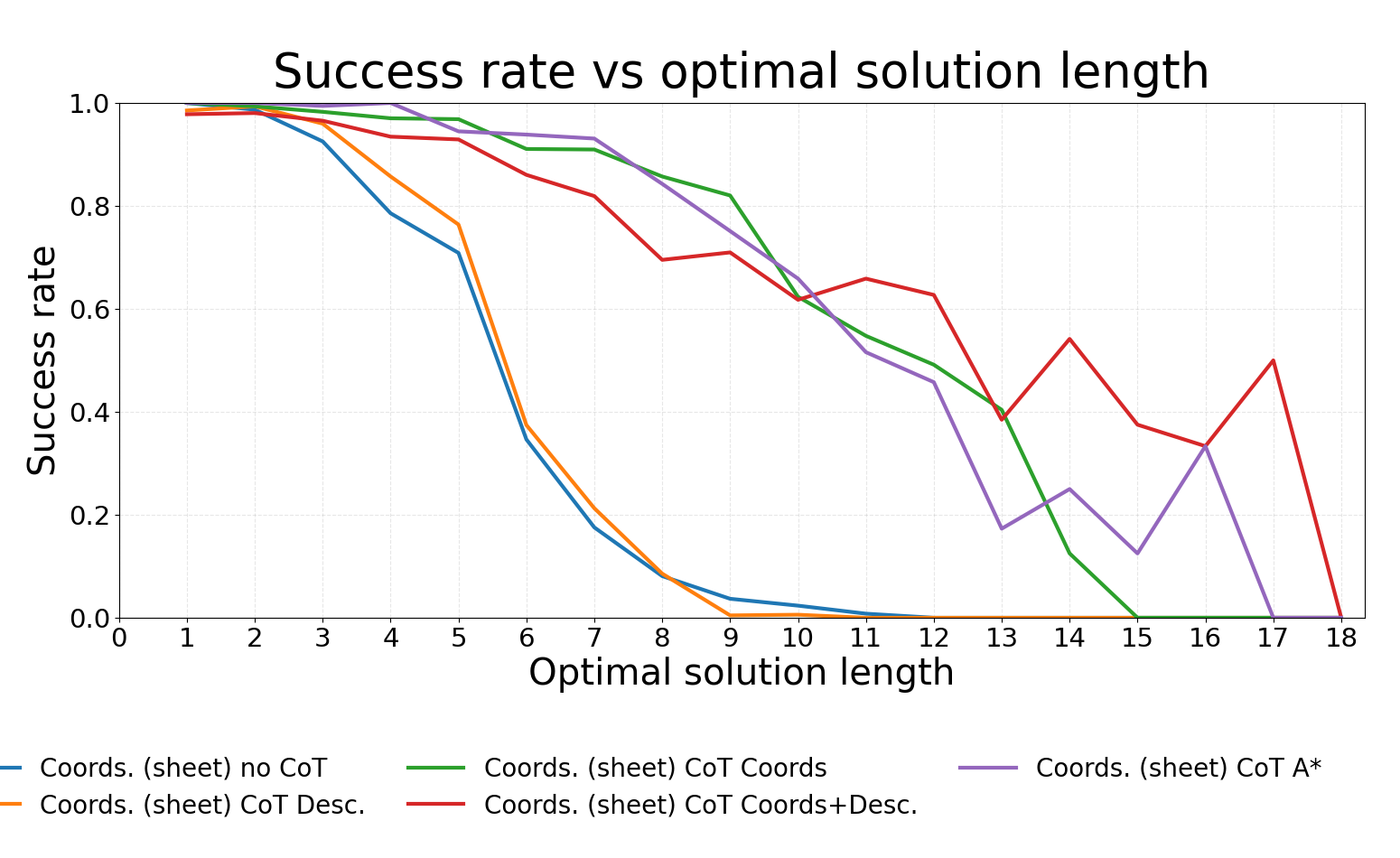}

    \end{tabular}
    
    \caption{
    \textbf{OOD generalization w.r.t. optimal solution length.} We show success rate of models using different input and CoT representation over solution length, complementing the results of Fig.~\ref{fig:sol-len-coord} for the \textit{coordinates} format. 
    }
    \label{fig:sol-len-all}

\end{figure*}

\begin{table*}[p]
\noindent\begin{minipage}[t]{\columnwidth}\vspace{0pt}
\normalsize
\setlength{\parskip}{\appendixproseparskip}

\subsection{Other analyses}
\label{sec:other_analyses}


\textbf{Extended comparison to Mirage on OOD maps.} To complement the results in Sec.~\ref{sec:additional_analyses}, Table~\ref{tab:comparison_mirage} further reports the performance on OOD of the Mirage models \citep{yang2025machine} we retrained (the original ones are not available).
Similar to ID performance, the continuous reasoning approach of Mirage does not provide benefit compared to our simple supervised fine-tuning with \textit{image} inputs, with or without \textit{description} CoT.

\textbf{Detailed results of other methods.}
Table~\ref{tab:other_baselines} reports the results of additional baseline models.
We see that frontier LLMs such as Gemini-1.0-Pro-Vision \cite{team2023gemini}, Claude-3 \cite{claude3} and GPT-4o \cite{hurst2024gpt}, have low zero-shot accuracy (the best is GPT-4o with 46\%) on the ID test maps (results are taken from \cite{wu2024vsp}).
Further, we report the three fine-tuning variants from \citet{li2025imagine}: \textit{Direct}, \textit{CoT}, and \textit{MVoT}. They use Anole-7B \cite{chern2024anole} as base model which can generate interleaved text and images autoregressively. \textit{Direct} corresponds to SFT without reasoning traces and achieves an accuracy of 78\%. For \textit{CoT}, the model is fine-tuned on reasoning traces incorporating coordinates and environment layout described in the text. In contrast to our experiments, they are not able to improve using such traces (accuracy 64\%). \textit{MVoT} also includes images of the grid in the reasoning traces and results in 86\%. Note that these methods were fine-tuned on a larger training set for 40 epochs each.

\end{minipage}\hfill\begin{minipage}[t]{\columnwidth}\vspace{0pt}
\normalsize
\setlength{\parskip}{\appendixproseparskip}

VPFT \citep{xu2025visual} achieves 75.4\% accuracy (exact match metric) with a specialized model, LVM-7B \citep{bai2024sequential}, which only handles images as both input and output.
VPRL \citep{xu2025visual} achieves 91.6\% using two-stage training with reinforcement learning (RL).
This is similar to what we achieve via supervised fine-tuning alone of the general-purpose Qwen2.5-VL-7B-Instruct with \textit{grid/table + description} CoT (91\%), see Table~\ref{tab:results_main}.
Since the models from \citet{xu2025visual} are not available, we could not test their OOD generalization.
\citet{jin2026latentum} fine-tune \mbox{LatentUM}, based on InternVL3.5-4B,
on 18,000 \frozenlake examples. With image input, the model autoregressively generates each action as text tokens, followed by latent visual tokens for the resulting map state. On our tests, LatentUM achieves 99\% ID accuracy but
solves at most 1\% of maps with $d_\infty \geq 6$ at each size from $8\times8$
to $10\times10$ (Table~\ref{tab:comparison_latentum}).

\medskip
\begingroup
    \setlength{\tabcolsep}{3.3pt}
    \centering
    \footnotesize
    \extrarowheight=.9pt
    \captionsetup{type=table}
    \caption{\textbf{Other baselines.}
    We report the performance of other baseline models and visual reasoning methods on the ID tasks.
    The results are taken from the previous works reporting them.
    }
    \label{tab:other_baselines}
    \begin{tabular}{l||cccc|c}
        \multirow{2}{*}{\makecell[l]{Model}} & \multicolumn{5}{c}{\textbf{ID test maps ($d_\infty \leq 5$)}} \\
         &  3x3 & {4x4} & {5x5} & {6x6} & {Avg}
         \\
        \toprule
        Gemini-1.0 \cite{xu2025visual} & 0.31 & 0.26 & 0.15 & 0.06 & 0.20\\
        Claude-3 \cite{xu2025visual} & 0.52 & 0.33 & 0.16 & 0.15 & 0.29\\
        GPT-4o \cite{xu2025visual} & 0.68 & 0.58 & 0.35 & 0.24 & 0.46\\
        \midrule
        Anole Direct \cite{li2025imagine}& 0.83 &0.80& 0.75& 0.75 & 0.78 \\
        Anole CoT \cite{li2025imagine} & 0.94 & 0.72& 0.50 & 0.39 & 0.64\\
        MVoT \cite{li2025imagine} &
        0.86 & 0.84 & 0.84 & 0.89 & 0.86  \\
        \midrule
        VPFT \cite{xu2025visual} & 0.92 & 0.83 & 0.67 & 0.58 & 0.75 \\
        VPRL \cite{xu2025visual} & 0.98 & 0.96 & 0.91 & 0.82 & 0.92 \\

        \bottomrule
    \end{tabular}
\par
\endgroup

\end{minipage}

\par\bigskip
\captionsetup{type=table}
  \begingroup
  \renewenvironment{table}[1][]{\par\begingroup}{\par\endgroup}%
  \renewenvironment{table*}[1][]{\par\begingroup}{\par\endgroup}%
  \begin{table*}
    \setlength{\tabcolsep}{2.7pt}
    \centering
    \footnotesize
    \extrarowheight=.9pt
    \vspace{4mm}
    \caption{\textbf{Comparison to Mirage.} We additionally report the OOD performance of the models in Table~\ref{tab:comparison_mirage_short}.
    The latent reasoning approach of Mirage does not provide better OOD generalization than simple supervised fine-tuning without CoT.
    }
    \label{tab:comparison_mirage}
    \begin{tabular}{l|| cccc|c|| cccc|c|| cccc|c}
        \multirow{2}{*}{\makecell[l]{Model}} & \multicolumn{5}{c||}{\textbf{ID test maps ($d_\infty \leq 5$)}} &\multicolumn{5}{c||}{\textbf{OOD maps (random $d_\infty$)}} & \multicolumn{5}{c}{\textbf{OOD maps ($d_\infty \geq 6$)}}\\
         &  3x3 & {4x4} & {5x5} & {6x6} & {Avg} & {7x7} & {8x8} & {9x9} & {10x10} & Avg & {7x7} & {8x8} & {9x9} & {10x10} & {Avg}\\
        \toprule

        Direct SFT \citep{yang2025machine} & 0.88 & 0.81 & 0.73 & 0.47 & 0.72 & - & - & - & - & - & - & - & - & - & -\\
        CoT SFT \citep{yang2025machine} & 0.68 & 0.53 & 0.35 & 0.31 & 0.47 & - & - & - & - & - & - & - & - & - & - \\
        Mirage Direct \citep{yang2025machine}  & 0.93 & 0.83 & 0.76 & 0.51 & 0.76 & - & - & - & - & - & - & - & - & - & - \\
Mirage Direct (retrained)  & 0.91 & 0.82 & 0.79 & 0.51 & 0.76 & 0.45 & 0.35 & 0.22 & 0.15 & 0.29 & 0.01 & 0.01 & 0.01 & 0.00 & 0.00\\
Mirage Direct shuffled& 0.92 & 0.85 & 0.82 & 0.5 & 0.77 & 0.46 & 0.41 & 0.26 & 0.14 & 0.31 & 0.05 & 0.06 & 0.02 & 0.00 & 0.03\\
\midrule
Image, no CoT (ours) &  0.91 & 0.85 & 0.76 & 0.59 & 0.78 & 0.46 & 0.41 & 0.23 & 0.16 & 0.31 & 0.04 & 0.01 & 0.00 & 0.00 & 0.01\\
Image, Descr. CoT (ours) & 0.95 & 0.87 & 0.76 & 0.62 & 0.80 & 0.47 & 0.35 & 0.21 & 0.14 & 0.29 & 0.07 & 0.03 & 0.00 & 0.01 & 0.03\\
\bottomrule
\end{tabular}
    \vspace{4mm}
\end{table*}
  \endgroup

\bigskip
\captionsetup{type=table}
  \begingroup
  \renewenvironment{table}[1][]{\par\begingroup}{\par\endgroup}%
  \renewenvironment{table*}[1][]{\par\begingroup}{\par\endgroup}%
  \begin{table*}[p]
    \centering
    \footnotesize
    \setlength{\tabcolsep}{2.5pt}
    \extrarowheight=.9pt
    \caption{\textbf{Detailed results for LatentUM.}
    We report accuracy by map size for the models in
    Table~\ref{tab:comparison_latentum_short}.
    }
    \label{tab:comparison_latentum}
    \begin{tabular}{l||cccc|c||cccc|c||cccc|c}
        \multirow{2}{*}{Model / CoT format}
        & \multicolumn{5}{c||}{\textbf{ID test maps ($d_\infty \leq 5$)}}
        & \multicolumn{5}{c||}{\textbf{OOD maps (random $d_\infty$)}}
        & \multicolumn{5}{c}{\textbf{OOD maps ($d_\infty \geq 6$)}} \\
        & 3x3 & 4x4 & 5x5 & 6x6 & Avg
        & 7x7 & 8x8 & 9x9 & 10x10 & Avg
        & 7x7 & 8x8 & 9x9 & 10x10 & Avg \\
        \midrule
        LatentUM-Vis-Plan
        & 0.99 & 1.00 & 1.00 & 0.98 & \textbf{0.99}
        & 0.56 & 0.22 & 0.17 & 0.16 & 0.27
        & 0.13 & 0.01 & 0.01 & 0.01 & 0.04 \\
        \midrule
        Coord. + Descr.
        & 1.00 & 0.95 & 0.98 & 0.86 & 0.95
        & 0.97 & 0.93 & 0.90 & 0.88 & \textbf{0.92}
        & 0.89 & 0.87 & 0.81 & 0.77 & \textbf{0.83} \\
        A* w/ Coord.
        & 1.00 & 0.99 & 0.96 & 0.95 & 0.98
        & 0.96 & 0.91 & 0.88 & 0.82 & 0.89
        & 0.87 & 0.76 & 0.73 & 0.62 & 0.74 \\
        \bottomrule
    \end{tabular}
\end{table*}%
  \endgroup

\end{table*}

\begin{table*}[p]
\begin{minipage}{\textwidth}
\normalsize
\begin{multicols}{2}
\setlength{\parskip}{\appendixproseparskip}

\textbf{Length of reasoning.}
Table~\ref{tab:response_length_success} reports the average token length of responses that lead to correct solutions.
We see no clear relation between response length and accuracy.
In the main OOD map-size setting, the best-performing method is numerical \textit{coordinates + description}, even though its responses are much shorter than A* traces and many \textit{grid}/\textit{table} multi-format CoTs.
A* traces are substantially longer and achieve the strongest performance in the fixed 10$\times$10 map size start-goal distance generalization setting, but they do not dominate in the OOD evaluation with varying map sizes.
Likewise, \textit{sheet-style coordinates} reduce output length compared to numerical coordinate formats, but do not consistently improve accuracy. The two \textit{coordinates} CoT variants perform similarly, while numerical \textit{coordinates + description} is clearly stronger than its sheet-style counterpart. Table~\ref{tab:response_length_training_variants} reports the output lengths for the solution-set augmentation variants. Unsurprisingly, the models trained with suboptimal solutions (\textit{up to 3 sol/map} and \textit{up to 5 sol/map}) produce longer outputs than the variants trained only with optimal solutions.

\end{multicols}
\end{minipage}
\medskip

  \begingroup
  \renewenvironment{table}[1][]{\par\begingroup}{\par\endgroup}%
  \renewenvironment{table*}[1][]{\par\begingroup}{\par\endgroup}%
  \begin{table*}[t]
    \setlength{\tabcolsep}{3.4pt}
    \centering
    \footnotesize
    \extrarowheight=.9pt
    \vspace{4mm}
    \caption{\textbf{Output length.} We report the average length (in tokens) of correct solutions, including reasoning, for the various fine-tuned models at each map size.
    }
    \label{tab:response_length_success}
    \begin{tabular}{ll|| cccc|c|| cccc|c|| cccc|c}
        \multirow{2}{*}{\makecell[l]{Input\\ format}} & \multirow{2}{*}{\makecell[l]{CoT\\ format}} & \multicolumn{5}{c||}{\textbf{ID test maps ($d_\infty \leq 5$)}} &\multicolumn{5}{c||}{\textbf{OOD maps (random $d_\infty$)}} & \multicolumn{5}{c}{\textbf{OOD maps ($d_\infty \geq 6$)}}\\
         &  & 3x3 & {4x4} & {5x5} & {6x6} & {Avg} & {7x7} & {8x8} & {9x9} & {10x10} & Avg & {7x7} & {8x8} & {9x9} & {10x10} & {Avg}\\
        \toprule
        
\multirow{3}{*}{Image} & zero-shot & 210 & 429 & 309 & 363 & 328  & 450 & 590 & 662 & 547 & 562  & 524 & 457 & 774 & 502 & 564 \\
 & no CoT & 9 & 10 & 10 & 11 & 10  & 12 & 11 & 10 & 9 & 11  & 19 & -  & -  & -  & 19 \\
 & Descr. & 52 & 59 & 64 & 86 & 65 & 81 & 74 & 65 & 54 & 69  & 196 & 217 & -  & 162 & 192 \\
 \midrule
 
\multirow{7}{*}{Descr.} & zero-shot & 222 & 264 & 505 & 541 & 383  & 588 & 836 & 832 & 675 & 732  & 848 & 791 & 1010 & 833 & 870\\
 & no CoT  &  9 & 10 & 11 & 12 & 10  & 13 & 13 & 13 & 13 & 13  & 21 & 19 & 18 & 22 & 20  \\
 & Descr. & 52 & 62 & 73 & 87 & 69  & 86 & 90 & 91 & 90 & 89  & 156 & 127 & 122 & 157 & 141 \\
 & Grid & 54 & 98 & 176 & 310 & 160 & 472 & 554 & 666 & 652 & 586 & 850 & 1023 & 1281 & 2116 & 1317 \\
 & Table & 155 & 272 & 497 & 824 & 437  & 1136 & 1307 & 1534 & 1268 & 1311  & 2071 & 2534 & 2967 & 4387 & 2990 \\
 & Grid + Descr. & 97 & 148 & 236 & 369 & 213  & 623 & 621 & 772 & 1069 & 771  & 1000 & 1371 & 1676 & 2413 & 1615 \\
 & Table + Descr. & 188 & 320 & 527 & 848 & 471  & 1213 & 1451 & 1760 & 2205 & 1657  & 2238 & 2613 & 3289 & 4353 & 3123  \\
\midrule

\multirow{5}{*}{Table} & zero-shot &  227 & 384 & 701 & 540 & 463  & 910 & 684 & 951 & 1027 & 893  & 848 & 918 & 886 & 1097 & 937 \\
 & no CoT  & 9 & 10 & 11 & 12 & 10  & 12 & 12 & 11 & 11 & 11  & 20 & 19 & -  & -  & 19 \\
 & Descr. & 52 & 64 & 68 & 91 & 69 & 82 & 80 & 76 & 64 & 75 & 171 & 189 & - & - & 180 \\
 & Table & 152 & 266 & 482 & 810 & 428  & 1242 & 1436 & 1687 & 1701 & 1517  & 2166 & 2652 & 3383 & 3498 & 2925 \\
 & Table + Descr. & 192 & 324 & 544 & 899 & 490  & 1265 & 1633 & 1982 & 2755 & 1909  & 2333 & 2979 & 3813 & 6489 & 3904 \\
\midrule

\multirow{5}{*}{Grid} & zero-shot & 355 & 668 & 1030 & 1086 & 785  & 1209 & 1314 & 1459 & 1567 & 1387  & 1067 & 1421 & 1523 & 1340 & 1338\\
 & no CoT  &  9 & 10 & 11 & 12 & 10  & 12 & 11 & 11 & 11 & 11  & 19 & 18 & -  & -  & 19  \\
 & Descr. & 52 & 61 & 72 & 83 & 67  & 81 & 80 & 79 & 71 & 77  & 161 & 145 & 171 & 150 & 157 \\
& Grid & 53 & 98 & 179 & 300 & 158  & 457 & 614 & 659 & 744 & 619  & 871 & 1081 & 1251 & 1572 & 1194 \\
 & Grid + Descr. & 98 & 151 & 239 & 393 & 220  & 582 & 751 & 882 & 1109 & 831  & 1025 & 1298 & 1632 & 1880 & 1459 \\
\midrule

\multirow{5}{*}{Coord.} & zero-shot & 138 & 170 & 151 & 170 & 157 & 176 & 223 & 275 & 225 & 225 & 297 & 230 & 258 & 238 & 256 \\
 & no CoT & 11 & 12 & 13 & 14 & 12 & 15 & 15 & 15 & 15 & 15 & 22 & 22 & 20 & 21 & 21 \\
 & Descr. & 54 & 63 & 76 & 90 & 71 & 89 & 88 & 93 & 96 & 92 & 160 & 146 & 137 & 173 & 154 \\
 & Coord. & 64 & 86 & 131 & 193 & 118 & 203 & 254 & 302 & 381 & 285 & 357 & 425 & 516 & 598 & 474 \\
 & Coord. + Descr. & 101 & 134 & 184 & 268 & 172 & 287 & 348 & 399 & 506 & 385 & 512 & 581 & 707 & 785 & 646 \\
\midrule

\multirow{5}{*}{\makecell[l]{Coord.\\ (sheet)}} & zero-shot & 134 & 116 & 144 & 137 & 133 & 143 & 143 & 175 & 161 & 156 & 169 & 171 & 182 & 191 & 178 \\
 & no CoT & 11 & 12 & 13 & 14 & 12 & 15 & 15 & 15 & 15 & 15 & 24 & 21 & 23 & 23 & 23 \\
 & Descr. & 52 & 62 & 72 & 88 & 68 & 90 & 89 & 88 & 90 & 89 & 138 & 152 & 150 & 122 & 140 \\
 & Coord. & 52 & 67 & 94 & 136 & 87 & 143 & 174 & 195 & 246 & 190 & 242 & 284 & 332 & 384 & 310 \\
 & Coord. + Descr. & 86 & 111 & 152 & 204 & 138 & 219 & 261 & 294 & 375 & 287 & 390 & 440 & 515 & 576 & 480 \\
\midrule

Coord. & A* w/ coord. & 476 & 589 & 724 & 960 & 687 & 1090 & 1221 & 1296 & 1484 & 1273 & 1856 & 2002 & 2400 & 2525 & 2196 \\
\midrule

\makecell[l]{Coord.\\ (sheet)} & \makecell[l]{A* w/ coord.\\ (sheet)} & 392 & 478 & 548 & 716 & 534 & 801 & 899 & 931 & 1055 & 922 & 1364 & 1418 & 1703 & 1794 & 1570 \\

\bottomrule
\end{tabular}
    
\end{table*} 
  \endgroup

\end{table*}

\begin{table*}[p]

  \begingroup
  \renewenvironment{table}[1][]{\par\begingroup}{\par\endgroup}%
  \renewenvironment{table*}[1][]{\par\begingroup}{\par\endgroup}%
  \begin{table*}[t]
    \setlength{\tabcolsep}{3.8pt}
    \centering
    \footnotesize
    \extrarowheight=.9pt
    \caption{\textbf{Output length for solution-set augmentation variants.}
    We report the average length, in tokens, of correct solutions for \textit{coordinates + description} models fine-tuned on the same 1000 maps but with different numbers of valid solutions per map.
    }
    \label{tab:response_length_training_variants}
    \begin{tabular}{l|| cccc|c|| cccc|c|| cccc|c}
        \multirow{2}{*}{Training data variant}
        & \multicolumn{5}{c||}{\textbf{ID test maps ($d_\infty \leq 5$)}}
        & \multicolumn{5}{c||}{\textbf{OOD maps (random $d_\infty$)}}
        & \multicolumn{5}{c}{\textbf{OOD maps ($d_\infty \geq 6$)}}\\
        & 3x3 & 4x4 & 5x5 & 6x6 & Avg
        & 7x7 & 8x8 & 9x9 & 10x10 & Avg
        & 7x7 & 8x8 & 9x9 & 10x10 & Avg\\
        \toprule
        1 sol/map (default)
        & 101 & 134 & 184 & 268 & 172
        & 287 & 348 & 399 & 506 & 385
        & 512 & 581 & 707 & 785 & 646 \\

        up to 3 sol/map
        & 148 & 183 & 238 & 308 & 219
        & 316 & 393 & 442 & 554 & 426
        & 509 & 604 & 735 & 817 & 666 \\

        up to 5 sol/map
        & 157 & 181 & 249 & 320 & 227
        & 336 & 407 & 475 & 567 & 446
        & 524 & 614 & 733 & 822 & 673 \\

        all optimal solutions
        & 101 & 131 & 180 & 267 & 170
        & 288 & 355 & 419 & 525 & 397
        & 504 & 589 & 712 & 650 & 614 \\
        \bottomrule
    \end{tabular}
\end{table*}
  \endgroup

\end{table*}

\begin{figure*}[p]
\begin{minipage}{\textwidth}
\normalsize
\begin{multicols}{2}
\setlength{\parskip}{\appendixproseparskip}

\textbf{Effect of number of training epochs.}
To test the effect of longer training, we fine-tune LLMs with various data formats and with or without CoT for 20 and 30 epochs (as default we use 10 epochs).
The detailed results are shown in Fig.~\ref{fig:num-epochs}.
While there are minimal improvements after 20 epochs on the ID test set and after 30 epochs on the OOD test set, the overall performance and model rankings remain largely unchanged.

\textbf{Details about training data variants.} In the previous experiments we have used a fixed training set of 1000 examples (see App.~\ref{sec:exp_details}), where for each map a single solution is shown to the model (with the exception of the A* CoT models which take a different approach).
However, the solution path is not generally unique.
Then, we test the effect of augmenting the training set with additional solutions for the same set of 1k maps.
In particular we consider three configurations: \textit{(i)} we sample up to three solutions for each map, if possible, which are up to four steps longer than the optimal one (total of training samples increases to 2761 when training on 3x3 to 6x6 maps, 2914 for the embedded 10x10 maps), \textit{(ii)} we sample up to five solutions for each map as above (4173 and 4656 training samples), \textit{(iii)} we use all optimal solutions, i.e. with shortest path, for each map (2756 and 2750 training samples).
We use \textit{coordinates} input and \textit{coordinates + description} CoT for these three training sets, as in Sec.~\ref{sec:exp_augmentation}.
We keep 10 epochs regardless of the number of training samples, therefore the number of training steps effectively increases.

\end{multicols}
\end{minipage}
\medskip

    \centering
    \vspace{4mm}
    \includegraphics[width=\dimexpr(\textwidth-\columnsep)/2\relax]{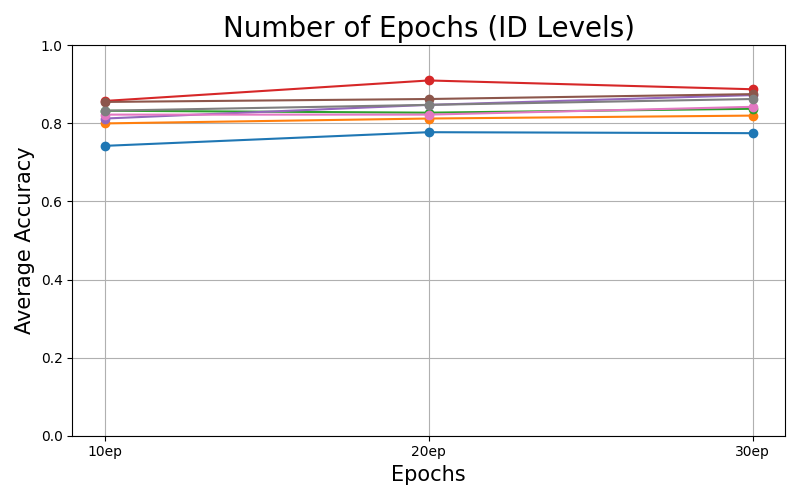}
    \includegraphics[width=\dimexpr(\textwidth-\columnsep)/2\relax]{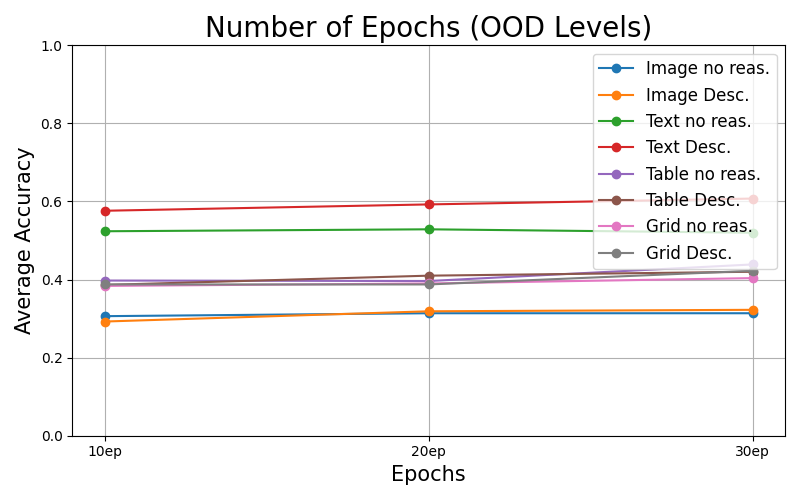}
    \caption{\textbf{Number of epochs.} We extended the fine-tuning to 20 and 30 epochs for some of the models and evaluated them on the ID and the OOD (random $d_\infty$) test maps. However, the overall performance and model ranking remain largely unchanged.
    }
    \label{fig:num-epochs}

\end{figure*}

\begin{table*}[p]
\begin{minipage}{\textwidth}
\normalsize
\begin{multicols}{2}
\setlength{\parskip}{\appendixproseparskip}
\subsection{Qwen3.5: detailed map-size and distance results}
\label{sec:qwen35_full_results}
Tables~\ref{tab:qwen35_map_full} and~\ref{tab:qwen35_distance_full} give the full map-size and distance results summarized in Table~\ref{tab:qwen35_map}.

\end{multicols}
\end{minipage}
\medskip

  \begingroup
  \renewenvironment{table}[1][]{\par\begingroup}{\par\endgroup}%
  \renewenvironment{table*}[1][]{\par\begingroup}{\par\endgroup}%
  \begin{table*}[!t]
\setlength{\tabcolsep}{4pt}
    \centering
    \footnotesize
    \extrarowheight=.9pt

\caption{\textbf{Qwen3.5 map-size results with coordinate inputs.}}
\label{tab:qwen35_map_full}
\begin{adjustbox}{max width=\textwidth}
\begin{tabular}{cl||cccc|c||cccc|c||cccc|c}
\multirow{2}{*}{Model} & \multirow{2}{*}{CoT format} & \multicolumn{5}{c||}{\textbf{ID test maps ($d_\infty \leq 5$)}} & \multicolumn{5}{c||}{\textbf{OOD maps (random $d_\infty$)}} & \multicolumn{5}{c}{\textbf{OOD maps ($d_\infty \geq 6$)}}\\
& & 3x3 & 4x4 & 5x5 & 6x6 & Avg & 7x7 & 8x8 & 9x9 & 10x10 & Avg & 7x7 & 8x8 & 9x9 & 10x10 & Avg\\
\toprule
\multirow{7}{*}{0.8B}
& zero-shot & 0.68 & 0.54 & 0.42 & 0.33 & 0.49 & 0.33 & 0.34 & 0.25 & 0.25 & 0.29 & 0.21 & 0.15 & 0.14 & 0.09 & 0.15\\
& Coord. & 0.98 & 0.92 & 0.95 & 0.81 & 0.92 & 0.90 & 0.83 & 0.81 & 0.68 & 0.80 & 0.67 & 0.58 & 0.41 & 0.36 & 0.50\\
& Coord. + Descr. & 0.98 & 0.96 & \textbf{0.99} & 0.91 & 0.96 & 0.96 & 0.95 & 0.88 & 0.78 & 0.89 & 0.78 & 0.77 & 0.56 & 0.52 & 0.66\\
& A* w/ Coord. & \textbf{1.00} & \textbf{1.00} & \textbf{0.99} & \textbf{0.95} & \textbf{0.99} & 0.97 & 0.94 & 0.86 & 0.83 & 0.90 & \textbf{0.90} & \textbf{0.81} & 0.66 & \textbf{0.62} & \textbf{0.75}\\
& $K=3$ aug. & 0.99 & 0.99 & \textbf{0.99} & 0.90 & 0.97 & \textbf{0.99} & \textbf{0.96} & 0.87 & \textbf{0.84} & \textbf{0.91} & 0.86 & 0.76 & 0.63 & 0.52 & 0.69\\
& $K=5$ aug. & \textbf{1.00} & 0.98 & 0.98 & 0.94 & 0.98 & 0.98 & 0.92 & 0.88 & \textbf{0.84} & 0.90 & 0.84 & 0.71 & 0.63 & 0.49 & 0.67\\
& all-opt aug. & 0.97 & 0.96 & 0.98 & 0.91 & 0.96 & 0.96 & 0.92 & \textbf{0.90} & 0.81 & 0.90 & 0.80 & 0.78 & \textbf{0.67} & 0.55 & 0.70\\
\midrule
\multirow{7}{*}{2B}
& zero-shot & \textbf{1.00} & \textbf{0.99} & 0.97 & 0.97 & 0.98 & 0.94 & 0.94 & 0.86 & 0.84 & 0.89 & 0.85 & 0.84 & 0.75 & 0.68 & 0.78\\
& Coord. & \textbf{1.00} & 0.96 & 0.97 & 0.87 & 0.95 & 0.96 & 0.88 & 0.87 & 0.81 & 0.88 & 0.76 & 0.65 & 0.56 & 0.49 & 0.61\\
& Coord. + Descr. & \textbf{1.00} & 0.97 & 0.98 & 0.91 & 0.97 & 0.95 & 0.91 & 0.85 & 0.79 & 0.87 & 0.62 & 0.56 & 0.43 & 0.42 & 0.50\\
& A* w/ Coord. & \textbf{1.00} & \textbf{0.99} & \textbf{0.99} & \textbf{0.98} & \textbf{0.99} & \textbf{1.00} & 0.96 & \textbf{0.93} & 0.90 & \textbf{0.95} & \textbf{0.92} & 0.87 & \textbf{0.80} & 0.73 & 0.83\\
& $K=3$ aug. & \textbf{1.00} & 0.96 & 0.98 & 0.94 & 0.97 & 0.99 & \textbf{0.98} & \textbf{0.93} & \textbf{0.91} & \textbf{0.95} & \textbf{0.92} & \textbf{0.91} & 0.77 & \textbf{0.75} & \textbf{0.84}\\
& $K=5$ aug. & \textbf{1.00} & \textbf{0.99} & \textbf{0.99} & 0.96 & \textbf{0.99} & 0.99 & 0.97 & 0.92 & 0.90 & 0.94 & 0.91 & 0.83 & 0.78 & 0.71 & 0.80\\
& all-opt aug. & 0.99 & 0.94 & 0.99 & 0.90 & 0.96 & 0.97 & 0.95 & 0.90 & 0.89 & 0.93 & 0.89 & 0.83 & 0.78 & 0.67 & 0.79\\
\midrule
\multirow{7}{*}{4B}
& zero-shot & 0.59 & 0.41 & 0.50 & 0.37 & 0.47 & 0.37 & 0.37 & 0.29 & 0.31 & 0.33 & 0.35 & 0.27 & 0.24 & 0.17 & 0.26\\
& Coord. & \textbf{1.00} & 0.99 & 0.97 & 0.91 & 0.97 & 0.98 & 0.95 & \textbf{0.97} & \textbf{0.95} & 0.96 & 0.91 & 0.94 & 0.87 & 0.89 & 0.90\\
& Coord. + Descr. & \textbf{1.00} & 0.98 & 0.98 & 0.91 & 0.97 & 0.98 & 0.97 & 0.96 & 0.92 & 0.95 & 0.91 & 0.93 & 0.85 & 0.81 & 0.87\\
& A* w/ Coord. & \textbf{1.00} & \textbf{1.00} & \textbf{1.00} & \textbf{0.98} & \textbf{1.00} & \textbf{1.00} & \textbf{0.98} & 0.96 & 0.94 & \textbf{0.97} & 0.95 & 0.90 & 0.85 & 0.83 & 0.88\\
& $K=3$ aug. & \textbf{1.00} & 0.96 & \textbf{1.00} & 0.93 & 0.97 & 0.98 & \textbf{0.98} & 0.96 & \textbf{0.95} & \textbf{0.97} & \textbf{0.96} & \textbf{0.97} & \textbf{0.89} & \textbf{0.91} & \textbf{0.93}\\
& $K=5$ aug. & \textbf{1.00} & 0.99 & 0.99 & 0.95 & 0.98 & 0.98 & \textbf{0.98} & \textbf{0.97} & 0.94 & 0.96 & 0.94 & 0.90 & 0.85 & 0.84 & 0.88\\
& all-opt aug. & \textbf{1.00} & 0.97 & 0.99 & 0.94 & 0.98 & 0.99 & \textbf{0.98} & 0.96 & 0.93 & 0.96 & 0.94 & \textbf{0.97} & \textbf{0.89} & 0.87 & 0.92\\
\bottomrule
\end{tabular}
\end{adjustbox}
\end{table*}
  \endgroup

\end{table*}

\begin{table*}[p]

  \begingroup
  \renewenvironment{table}[1][]{\par\begingroup}{\par\endgroup}%
  \renewenvironment{table*}[1][]{\par\begingroup}{\par\endgroup}%
  \begin{table*}[!tbp]
\setlength{\tabcolsep}{6pt}
    \centering
    \footnotesize
    \extrarowheight=.9pt
\caption{\textbf{Qwen3.5 start-goal distance results with \textit{coordinates} input on fixed $10\times10$ maps.}}
\label{tab:qwen35_distance_full}
\begin{tabular}{cl||cccc|c||cccc|c}
\multirow{2}{*}{Model} & \multirow{2}{*}{CoT format} & \multicolumn{5}{c||}{\textbf{ID test maps ($d_\infty \leq 5$)}} & \multicolumn{5}{c}{\textbf{OOD maps ($d_\infty \geq 6$)}}\\
& & 2 & 3 & 4 & 5 & Avg & 6 & 7 & 8 & 9 & Avg\\
\toprule
\multirow{7}{*}{0.8B}
& zero-shot & 0.34 & 0.29 & 0.25 & 0.12 & 0.25 & 0.18 & 0.14 & 0.11 & 0.05 & 0.12\\
& Coord. & 0.92 & 0.90 & 0.84 & 0.62 & 0.82 & 0.48 & 0.24 & 0.06 & 0.02 & 0.20\\
& Coord. + Descr. & 0.96 & 0.94 & 0.89 & 0.74 & 0.88 & 0.47 & 0.23 & 0.08 & 0.04 & 0.20\\
& A* w/ Coord. & \textbf{0.99} & 0.99 & 0.97 & 0.91 & 0.96 & 0.78 & 0.56 & 0.32 & 0.21 & 0.47\\
& $K=3$ aug. & 0.98 & \textbf{1.00} & 0.97 & 0.90 & 0.96 & 0.85 & \textbf{0.67} & 0.37 & \textbf{0.22} & 0.52\\
& $K=5$ aug. & \textbf{0.99} & \textbf{1.00} & \textbf{0.99} & \textbf{0.95} & \textbf{0.98} & \textbf{0.89} & 0.66 & 0.34 & 0.19 & 0.52\\
& all-opt aug. & \textbf{0.99} & \textbf{1.00} & 0.97 & 0.91 & 0.97 & 0.87 & 0.66 & \textbf{0.40} & 0.18 & \textbf{0.53}\\
\midrule
\multirow{7}{*}{2B}
& zero-shot & 0.97 & 0.94 & 0.92 & 0.84 & 0.92 & 0.72 & 0.71 & 0.56 & 0.58 & 0.64\\
& Coord. & 0.96 & 0.95 & 0.88 & 0.73 & 0.88 & 0.61 & 0.28 & 0.13 & 0.04 & 0.26\\
& Coord. + Descr. & 0.98 & 0.98 & 0.94 & 0.87 & 0.94 & 0.72 & 0.42 & 0.22 & 0.12 & 0.37\\
& A* w/ Coord. & \textbf{1.00} & \textbf{1.00} & \textbf{0.99} & 0.92 & \textbf{0.98} & 0.91 & \textbf{0.82} & \textbf{0.73} & \textbf{0.60} & \textbf{0.76}\\
& $K=3$ aug. & 0.99 & 0.99 & \textbf{0.99} & 0.92 & 0.97 & 0.84 & 0.67 & 0.50 & 0.37 & 0.59\\
& $K=5$ aug. & 0.99 & 0.99 & \textbf{0.99} & \textbf{0.95} & \textbf{0.98} & \textbf{0.93} & 0.71 & 0.42 & 0.23 & 0.57\\
& all-opt aug. & 0.98 & 0.99 & 0.97 & 0.89 & 0.96 & \textbf{0.93} & 0.72 & 0.47 & 0.26 & 0.59\\
\midrule
\multirow{7}{*}{4B}
& zero-shot & 0.40 & 0.34 & 0.30 & 0.25 & 0.32 & 0.24 & 0.12 & 0.19 & 0.17 & 0.18\\
& Coord. & 0.97 & 0.97 & 0.97 & 0.90 & 0.95 & 0.87 & 0.81 & 0.78 & 0.71 & 0.79\\
& Coord. + Descr. & 0.98 & 0.98 & 0.97 & 0.87 & 0.95 & 0.90 & 0.76 & 0.56 & 0.49 & 0.68\\
& A* w/ Coord. & \textbf{1.00} & \textbf{1.00} & \textbf{0.99} & \textbf{0.97} & \textbf{0.99} & 0.97 & \textbf{0.92} & \textbf{0.85} & \textbf{0.81} & \textbf{0.89}\\
& $K=3$ aug. & 0.99 & \textbf{1.00} & 0.98 & 0.94 & 0.98 & 0.95 & 0.83 & 0.79 & 0.63 & 0.80\\
& $K=5$ aug. & 0.99 & \textbf{1.00} & 0.98 & 0.95 & 0.98 & \textbf{0.98} & 0.90 & 0.79 & 0.56 & 0.80\\
& all-opt aug. & 0.98 & 0.99 & 0.97 & 0.91 & 0.96 & 0.95 & 0.83 & 0.61 & 0.47 & 0.71\\
\bottomrule
\end{tabular}
\end{table*}
  \endgroup

\end{table*}

\begin{table*}[p]
\begin{minipage}{\textwidth}
\normalsize
\begin{multicols}{2}
\setlength{\parskip}{\appendixproseparskip}
\subsection{Structural generalization: detailed results}
\label{sec:structural_full_results}
Tables~\ref{tab:direction_complexity_map_full}--\ref{tab:turn_complexity_distance_full} provide the per-size and per-distance results corresponding to the averages in Tables~\ref{tab:direction_complexity_map} and~\ref{tab:turn_complexity_map}. A dash indicates an unavailable configuration; the reported averages exclude these entries.

\end{multicols}
\end{minipage}
\medskip

  \begingroup
  \renewenvironment{table}[1][]{\par\begingroup}{\par\endgroup}%
  \renewenvironment{table*}[1][]{\par\begingroup}{\par\endgroup}%
  \begin{table*}[!tbp]
\setlength{\tabcolsep}{4pt}
    \centering
    \footnotesize
    \extrarowheight=.9pt
    \vspace{4mm}
\caption{\textbf{Results on directionally complex maps by map size.} All models use \textit{coordinates} input. ``Dirs.'' is the minimum number of directions among shortest paths. ID and OOD refer to map size and start-goal distance.}
\label{tab:direction_complexity_map_full}
\begin{tabular}{ll||cccc|c||cccc|c||cccc|c}
\multirow{2}{*}{Dirs.} & \multirow{2}{*}{CoT format} & \multicolumn{5}{c||}{ID maps} & \multicolumn{5}{c||}{OOD maps, random $d_\infty$} & \multicolumn{5}{c}{OOD maps, $d_\infty\geq6$}\\
& & 3x3 & 4x4 & 5x5 & 6x6 & Avg & 7x7 & 8x8 & 9x9 & 10x10 & Avg & 7x7 & 8x8 & 9x9 & 10x10 & Avg\\
\toprule
\multirow{5}{*}{3}
& Coord. + Descr. & 0.80 & 0.68 & 0.53 & 0.43 & 0.61 & 0.33 & 0.30 & 0.19 & 0.06 & 0.22 & 0.30 & 0.20 & 0.08 & 0.04 & 0.15\\
& A* w/ coord. & \textbf{0.99} & \textbf{0.92} & \textbf{0.89} & \textbf{0.72} & \textbf{0.88} & \textbf{0.62} & \textbf{0.54} & 0.43 & \textbf{0.34} & \textbf{0.48} & \textbf{0.56} & 0.38 & 0.27 & \textbf{0.21} & \textbf{0.35}\\
& $K=3$ aug. & 0.93 & 0.87 & 0.73 & 0.52 & 0.76 & 0.50 & 0.44 & 0.34 & 0.21 & 0.37 & 0.43 & 0.38 & \textbf{0.31} & 0.16 & 0.32\\
& $K=5$ aug. & 0.91 & 0.85 & 0.82 & 0.65 & 0.81 & 0.55 & 0.44 & \textbf{0.45} & 0.28 & 0.43 & 0.49 & \textbf{0.45} & 0.26 & \textbf{0.21} & \textbf{0.35}\\
& all-opt aug. & 0.60 & 0.65 & 0.57 & 0.50 & 0.58 & 0.40 & 0.33 & 0.21 & 0.10 & 0.26 & 0.35 & 0.31 & 0.16 & 0.10 & 0.23\\
\midrule
\multirow{5}{*}{4}
& Coord. + Descr. & -- & 0.25 & 0.08 & 0.09 & 0.14 & 0.02 & 0.04 & 0.01 & 0.02 & 0.02 & 0.00 & 0.00 & 0.01 & 0.00 & 0.00\\
& A* w/ coord. & -- & \textbf{0.66} & \textbf{0.49} & \textbf{0.42} & \textbf{0.52} & \textbf{0.25} & \textbf{0.22} & \textbf{0.13} & \textbf{0.10} & \textbf{0.17} & \textbf{0.04} & 0.06 & \textbf{0.06} & \textbf{0.02} & \textbf{0.04}\\
& $K=3$ aug. & -- & 0.53 & 0.34 & 0.20 & 0.36 & 0.14 & 0.10 & 0.05 & 0.03 & 0.08 & 0.02 & 0.02 & 0.01 & 0.01 & 0.01\\
& $K=5$ aug. & -- & 0.60 & 0.43 & 0.32 & 0.45 & 0.23 & 0.15 & 0.09 & 0.06 & 0.13 & 0.01 & \textbf{0.07} & 0.03 & 0.01 & 0.03\\
& all-opt aug. & -- & 0.16 & 0.18 & 0.06 & 0.13 & 0.08 & 0.03 & 0.01 & 0.01 & 0.03 & 0.00 & 0.00 & 0.01 & 0.00 & 0.00\\
\bottomrule
\end{tabular}
\end{table*}%
  \endgroup

\end{table*}

\begin{table*}[p]

  \begingroup
  \renewenvironment{table}[1][]{\par\begingroup}{\par\endgroup}%
  \renewenvironment{table*}[1][]{\par\begingroup}{\par\endgroup}%
  \begin{table*}[!tbp]
\setlength{\tabcolsep}{4.5pt}
    \centering
    \footnotesize
    \extrarowheight=.9pt
\caption{\textbf{Results on directionally complex maps by start-goal distance on fixed $10\times10$ maps.}}
\label{tab:direction_complexity_distance_full}
\begin{tabular}{ll||cccc|c||cccc|c}
\multirow{2}{*}{Dirs.} & \multirow{2}{*}{CoT format} & \multicolumn{5}{c||}{ID distances} & \multicolumn{5}{c}{OOD distances}\\
& & 2 & 3 & 4 & 5 & Avg & 6 & 7 & 8 & 9 & Avg\\
\toprule
\multirow{5}{*}{3}
& Coord. + Descr. & 0.32 & 0.22 & 0.20 & 0.07 & 0.20 & 0.07 & 0.03 & 0.02 & 0.01 & 0.03\\
& A* w/ coord. & \textbf{0.85} & \textbf{0.62} & \textbf{0.52} & \textbf{0.40} & \textbf{0.60} & 0.24 & 0.22 & 0.16 & 0.09 & 0.18\\
& $K=3$ aug. & 0.53 & 0.39 & 0.41 & 0.23 & 0.39 & 0.24 & 0.18 & 0.10 & 0.12 & 0.16\\
& $K=5$ aug. & 0.61 & 0.57 & 0.51 & 0.33 & 0.50 & \textbf{0.36} & \textbf{0.31} & \textbf{0.24} & \textbf{0.19} & \textbf{0.27}\\
& all-opt aug. & 0.46 & 0.38 & 0.33 & 0.21 & 0.34 & 0.24 & 0.16 & 0.10 & 0.11 & 0.15\\
\midrule
\multirow{5}{*}{4}
& Coord. + Descr. & 0.01 & 0.00 & 0.01 & 0.00 & 0.00 & 0.00 & 0.00 & 0.00 & \textbf{0.00} & 0.00\\
& A* w/ coord. & \textbf{0.20} & \textbf{0.14} & \textbf{0.09} & \textbf{0.06} & \textbf{0.12} & \textbf{0.03} & 0.00 & \textbf{0.01} & \textbf{0.00} & \textbf{0.01}\\
& $K=3$ aug. & 0.10 & 0.05 & 0.02 & 0.02 & 0.04 & 0.00 & 0.01 & 0.00 & \textbf{0.00} & 0.00\\
& $K=5$ aug. & 0.09 & 0.07 & 0.05 & 0.02 & 0.06 & 0.01 & \textbf{0.02} & \textbf{0.01} & \textbf{0.00} & \textbf{0.01}\\
& all-opt aug. & 0.04 & 0.02 & 0.01 & 0.01 & 0.02 & 0.00 & 0.00 & 0.00 & \textbf{0.00} & 0.00\\
\bottomrule
\end{tabular}
\end{table*}%
  \endgroup

\end{table*}

\begin{table*}[p]

  \begingroup
  \renewenvironment{table}[1][]{\par\begingroup}{\par\endgroup}%
  \renewenvironment{table*}[1][]{\par\begingroup}{\par\endgroup}%
  \begin{table*}[!tbp]
\setlength{\tabcolsep}{4.5pt}
    \centering
    \footnotesize
    \extrarowheight=.9pt
\caption{\textbf{Results by number of turns and map size.} All models use coordinate inputs. ``Turns'' is the minimum number of direction changes among shortest paths.}
\label{tab:turn_complexity_map_full}
\begin{adjustbox}{max width=\textwidth}
\begin{tabular}{ll||cccc|c||cccc|c||cccc|c}
\multirow{2}{*}{Turns} & \multirow{2}{*}{CoT format} & \multicolumn{5}{c||}{ID maps} & \multicolumn{5}{c||}{OOD maps, random $d_\infty$} & \multicolumn{5}{c}{OOD maps, $d_\infty\geq6$}\\
& & 3x3 & 4x4 & 5x5 & 6x6 & Avg & 7x7 & 8x8 & 9x9 & 10x10 & Avg & 7x7 & 8x8 & 9x9 & 10x10 & Avg\\
\toprule
\multirow{3}{*}{5}
& Coord. + Descr. & -- & 0.05 & 0.09 & 0.04 & 0.06 & 0.06 & 0.01 & 0.02 & 0.00 & 0.02 & 0.06 & 0.05 & 0.03 & 0.01 & 0.04\\
& A* w/ coord. & -- & \textbf{0.65} & \textbf{0.49} & \textbf{0.23} & \textbf{0.46} & \textbf{0.22} & 0.10 & 0.08 & 0.03 & 0.11 & \textbf{0.19} & 0.15 & \textbf{0.12} & \textbf{0.06} & \textbf{0.13}\\
& $K=3$ aug. & -- & 0.26 & 0.32 & 0.14 & 0.24 & 0.15 & \textbf{0.13} & \textbf{0.13} & \textbf{0.07} & \textbf{0.12} & \textbf{0.19} & \textbf{0.17} & 0.10 & \textbf{0.06} & \textbf{0.13}\\
\midrule
\multirow{3}{*}{6}
& Coord. + Descr. & -- & -- & 0.03 & 0.01 & 0.02 & 0.02 & 0.01 & 0.01 & 0.00 & 0.01 & 0.04 & 0.01 & 0.01 & 0.00 & 0.02\\
& A* w/ coord. & -- & -- & \textbf{0.32} & \textbf{0.14} & \textbf{0.23} & \textbf{0.14} & 0.04 & 0.03 & \textbf{0.03} & 0.06 & \textbf{0.17} & 0.07 & 0.03 & 0.03 & \textbf{0.08}\\
& $K=3$ aug. & -- & -- & 0.15 & 0.12 & 0.14 & 0.10 & \textbf{0.10} & \textbf{0.04} & 0.02 & \textbf{0.07} & 0.15 & \textbf{0.08} & \textbf{0.04} & \textbf{0.05} & \textbf{0.08}\\
\midrule
\multirow{3}{*}{7}
& Coord. + Descr. & -- & -- & 0.00 & 0.02 & 0.01 & 0.02 & 0.00 & 0.00 & 0.00 & 0.01 & 0.01 & 0.01 & 0.00 & 0.00 & 0.01\\
& A* w/ coord. & -- & -- & \textbf{0.27} & \textbf{0.12} & \textbf{0.20} & \textbf{0.05} & 0.02 & \textbf{0.02} & 0.00 & 0.02 & \textbf{0.11} & 0.04 & 0.01 & 0.01 & 0.04\\
& $K=3$ aug. & -- & -- & 0.04 & 0.10 & 0.07 & \textbf{0.05} & \textbf{0.03} & 0.01 & \textbf{0.01} & \textbf{0.03} & 0.10 & \textbf{0.05} & \textbf{0.02} & \textbf{0.02} & \textbf{0.05}\\
\midrule
\multirow{3}{*}{8}
& Coord. + Descr. & -- & -- & 0.00 & 0.00 & 0.00 & 0.00 & 0.00 & 0.00 & 0.00 & 0.00 & 0.00 & 0.00 & 0.00 & 0.00 & 0.00\\
& A* w/ coord. & -- & -- & \textbf{0.18} & 0.01 & \textbf{0.10} & \textbf{0.01} & \textbf{0.01} & \textbf{0.01} & \textbf{0.01} & \textbf{0.01} & 0.04 & \textbf{0.02} & \textbf{0.02} & \textbf{0.00} & \textbf{0.02}\\
& $K=3$ aug. & -- & -- & 0.07 & \textbf{0.03} & 0.05 & 0.00 & \textbf{0.01} & \textbf{0.01} & \textbf{0.01} & \textbf{0.01} & \textbf{0.06} & \textbf{0.02} & 0.01 & \textbf{0.00} & \textbf{0.02}\\
\bottomrule
\end{tabular}
\end{adjustbox}
\end{table*}%
  \endgroup

\end{table*}

\begin{table*}[p]

  \begingroup
  \renewenvironment{table}[1][]{\par\begingroup}{\par\endgroup}%
  \renewenvironment{table*}[1][]{\par\begingroup}{\par\endgroup}%
  \begin{table*}[!tbp]
\setlength{\tabcolsep}{6pt}
    \centering
    \footnotesize
    \extrarowheight=.9pt
\caption{\textbf{Results by number of turns and start-goal distance on fixed $10\times10$ maps.}}
\label{tab:turn_complexity_distance_full}
\begin{tabular}{ll||cccc|c||cccc|c}
\multirow{2}{*}{Turns} & \multirow{2}{*}{CoT format} & \multicolumn{5}{c||}{ID distances} & \multicolumn{5}{c}{OOD distances}\\
& & 2 & 3 & 4 & 5 & Avg & 6 & 7 & 8 & 9 & Avg\\
\toprule
\multirow{3}{*}{5}
& Coord. + Descr. & 0.00 & 0.00 & 0.00 & 0.00 & 0.00 & 0.00 & 0.00 & 0.00 & 0.00 & 0.00\\
& A* w/ coord. & \textbf{0.11} & \textbf{0.06} & \textbf{0.11} & \textbf{0.10} & \textbf{0.10} & \textbf{0.14} & \textbf{0.08} & \textbf{0.06} & \textbf{0.08} & \textbf{0.09}\\
& $K=3$ aug. & 0.03 & 0.02 & 0.04 & 0.08 & 0.04 & 0.07 & 0.06 & 0.05 & 0.04 & 0.06\\
\midrule
\multirow{3}{*}{6}
& Coord. + Descr. & 0.00 & 0.00 & 0.00 & 0.00 & 0.00 & 0.00 & 0.00 & 0.00 & 0.00 & 0.00\\
& A* w/ coord. & \textbf{0.01} & \textbf{0.03} & \textbf{0.07} & \textbf{0.05} & \textbf{0.04} & 0.03 & \textbf{0.03} & \textbf{0.04} & \textbf{0.03} & \textbf{0.03}\\
& $K=3$ aug. & 0.00 & 0.01 & 0.03 & \textbf{0.05} & 0.02 & \textbf{0.05} & 0.02 & 0.02 & 0.00 & 0.02\\
\midrule
\multirow{3}{*}{7}
& Coord. + Descr. & 0.00 & 0.00 & 0.00 & 0.00 & 0.00 & 0.00 & 0.00 & 0.00 & 0.00 & 0.00\\
& A* w/ coord. & \textbf{0.01} & \textbf{0.03} & \textbf{0.01} & \textbf{0.03} & \textbf{0.02} & \textbf{0.03} & \textbf{0.01} & 0.01 & \textbf{0.00} & \textbf{0.01}\\
& $K=3$ aug. & 0.00 & 0.00 & \textbf{0.01} & 0.01 & 0.01 & 0.00 & \textbf{0.01} & \textbf{0.02} & \textbf{0.00} & \textbf{0.01}\\
\midrule
\multirow{3}{*}{8}
& Coord. + Descr. & \textbf{0.00} & \textbf{0.00} & \textbf{0.00} & \textbf{0.00} & \textbf{0.00} & \textbf{0.00} & 0.00 & 0.00 & 0.00 & 0.00\\
& A* w/ coord. & \textbf{0.00} & \textbf{0.00} & \textbf{0.00} & \textbf{0.00} & \textbf{0.00} & \textbf{0.00} & \textbf{0.01} & \textbf{0.01} & 0.00 & \textbf{0.01}\\
& $K=3$ aug. & \textbf{0.00} & \textbf{0.00} & \textbf{0.00} & \textbf{0.00} & \textbf{0.00} & \textbf{0.00} & \textbf{0.01} & 0.00 & \textbf{0.01} & \textbf{0.01}\\
\bottomrule
\end{tabular}
\end{table*}%
  \endgroup

\end{table*}

\begin{figure*}[p]
\begin{minipage}{\textwidth}
\normalsize
\begin{multicols}{2}
\setlength{\parskip}{\appendixproseparskip}
\subsection{Continued fine-tuning: data scaling and detailed results}
\label{sec:continued_scaling}
We also evaluate fine-tuning with 10 or 100 additional three- or four-direction maps (Tables~\ref{tab:hard_scaling_map_full}--\ref{tab:hard_scaling_coorddescr_distance_full}).
With A*, accuracy generally improves with the number of additional examples. In contrast, 10 additional examples do not consistently improve the original model.
\end{multicols}
\end{minipage}
\medskip

    \centering
    \includegraphics[width=\dimexpr(\textwidth-\columnsep)/2\relax]{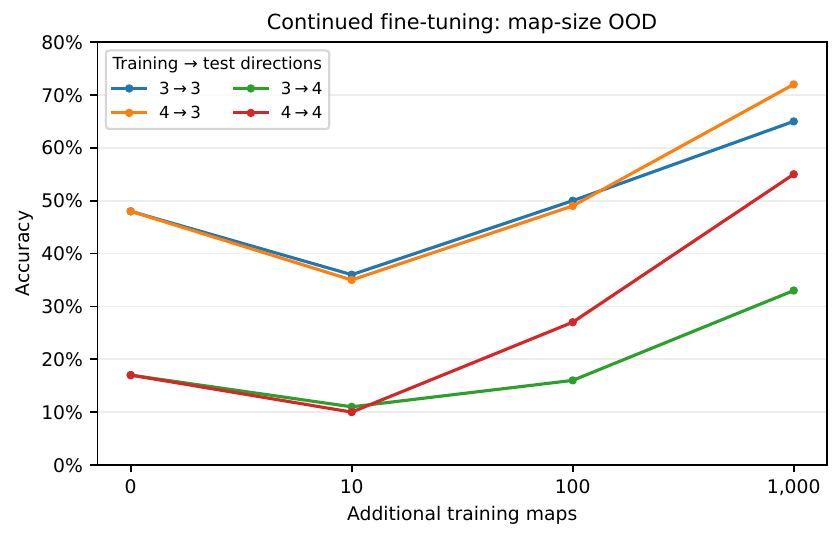}
    \includegraphics[width=\dimexpr(\textwidth-\columnsep)/2\relax]{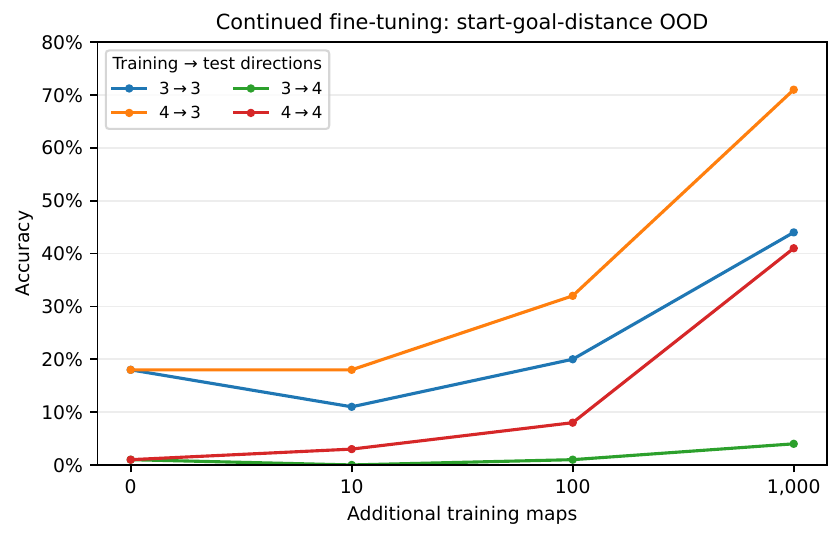}
    \caption{\textbf{A* continued fine-tuning on directionally complex maps.} OOD accuracy after adding 10, 100, or 1,000 three- or four-direction maps; zero denotes the base checkpoint.}
    \label{fig:hard-scaling}

\end{figure*}

\begin{table*}[p]

  \begingroup
  \renewenvironment{table}[1][]{\par\begingroup}{\par\endgroup}%
  \renewenvironment{table*}[1][]{\par\begingroup}{\par\endgroup}%
  \begin{table*}[!tbp]
\setlength{\tabcolsep}{4.7pt}
    \centering
    \footnotesize
    \extrarowheight=.9pt
\caption{\textbf{A* continued fine-tuning on directionally complex maps: results by map size.} Training and test directions give the direction counts of the additional training and evaluation maps. ``Original'' is the checkpoint before continued fine-tuning.}
\label{tab:hard_scaling_map_full}
\begin{adjustbox}{max width=\textwidth}
\begin{tabular}{ccc||cccc|c||cccc|c||cccc|c}
\multirow{2}{*}{\makecell{Train\\dirs.}} & \multirow{2}{*}{\makecell{Test\\dirs.}} & \multirow{2}{*}{Maps} & \multicolumn{5}{c||}{\textbf{ID test maps ($d_\infty \leq 5$)}} & \multicolumn{5}{c||}{\textbf{OOD maps (random $d_\infty$)}} & \multicolumn{5}{c}{\textbf{OOD maps ($d_\infty \geq 6$)}}\\
& & & 3x3 & 4x4 & 5x5 & 6x6 & Avg & 7x7 & 8x8 & 9x9 & 10x10 & Avg & 7x7 & 8x8 & 9x9 & 10x10 & Avg\\
\toprule
\multirow{2}{*}{orig.} & 3 & 0 & 0.99 & 0.92 & 0.89 & 0.72 & 0.88 & 0.62 & 0.54 & 0.43 & 0.34 & 0.48 & 0.56 & 0.38 & 0.27 & 0.21 & 0.35\\
\cmidrule{2-18}
& 4 & 0 & -- & 0.66 & 0.49 & 0.42 & 0.52 & 0.25 & 0.22 & 0.13 & 0.10 & 0.17 & 0.04 & 0.06 & 0.06 & 0.02 & 0.04\\
\midrule
\multirow{6}{*}{3}
& \multirow{3}{*}{3} & 10 & 0.92 & 0.84 & 0.86 & 0.57 & 0.80 & 0.56 & 0.42 & 0.31 & 0.16 & 0.36 & 0.34 & 0.31 & 0.17 & 0.13 & 0.24\\
& & 100 & 0.99 & 0.94 & 0.91 & 0.77 & 0.90 & 0.72 & 0.57 & 0.41 & 0.29 & 0.50 & 0.47 & 0.37 & 0.21 & 0.17 & 0.30\\
& & 1000 & \textbf{1.00} & 0.97 & 0.97 & \textbf{0.91} & 0.96 & 0.89 & 0.72 & 0.59 & 0.42 & 0.65 & 0.70 & 0.57 & 0.35 & 0.28 & 0.47\\
\cmidrule{2-18}
& \multirow{3}{*}{4} & 10 & -- & 0.73 & 0.55 & 0.33 & 0.54 & 0.18 & 0.15 & 0.07 & 0.07 & 0.11 & 0.02 & 0.07 & 0.02 & 0.02 & 0.03\\
& & 100 & -- & 0.78 & 0.54 & 0.42 & 0.58 & 0.22 & 0.22 & 0.11 & 0.08 & 0.16 & 0.03 & 0.02 & 0.07 & 0.02 & 0.03\\
& & 1000 & -- & 0.99 & 0.87 & 0.67 & 0.84 & 0.48 & 0.43 & 0.23 & 0.18 & 0.33 & 0.08 & 0.15 & 0.11 & 0.09 & 0.10\\
\midrule
\multirow{6}{*}{4}
& \multirow{3}{*}{3} & 10 & 0.91 & 0.82 & 0.78 & 0.62 & 0.78 & 0.54 & 0.41 & 0.27 & 0.19 & 0.35 & 0.41 & 0.27 & 0.17 & 0.08 & 0.23\\
& & 100 & 0.99 & 0.90 & 0.87 & 0.72 & 0.87 & 0.70 & 0.57 & 0.44 & 0.28 & 0.49 & 0.57 & 0.43 & 0.33 & 0.20 & 0.38\\
& & 1000 & \textbf{1.00} & \textbf{1.00} & \textbf{0.98} & \textbf{0.91} & \textbf{0.97} & \textbf{0.91} & \textbf{0.82} & \textbf{0.66} & \textbf{0.51} & \textbf{0.72} & \textbf{0.79} & \textbf{0.67} & \textbf{0.54} & \textbf{0.41} & \textbf{0.60}\\
\cmidrule{2-18}
& \multirow{3}{*}{4} & 10 & -- & 0.60 & 0.47 & 0.33 & 0.47 & 0.18 & 0.13 & 0.05 & 0.05 & 0.10 & 0.01 & 0.04 & 0.03 & 0.03 & 0.03\\
& & 100 & -- & 0.95 & 0.80 & 0.61 & 0.79 & 0.42 & 0.34 & 0.19 & 0.15 & 0.27 & 0.12 & 0.09 & 0.08 & 0.05 & 0.08\\
& & 1000 & -- & \textbf{1.00} & \textbf{1.00} & \textbf{0.88} & \textbf{0.96} & \textbf{0.69} & \textbf{0.67} & \textbf{0.49} & \textbf{0.36} & \textbf{0.55} & \textbf{0.47} & \textbf{0.39} & \textbf{0.23} & \textbf{0.18} & \textbf{0.32}\\
\bottomrule
\end{tabular}
\end{adjustbox}
\end{table*}%
  \endgroup

\end{table*}

\begin{table*}[p]

  \begingroup
  \renewenvironment{table}[1][]{\par\begingroup}{\par\endgroup}%
  \renewenvironment{table*}[1][]{\par\begingroup}{\par\endgroup}%
  \begin{table*}[!tbp]
\setlength{\tabcolsep}{6pt}
    \centering
    \footnotesize
    \extrarowheight=.9pt
\caption{\textbf{A* continued fine-tuning on directionally complex maps: results by start-goal distance on fixed $10\times10$ maps.}}
\label{tab:hard_scaling_distance_full}
\begin{tabular}{ccc||cccc|c||cccc|c}
\multirow{2}{*}{\makecell{Train\\dirs.}} & \multirow{2}{*}{\makecell{Test\\dirs.}} & \multirow{2}{*}{Maps} & \multicolumn{5}{c||}{\textbf{ID test maps ($d_\infty \leq 5$)}} & \multicolumn{5}{c}{\textbf{OOD maps ($d_\infty \geq 6$)}}\\
& & & 2 & 3 & 4 & 5 & Avg & 6 & 7 & 8 & 9 & Avg\\
\toprule
\multirow{2}{*}{orig.} & 3 & 0 & 0.85 & 0.62 & 0.52 & 0.40 & 0.60 & 0.24 & 0.22 & 0.16 & 0.09 & 0.18\\
\cmidrule{2-13}
& 4 & 0 & 0.20 & 0.14 & 0.09 & 0.06 & 0.12 & 0.03 & 0.00 & 0.01 & 0.00 & 0.01\\
\midrule
\multirow{6}{*}{3}
& \multirow{3}{*}{3} & 10 & 0.87 & 0.56 & 0.49 & 0.31 & 0.56 & 0.18 & 0.16 & 0.08 & 0.02 & 0.11\\
& & 100 & 0.91 & 0.66 & 0.59 & 0.36 & 0.63 & 0.31 & 0.26 & 0.16 & 0.06 & 0.20\\
& & 1000 & \textbf{1.00} & 0.88 & 0.83 & 0.70 & 0.85 & 0.60 & 0.55 & 0.36 & 0.25 & 0.44\\
\cmidrule{2-13}
& \multirow{3}{*}{4} & 10 & 0.17 & 0.09 & 0.06 & 0.04 & 0.09 & 0.01 & 0.01 & 0.00 & 0.00 & 0.00\\
& & 100 & 0.30 & 0.22 & 0.09 & 0.05 & 0.16 & 0.03 & 0.02 & 0.02 & 0.00 & 0.01\\
& & 1000 & 0.50 & 0.36 & 0.22 & 0.14 & 0.30 & 0.12 & 0.04 & 0.01 & 0.01 & 0.04\\
\midrule
\multirow{6}{*}{4}
& \multirow{3}{*}{3} & 10 & 0.73 & 0.59 & 0.43 & 0.32 & 0.52 & 0.24 & 0.27 & 0.13 & 0.10 & 0.18\\
& & 100 & 0.92 & 0.70 & 0.58 & 0.55 & 0.68 & 0.42 & 0.36 & 0.28 & 0.24 & 0.32\\
& & 1000 & 0.95 & \textbf{0.93} & \textbf{0.90} & \textbf{0.84} & \textbf{0.91} & \textbf{0.82} & \textbf{0.74} & \textbf{0.67} & \textbf{0.63} & \textbf{0.71}\\
\cmidrule{2-13}
& \multirow{3}{*}{4} & 10 & 0.22 & 0.16 & 0.10 & 0.07 & 0.14 & 0.05 & 0.03 & 0.02 & 0.01 & 0.03\\
& & 100 & 0.44 & 0.28 & 0.22 & 0.21 & 0.29 & 0.15 & 0.08 & 0.05 & 0.03 & 0.08\\
& & 1000 & \textbf{0.86} & \textbf{0.76} & \textbf{0.68} & \textbf{0.60} & \textbf{0.72} & \textbf{0.50} & \textbf{0.48} & \textbf{0.39} & \textbf{0.27} & \textbf{0.41}\\
\bottomrule
\end{tabular}
\end{table*}%
  \endgroup

\end{table*}

\begin{table*}[p]

  \begingroup
  \renewenvironment{table}[1][]{\par\begingroup}{\par\endgroup}%
  \renewenvironment{table*}[1][]{\par\begingroup}{\par\endgroup}%
  \begin{table*}[!tbp]
\setlength{\tabcolsep}{4.7pt}
    \centering
    \footnotesize
    \extrarowheight=.9pt
\caption{\textbf{Coord. + Descr. continued fine-tuning on directionally complex maps: results by map size.}}
\label{tab:hard_scaling_coorddescr_map_full}
\begin{adjustbox}{max width=\textwidth}
\begin{tabular}{ccc||cccc|c||cccc|c||cccc|c}
\multirow{2}{*}{\makecell{Train\\dirs.}} & \multirow{2}{*}{\makecell{Test\\dirs.}} & \multirow{2}{*}{Maps} & \multicolumn{5}{c||}{\textbf{ID test maps ($d_\infty \leq 5$)}} & \multicolumn{5}{c||}{\textbf{OOD maps (random $d_\infty$)}} & \multicolumn{5}{c}{\textbf{OOD maps ($d_\infty \geq 6$)}}\\
& & & 3x3 & 4x4 & 5x5 & 6x6 & Avg & 7x7 & 8x8 & 9x9 & 10x10 & Avg & 7x7 & 8x8 & 9x9 & 10x10 & Avg\\
\toprule
\multirow{2}{*}{orig.} & 3 & 0 & 0.80 & 0.68 & 0.53 & 0.43 & 0.61 & 0.33 & 0.30 & 0.19 & 0.06 & 0.22 & 0.30 & 0.20 & 0.08 & 0.04 & 0.15\\
\cmidrule{2-18}
& 4 & 0 & -- & 0.25 & 0.08 & 0.09 & 0.14 & 0.02 & 0.04 & 0.01 & 0.02 & 0.02 & 0.00 & 0.00 & 0.01 & 0.00 & 0.00\\
\midrule
\multirow{6}{*}{3}
& \multirow{3}{*}{3} & 10 & 0.38 & 0.41 & 0.26 & 0.20 & 0.31 & 0.18 & 0.14 & 0.12 & 0.04 & 0.12 & 0.07 & 0.07 & 0.02 & 0.03 & 0.05\\
& & 100 & 0.78 & 0.70 & 0.58 & 0.48 & 0.64 & 0.45 & 0.37 & 0.26 & 0.18 & 0.31 & 0.27 & 0.23 & 0.17 & 0.14 & 0.20\\
& & 1000 & \textbf{1.00} & \textbf{0.94} & \textbf{0.88} & \textbf{0.78} & \textbf{0.90} & \textbf{0.73} & \textbf{0.58} & \textbf{0.42} & 0.26 & \textbf{0.50} & \textbf{0.56} & \textbf{0.45} & \textbf{0.31} & \textbf{0.28} & \textbf{0.40}\\
\cmidrule{2-18}
& \multirow{3}{*}{4} & 10 & -- & 0.02 & 0.00 & 0.01 & 0.01 & 0.00 & 0.00 & 0.00 & 0.00 & 0.00 & 0.00 & 0.00 & 0.00 & 0.00 & 0.00\\
& & 100 & -- & 0.37 & 0.22 & 0.17 & 0.25 & 0.06 & 0.06 & 0.04 & 0.02 & 0.05 & 0.00 & 0.01 & 0.01 & 0.00 & 0.00\\
& & 1000 & -- & 0.65 & 0.51 & 0.41 & 0.52 & 0.23 & 0.13 & 0.06 & 0.07 & 0.12 & 0.07 & 0.08 & 0.06 & 0.03 & 0.06\\
\midrule
\multirow{6}{*}{4}
& \multirow{3}{*}{3} & 10 & 0.24 & 0.26 & 0.32 & 0.30 & 0.28 & 0.23 & 0.12 & 0.10 & 0.11 & 0.14 & 0.09 & 0.08 & 0.06 & 0.04 & 0.07\\
& & 100 & 0.53 & 0.45 & 0.39 & 0.52 & 0.47 & 0.29 & 0.37 & 0.26 & 0.15 & 0.26 & 0.27 & 0.25 & 0.20 & 0.14 & 0.21\\
& & 1000 & 0.82 & 0.74 & 0.66 & 0.68 & 0.73 & 0.58 & 0.44 & 0.33 & \textbf{0.29} & 0.41 & 0.41 & 0.36 & 0.27 & 0.20 & 0.31\\
\cmidrule{2-18}
& \multirow{3}{*}{4} & 10 & -- & 0.34 & 0.09 & 0.14 & 0.19 & 0.03 & 0.04 & 0.02 & 0.04 & 0.03 & 0.00 & 0.01 & 0.01 & 0.00 & 0.00\\
& & 100 & -- & 0.74 & 0.52 & 0.42 & 0.56 & 0.27 & 0.19 & 0.13 & 0.09 & 0.17 & 0.00 & 0.07 & 0.04 & 0.04 & 0.03\\
& & 1000 & -- & \textbf{0.94} & \textbf{0.87} & \textbf{0.73} & \textbf{0.85} & \textbf{0.55} & \textbf{0.39} & \textbf{0.31} & \textbf{0.22} & \textbf{0.36} & \textbf{0.17} & \textbf{0.22} & \textbf{0.17} & \textbf{0.16} & \textbf{0.18}\\
\bottomrule
\end{tabular}
\end{adjustbox}
\end{table*}%
  \endgroup

\end{table*}

\begin{table*}[p]

  \begingroup
  \renewenvironment{table}[1][]{\par\begingroup}{\par\endgroup}%
  \renewenvironment{table*}[1][]{\par\begingroup}{\par\endgroup}%
  \begin{table*}[!tbp]
\setlength{\tabcolsep}{6pt}
    \centering
    \footnotesize
    \extrarowheight=.9pt
\caption{\textbf{Coord. + Descr. continued fine-tuning on directionally complex maps: results by start-goal distance on fixed $10\times10$ maps.}}
\label{tab:hard_scaling_coorddescr_distance_full}
\begin{tabular}{ccc||cccc|c||cccc|c}
\multirow{2}{*}{\makecell{Train\\dirs.}} & \multirow{2}{*}{\makecell{Test\\dirs.}} & \multirow{2}{*}{Maps} & \multicolumn{5}{c||}{\textbf{ID test maps ($d_\infty \leq 5$)}} & \multicolumn{5}{c}{\textbf{OOD maps ($d_\infty \geq 6$)}}\\
& & & 2 & 3 & 4 & 5 & Avg & 6 & 7 & 8 & 9 & Avg\\
\toprule
\multirow{2}{*}{orig.} & 3 & 0 & 0.32 & 0.22 & 0.20 & 0.07 & 0.20 & 0.07 & 0.03 & 0.02 & 0.01 & 0.03\\
\cmidrule{2-13}
& 4 & 0 & 0.01 & 0.00 & 0.01 & 0.00 & 0.00 & 0.00 & 0.00 & 0.00 & 0.00 & 0.00\\
\midrule
\multirow{6}{*}{3}
& \multirow{3}{*}{3} & 10 & 0.34 & 0.14 & 0.11 & 0.06 & 0.16 & 0.01 & 0.02 & 0.00 & 0.00 & 0.01\\
& & 100 & 0.57 & 0.33 & 0.21 & 0.19 & 0.32 & 0.14 & 0.04 & 0.03 & 0.01 & 0.05\\
& & 1000 & \textbf{0.86} & \textbf{0.71} & \textbf{0.63} & \textbf{0.52} & \textbf{0.68} & \textbf{0.45} & \textbf{0.40} & 0.33 & \textbf{0.29} & \textbf{0.36}\\
\cmidrule{2-13}
& \multirow{3}{*}{4} & 10 & 0.02 & 0.01 & 0.00 & 0.00 & 0.01 & 0.00 & 0.00 & 0.00 & 0.00 & 0.00\\
& & 100 & 0.06 & 0.01 & 0.01 & 0.01 & 0.02 & 0.00 & 0.00 & 0.00 & 0.00 & 0.00\\
& & 1000 & 0.26 & 0.18 & 0.12 & 0.09 & 0.16 & 0.09 & 0.04 & 0.02 & 0.01 & 0.04\\
\midrule
\multirow{6}{*}{4}
& \multirow{3}{*}{3} & 10 & 0.12 & 0.08 & 0.06 & 0.06 & 0.08 & 0.01 & 0.02 & 0.02 & 0.00 & 0.01\\
& & 100 & 0.30 & 0.29 & 0.22 & 0.24 & 0.26 & 0.09 & 0.07 & 0.10 & 0.06 & 0.08\\
& & 1000 & 0.62 & 0.54 & 0.54 & 0.43 & 0.53 & 0.34 & 0.36 & \textbf{0.39} & \textbf{0.29} & 0.34\\
\cmidrule{2-13}
& \multirow{3}{*}{4} & 10 & 0.04 & 0.01 & 0.01 & 0.00 & 0.01 & 0.01 & 0.00 & 0.00 & 0.00 & 0.00\\
& & 100 & 0.19 & 0.15 & 0.09 & 0.07 & 0.12 & 0.02 & 0.01 & 0.01 & 0.00 & 0.01\\
& & 1000 & \textbf{0.53} & \textbf{0.53} & \textbf{0.40} & \textbf{0.35} & \textbf{0.45} & \textbf{0.26} & \textbf{0.21} & \textbf{0.21} & \textbf{0.09} & \textbf{0.19}\\
\bottomrule
\end{tabular}
\end{table*}%
  \endgroup

\end{table*}

\begin{table*}[p]
\begin{minipage}{\textwidth}
\normalsize
\begin{multicols}{2}
\setlength{\parskip}{\appendixproseparskip}
\textbf{Transfer and retention after 1,000-example fine-tuning.}

\end{multicols}
\end{minipage}
\medskip

  \begingroup
  \renewenvironment{table}[1][]{\par\begingroup}{\par\endgroup}%
  \renewenvironment{table*}[1][]{\par\begingroup}{\par\endgroup}%
  \begin{table*}[!tbp]
\centering\footnotesize
\setlength{\tabcolsep}{3pt}
\extrarowheight=.9pt
\caption{\textbf{Transfer to turn complexity after A* fine-tuning on 1,000 four-direction maps: varying map size.} No turn-targeted fine-tuning is used.}
\label{tab:direction_to_turn_transfer_map}
\begin{adjustbox}{max width=\textwidth}
\begin{tabular}{cc||cccc|c||cccc|c||cccc|c}
\multirow{2}{*}{Turns} & \multirow{2}{*}{Checkpoint} & \multicolumn{5}{c||}{\textbf{ID test maps ($d_\infty \leq 5$)}} & \multicolumn{5}{c||}{\textbf{OOD maps (random $d_\infty$)}} & \multicolumn{5}{c}{\textbf{OOD maps ($d_\infty \geq 6$)}}\\
& & 3x3 & 4x4 & 5x5 & 6x6 & Avg & 7x7 & 8x8 & 9x9 & 10x10 & Avg & 7x7 & 8x8 & 9x9 & 10x10 & Avg\\
\toprule
\multirow{2}{*}{5} & orig. & -- & 0.65 & 0.49 & 0.23 & 0.46 & 0.22 & 0.10 & 0.08 & 0.03 & 0.11 & 0.19 & 0.15 & 0.12 & 0.06 & 0.13\\
& +1k 4-dir & -- & \textbf{1.00} & \textbf{0.95} & \textbf{0.83} & \textbf{0.93} & \textbf{0.74} & \textbf{0.61} & \textbf{0.43} & \textbf{0.31} & \textbf{0.52} & \textbf{0.70} & \textbf{0.61} & \textbf{0.43} & \textbf{0.31} & \textbf{0.51}\\
\midrule
\multirow{2}{*}{6} & orig. & -- & -- & 0.32 & 0.14 & 0.23 & 0.14 & 0.04 & 0.03 & 0.03 & 0.06 & 0.17 & 0.07 & 0.03 & 0.03 & 0.08\\
& +1k 4-dir & -- & -- & \textbf{0.96} & \textbf{0.77} & \textbf{0.87} & \textbf{0.68} & \textbf{0.48} & \textbf{0.32} & \textbf{0.16} & \textbf{0.41} & \textbf{0.62} & \textbf{0.52} & \textbf{0.36} & \textbf{0.21} & \textbf{0.43}\\
\midrule
\multirow{2}{*}{7} & orig. & -- & -- & 0.27 & 0.12 & 0.20 & 0.05 & 0.02 & 0.02 & 0.00 & 0.02 & 0.11 & 0.04 & 0.01 & 0.01 & 0.04\\
& +1k 4-dir & -- & -- & \textbf{0.98} & \textbf{0.80} & \textbf{0.89} & \textbf{0.57} & \textbf{0.35} & \textbf{0.24} & \textbf{0.12} & \textbf{0.32} & \textbf{0.62} & \textbf{0.42} & \textbf{0.25} & \textbf{0.18} & \textbf{0.37}\\
\midrule
\multirow{2}{*}{8} & orig. & -- & -- & 0.18 & 0.01 & 0.10 & 0.01 & 0.01 & 0.01 & 0.01 & 0.01 & 0.04 & 0.02 & 0.02 & 0.00 & 0.02\\
& +1k 4-dir & -- & -- & \textbf{0.91} & \textbf{0.77} & \textbf{0.84} & \textbf{0.49} & \textbf{0.28} & \textbf{0.21} & \textbf{0.08} & \textbf{0.27} & \textbf{0.51} & \textbf{0.27} & \textbf{0.16} & \textbf{0.07} & \textbf{0.25}\\
\bottomrule
\end{tabular}
\end{adjustbox}
\end{table*}
  \endgroup

\end{table*}

\begin{table*}[p]

  \begingroup
  \renewenvironment{table}[1][]{\par\begingroup}{\par\endgroup}%
  \renewenvironment{table*}[1][]{\par\begingroup}{\par\endgroup}%
  \begin{table*}[!tbp]
\centering\footnotesize
\setlength{\tabcolsep}{3pt}
\extrarowheight=.9pt
\caption{\textbf{Transfer to turn complexity after A* fine-tuning on 1,000 four-direction maps: fixed $10\times10$ size.} No turn-targeted fine-tuning is used.}
\label{tab:direction_to_turn_transfer_distance}
\begin{adjustbox}{max width=\textwidth}
\begin{tabular}{cc||cccc|c||cccc|c}
\multirow{2}{*}{Turns} & \multirow{2}{*}{Checkpoint} & \multicolumn{5}{c||}{\textbf{ID test maps ($d_\infty \leq 5$)}} & \multicolumn{5}{c}{\textbf{OOD maps ($d_\infty \geq 6$)}}\\
& & 2 & 3 & 4 & 5 & Avg & 6 & 7 & 8 & 9 & Avg\\
\toprule
\multirow{2}{*}{5} & orig. & 0.11 & 0.06 & 0.11 & 0.10 & 0.10 & 0.14 & 0.08 & 0.06 & 0.08 & 0.09\\
& +1k 4-dir & \textbf{0.44} & \textbf{0.37} & \textbf{0.39} & \textbf{0.34} & \textbf{0.39} & \textbf{0.37} & \textbf{0.30} & \textbf{0.30} & \textbf{0.27} & \textbf{0.31}\\
\midrule
\multirow{2}{*}{6} & orig. & 0.01 & 0.03 & 0.07 & 0.05 & 0.04 & 0.03 & 0.03 & 0.04 & 0.03 & 0.03\\
& +1k 4-dir & \textbf{0.21} & \textbf{0.22} & \textbf{0.27} & \textbf{0.25} & \textbf{0.24} & \textbf{0.24} & \textbf{0.16} & \textbf{0.21} & \textbf{0.12} & \textbf{0.18}\\
\midrule
\multirow{2}{*}{7} & orig. & 0.01 & 0.03 & 0.01 & 0.03 & 0.02 & 0.03 & 0.01 & 0.01 & 0.00 & 0.01\\
& +1k 4-dir & \textbf{0.15} & \textbf{0.19} & \textbf{0.20} & \textbf{0.14} & \textbf{0.17} & \textbf{0.16} & \textbf{0.11} & \textbf{0.13} & \textbf{0.18} & \textbf{0.15}\\
\midrule
\multirow{2}{*}{8} & orig. & 0.00 & 0.00 & 0.00 & 0.00 & 0.00 & 0.00 & 0.01 & 0.01 & 0.00 & 0.01\\
& +1k 4-dir & \textbf{0.04} & \textbf{0.13} & \textbf{0.07} & \textbf{0.08} & \textbf{0.08} & \textbf{0.09} & \textbf{0.10} & \textbf{0.12} & \textbf{0.08} & \textbf{0.10}\\
\bottomrule
\end{tabular}
\end{adjustbox}
\end{table*}
  \endgroup

\end{table*}

\begin{table*}[p]

  \begingroup
  \renewenvironment{table}[1][]{\par\begingroup}{\par\endgroup}%
  \renewenvironment{table*}[1][]{\par\begingroup}{\par\endgroup}%
  \begin{table*}[!tbp]
\centering\footnotesize
\setlength{\tabcolsep}{3pt}
\extrarowheight=.9pt
\caption{\textbf{Retention on the original evaluations after fine-tuning on 1,000 four-direction maps: varying map size.}}
\label{tab:hard_scaling_original_splits_map}
\begin{adjustbox}{max width=\textwidth}
\begin{tabular}{ll||cccc|c||cccc|c||cccc|c}
\multirow{2}{*}{CoT} & \multirow{2}{*}{Checkpoint} & \multicolumn{5}{c||}{\textbf{ID test maps ($d_\infty \leq 5$)}} & \multicolumn{5}{c||}{\textbf{OOD maps (random $d_\infty$)}} & \multicolumn{5}{c}{\textbf{OOD maps ($d_\infty \geq 6$)}}\\
& & 3x3 & 4x4 & 5x5 & 6x6 & Avg & 7x7 & 8x8 & 9x9 & 10x10 & Avg & 7x7 & 8x8 & 9x9 & 10x10 & Avg\\
\toprule
\multirow{2}{*}{A* w/ Coord.} & orig. & \textbf{1.00} & 0.99 & 0.96 & 0.95 & 0.98 & \textbf{0.96} & 0.91 & \textbf{0.88} & 0.82 & 0.89 & 0.87 & 0.76 & \textbf{0.73} & 0.62 & 0.74\\
& +1k 4-dir & \textbf{1.00} & \textbf{1.00} & \textbf{1.00} & \textbf{0.98} & \textbf{1.00} & \textbf{0.96} & \textbf{0.93} & \textbf{0.88} & \textbf{0.85} & \textbf{0.90} & \textbf{0.92} & \textbf{0.87} & 0.71 & \textbf{0.68} & \textbf{0.79}\\
\midrule
\multirow{2}{*}{Coord. + Descr.} & orig. & \textbf{1.00} & \textbf{0.95} & \textbf{0.98} & \textbf{0.86} & \textbf{0.95} & \textbf{0.97} & \textbf{0.93} & \textbf{0.90} & \textbf{0.88} & \textbf{0.92} & \textbf{0.89} & \textbf{0.87} & \textbf{0.81} & \textbf{0.77} & \textbf{0.83}\\
& +1k 4-dir & 0.72 & 0.79 & 0.76 & 0.67 & 0.74 & 0.83 & 0.81 & 0.76 & 0.76 & 0.79 & 0.74 & 0.65 & 0.65 & 0.61 & 0.66\\
\bottomrule
\end{tabular}
\end{adjustbox}
\end{table*}
  \endgroup

\end{table*}

\begin{table*}[p]

  \begingroup
  \renewenvironment{table}[1][]{\par\begingroup}{\par\endgroup}%
  \renewenvironment{table*}[1][]{\par\begingroup}{\par\endgroup}%
  \begin{table*}[!tbp]
\centering\footnotesize
\setlength{\tabcolsep}{3pt}
\extrarowheight=.9pt
\caption{\textbf{Retention on the original evaluations after fine-tuning on 1,000 four-direction maps: fixed $10\times10$ size.}}
\label{tab:hard_scaling_original_splits_distance}
\begin{adjustbox}{max width=\textwidth}
\begin{tabular}{ll||cccc|c||cccc|c}
\multirow{2}{*}{CoT} & \multirow{2}{*}{Checkpoint} & \multicolumn{5}{c||}{\textbf{ID test maps ($d_\infty \leq 5$)}} & \multicolumn{5}{c}{\textbf{OOD maps ($d_\infty \geq 6$)}}\\
& & 2 & 3 & 4 & 5 & Avg & 6 & 7 & 8 & 9 & Avg\\
\toprule
\multirow{2}{*}{A* w/ Coord.} & orig. & \textbf{0.99} & 0.98 & 0.96 & 0.89 & \textbf{0.95} & 0.85 & 0.63 & 0.56 & 0.46 & 0.62\\
& +1k 4-dir & 0.85 & \textbf{0.99} & \textbf{0.98} & \textbf{0.98} & \textbf{0.95} & \textbf{0.96} & \textbf{0.92} & \textbf{0.90} & \textbf{0.85} & \textbf{0.91}\\
\midrule
\multirow{2}{*}{Coord. + Descr.} & orig. & \textbf{0.95} & \textbf{0.94} & \textbf{0.89} & \textbf{0.74} & \textbf{0.88} & \textbf{0.75} & 0.56 & 0.44 & 0.44 & 0.55\\
& +1k 4-dir & 0.76 & 0.71 & 0.70 & 0.66 & 0.71 & 0.65 & \textbf{0.60} & \textbf{0.49} & \textbf{0.49} & \textbf{0.56}\\
\bottomrule
\end{tabular}
\end{adjustbox}
\end{table*}
  \endgroup

\end{table*}

\begin{table*}[p]
\begin{minipage}{\textwidth}
\normalsize
\begin{multicols}{2}
\setlength{\parskip}{\appendixproseparskip}
\subsection{Structural diagnostics}
\label{sec:structural_diagnostics}

Table~\ref{tab:structural_counts} and Fig.~\ref{fig:structural-distributions} report the direction and turn counts in the standard datasets. Table~\ref{tab:structural_confounds} compares shortest-path lengths and hole counts across the structural test sets, while Fig.~\ref{fig:hard-scaling} in App.~\ref{sec:continued_scaling} shows the main trends from continued fine-tuning on directionally complex maps.

Table~\ref{tab:structural_failure_modes} shows that most failures contain parseable plans and most of these enter a hole.

\end{multicols}
\end{minipage}
\par\medskip
\noindent\begin{minipage}[t]{\dimexpr(\textwidth-\columnsep)/2\relax}\vspace{0pt}

  \begingroup
  \renewenvironment{table}[1][]{\par\begingroup}{\par\endgroup}%
  \renewenvironment{table*}[1][]{\par\begingroup}{\par\endgroup}%
  \begin{table}[t]
\centering
\footnotesize
\setlength{\tabcolsep}{1.5pt} 
\vspace{4mm}
\caption{\textbf{Confounds in the map-size structural test sets.} Values are averaged over OOD maps. $D^*$ and $T^*$ denote the minimum direction and turn counts over shortest paths.}
\label{tab:structural_confounds}
\begin{tabular}{cc||cc||cc}
\multirow{2}{*}{Test} & \multirow{2}{*}{Bucket} & \multicolumn{2}{c||}{\makecell{OOD maps, random\\$d_\infty$}} & \multicolumn{2}{c}{\makecell{OOD maps,\\$d_\infty\geq6$}}\\
& & shortest length & holes & shortest length & holes\\
\toprule
\multirow{2}{*}{Direction} & $D^*=3$ & 9.05 & 18.15 & 11.28 & 18.11\\
& $D^*=4$ & 12.79 & 19.13 & 15.99 & 19.39\\
\midrule
\multirow{4}{*}{Turn} & $T^*=5$ & 12.61 & 19.25 & 12.72 & 19.43\\
& $T^*=6$ & 14.24 & 20.07 & 13.87 & 20.07\\
& $T^*=7$ & 15.91 & 20.37 & 15.36 & 20.48\\
& $T^*=8$ & 17.35 & 20.65 & 16.94 & 20.89\\
\bottomrule
\end{tabular}
\end{table}
  \endgroup

\medskip

  \begingroup
  \renewenvironment{table}[1][]{\par\begingroup}{\par\endgroup}%
  \renewenvironment{table*}[1][]{\par\begingroup}{\par\endgroup}%
  \begin{table}[t]
\centering
\footnotesize
\setlength{\tabcolsep}{3pt}
\vspace{2mm}
\caption{\textbf{Shortest-path structure in the standard datasets.} These counts determine the direction and turn buckets used in the structural test sets.}
\label{tab:structural_counts}
\begin{tabular}{ll||rr||rr}
\multirow{2}{*}{Measure} & \multirow{2}{*}{Count} & \multicolumn{2}{c||}{Map size} & \multicolumn{2}{c}{Start-goal distance}\\
& & Train & Test & Train & Test\\
\toprule
\multirow{4}{*}{Direction} & 1 & 292 & 356 & 292 & 164\\
& 2 & 610 & 1448 & 610 & 1261\\
& 3 & 87 & 183 & 88 & 161\\
& 4 & 11 & 13 & 10 & 12\\
\midrule
\multirow{8}{*}{Turn} & 0 & 292 & 356 & 292 & 164\\
& 1 & 538 & 1137 & 538 & 955\\
& 2 & 125 & 388 & 131 & 375\\
& 3 & 35 & 85 & 31 & 73\\
& 4 & 9 & 24 & 8 & 24\\
& 5 & 0 & 7 & 0 & 3\\
& 6 & 1 & 2 & 0 & 3\\
& 7 & 0 & 1 & 0 & 1\\
\bottomrule
\end{tabular}
\end{table}
  \endgroup

\end{minipage}\hfill\begin{minipage}[t]{\dimexpr(\textwidth-\columnsep)/2\relax}\vspace{0pt}
\captionsetup{type=figure}

    \centering
    \includegraphics[width=\columnwidth]{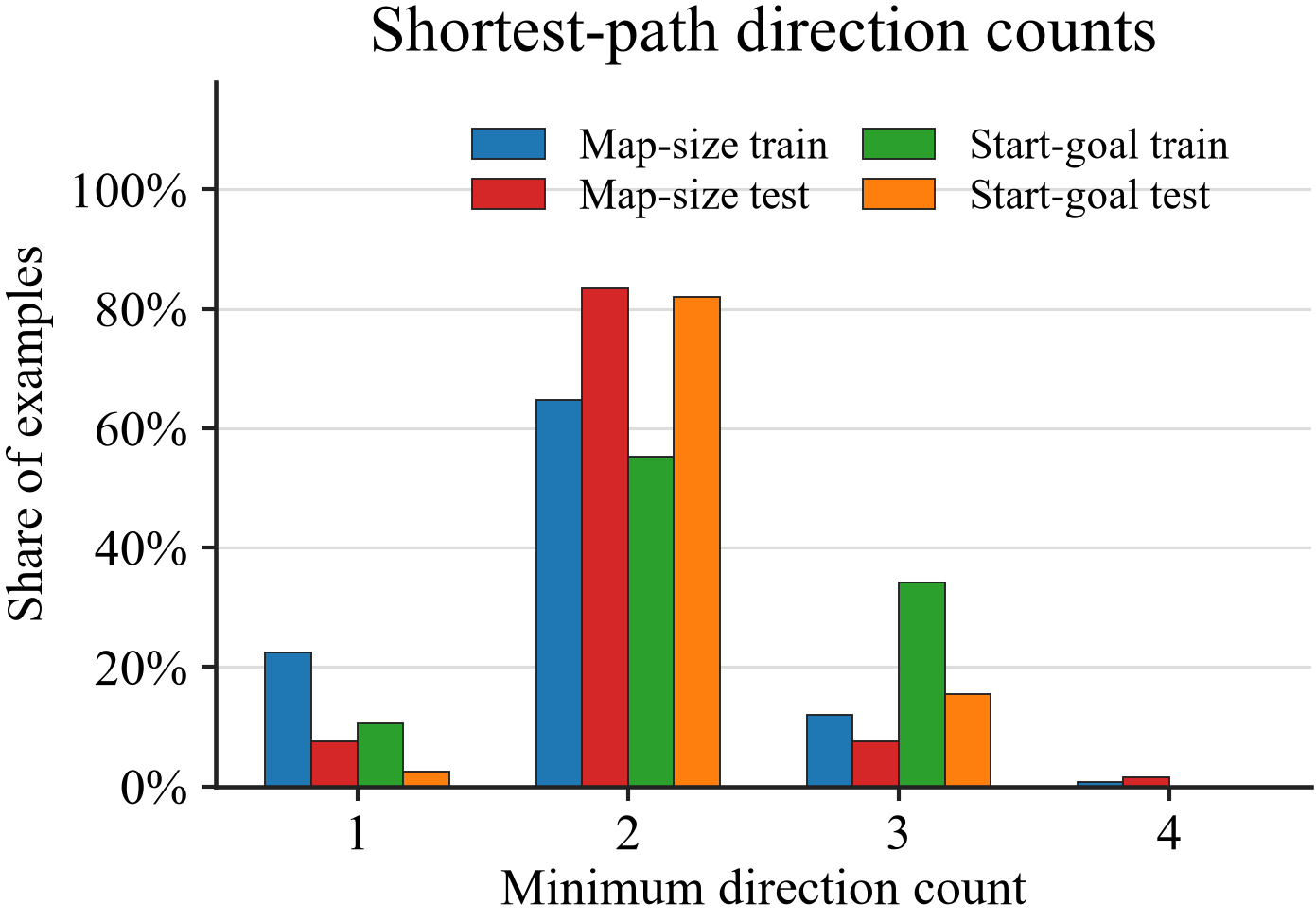}
    \includegraphics[width=\columnwidth]{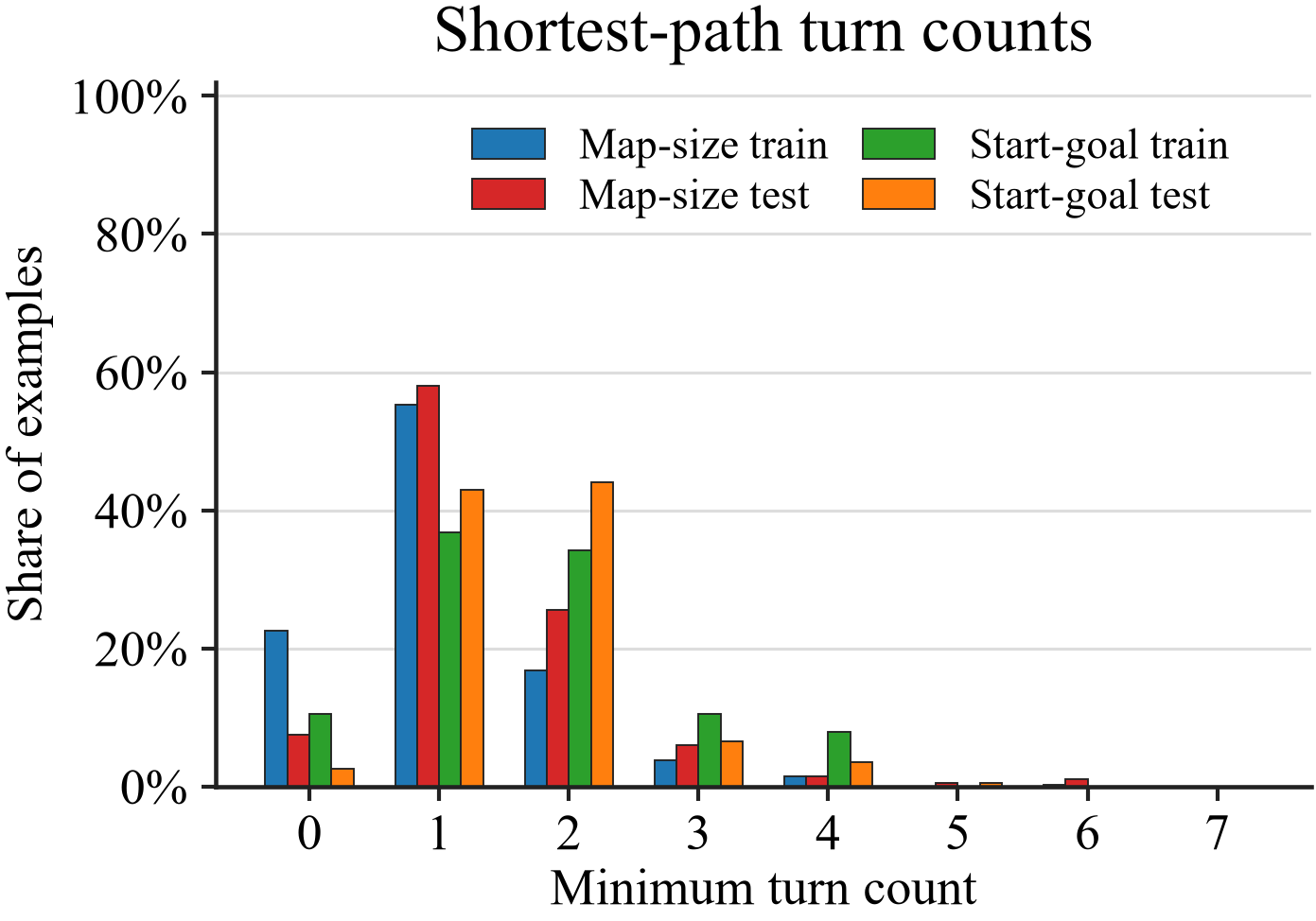}
    \caption{\textbf{Shortest-path structure in the standard datasets.} Counts from Table~\ref{tab:structural_counts} are normalized within each train/test split.}
    \label{fig:structural-distributions}

\end{minipage}
\par\medskip

  \begingroup
  \renewenvironment{table}[1][]{\par\begingroup}{\par\endgroup}%
  \renewenvironment{table*}[1][]{\par\begingroup}{\par\endgroup}%
  \begin{table*}[t]
\centering
\footnotesize
\setlength{\tabcolsep}{4pt}
\caption{\textbf{Failure modes after continued fine-tuning on 1,000 directionally complex maps.} Map-size rows pool the ID and random-OOD test sets; start-goal-distance rows pool the $d_\infty\geq6$ larger-map and fixed-$10\times10$ test sets. ``No answer'' and ``Wrong'' are shares of all failures; ``Hole'' and ``No goal'' are shares of parseable wrong answers.}
\label{tab:structural_failure_modes}
\begin{tabular}{cl||lr||rr||rr}
Evaluation & CoT format & Train$\to$test & Failures & No answer & Wrong & Hole & No goal\\
\toprule
\multirow{8}{*}{Map size}
& \multirow{4}{*}{A*} & $3\to3$ & 292 & 0.05 & 0.95 & 0.84 & 0.16\\
& & $4\to3$ & 234 & 0.06 & 0.94 & 0.93 & 0.07\\
& & $3\to4$ & 586 & 0.11 & 0.89 & 0.88 & 0.12\\
& & $4\to4$ & 372 & 0.17 & 0.83 & 0.94 & 0.06\\
\noalign{\vskip-1.25pt}\cline{2-8}\noalign{\vskip1.25pt}
& \multirow{4}{*}{Coord. + Descr.} & $3\to3$ & 443 & 0.20 & 0.80 & 0.95 & 0.05\\
& & $4\to3$ & 584 & 0.21 & 0.79 & 0.97 & 0.03\\
& & $3\to4$ & 845 & 0.21 & 0.79 & 0.99 & 0.01\\
& & $4\to4$ & 555 & 0.22 & 0.78 & 0.98 & 0.02\\
\midrule
\multirow{8}{*}{\makecell[c]{Start-goal\\distance}}
& \multirow{4}{*}{A*} & $3\to3$ & 993 & 0.06 & 0.94 & 0.91 & 0.09\\
& & $4\to3$ & 627 & 0.06 & 0.94 & 0.94 & 0.06\\
& & $3\to4$ & 2042 & 0.18 & 0.82 & 0.95 & 0.05\\
& & $4\to4$ & 1244 & 0.13 & 0.87 & 0.95 & 0.05\\
\noalign{\vskip-1.25pt}\cline{2-8}\noalign{\vskip1.25pt}
& \multirow{4}{*}{Coord. + Descr.} & $3\to3$ & 1253 & 0.13 & 0.87 & 0.97 & 0.03\\
& & $4\to3$ & 1457 & 0.32 & 0.68 & 0.98 & 0.02\\
& & $3\to4$ & 2196 & 0.20 & 0.80 & 0.98 & 0.02\\
& & $4\to4$ & 1746 & 0.24 & 0.76 & 0.98 & 0.02\\
\bottomrule
\end{tabular}
\end{table*}
  \endgroup

\end{table*}

\begin{table*}[p]
\noindent\begin{minipage}[t]{\dimexpr(\textwidth-\columnsep)/2\relax}\vspace{0pt}
\normalsize
\setlength{\parskip}{\appendixproseparskip}
\subsection{Qwen3.5 supplementary analyses}
\label{sec:qwen35_appendix}

Tables~\ref{tab:qwen35_structural_map_full} and~\ref{tab:qwen35_structural_distance_full} report the zero-shot results on the structural test sets for Qwen3.5-2B. Four-direction maps are more challenging than three-direction maps, and accuracy generally decreases as the number of turns increases. Table~\ref{tab:answer_extraction_sensitivity} reports the sensitivity of the results to the position of the extracted answer.

\end{minipage}\hfill\begin{minipage}[t]{\dimexpr(\textwidth-\columnsep)/2\relax}\vspace{0pt}

  \begingroup
  \renewenvironment{table}[1][]{\par\begingroup}{\par\endgroup}%
  \renewenvironment{table*}[1][]{\par\begingroup}{\par\endgroup}%
  \begin{table}[t]
\centering
\footnotesize
\setlength{\tabcolsep}{1.5pt}
\caption{\textbf{Answer-extraction sensitivity.} Accuracy is aggregated over evaluations of map size and start-goal distance.}
\label{tab:answer_extraction_sensitivity}
\begin{tabular}{@{}ll||rrr@{}}
Model & CoT format & First boxed & Last boxed & Change\\
\toprule
\multirow{6}{*}{Qwen2.5-VL} & zero-shot & 0.072 & 0.072 & 0.000\\
& no CoT & 0.403 & 0.403 & 0.000\\
& Descr. & 0.393 & 0.393 & 0.000\\
& Coord. & 0.720 & 0.720 & 0.000\\
& Coord. + Descr. & 0.812 & 0.812 & 0.000\\
& A* w/ coord. & 0.822 & 0.822 & 0.000\\
\midrule
Qwen3.5-0.8B & zero-shot & 0.213 & 0.233 & +0.020\\
Qwen3.5-2B & zero-shot & 0.806 & 0.826 & +0.020\\
Qwen3.5-4B & zero-shot & 0.032 & 0.293 & +0.261\\
\bottomrule
\end{tabular}
\end{table}
  \endgroup

\end{minipage}

\end{table*}

\begin{table*}[p]

\centering
\footnotesize
\setlength{\tabcolsep}{1.8pt}
\caption{\textbf{Qwen3.5-2B zero-shot results on structural test sets by map size.} The last group contains larger maps with $d_\infty\geq6$.}
\label{tab:qwen35_structural_map_full}
\begin{adjustbox}{max width=\textwidth}
\begin{tabular}{ll||cccc|c||cccc|c||cccc|c}
\multirow{2}{*}{Test} & \multirow{2}{*}{Bucket} & \multicolumn{5}{c||}{\textbf{ID test maps ($d_\infty \leq 5$)}} & \multicolumn{5}{c||}{\textbf{OOD maps (random $d_\infty$)}} & \multicolumn{5}{c}{\textbf{OOD maps ($d_\infty \geq 6$)}}\\
& & 3x3 & 4x4 & 5x5 & 6x6 & Avg & 7x7 & 8x8 & 9x9 & 10x10 & Avg & 7x7 & 8x8 & 9x9 & 10x10 & Avg\\
\toprule
\multirow{2}{*}{Direction} & 3 directions & 0.95 & 0.95 & 0.88 & 0.85 & 0.91 & 0.79 & 0.69 & 0.57 & 0.45 & 0.63 & 0.70 & 0.64 & 0.49 & 0.44 & 0.57\\
& 4 directions & -- & 0.85 & 0.75 & 0.71 & 0.77 & 0.54 & 0.50 & 0.34 & 0.30 & 0.42 & 0.39 & 0.34 & 0.25 & 0.21 & 0.30\\
\midrule
\multirow{4}{*}{Turn}
& 5 turns & -- & 0.90 & 0.80 & 0.69 & 0.80 & 0.54 & 0.45 & 0.40 & 0.37 & 0.44 & 0.60 & 0.50 & 0.37 & 0.30 & 0.44\\
& 6 turns & -- & -- & 0.79 & 0.61 & 0.70 & 0.53 & 0.47 & 0.32 & 0.18 & 0.38 & 0.58 & 0.44 & 0.31 & 0.21 & 0.39\\
& 7 turns & -- & -- & 0.80 & 0.63 & 0.72 & 0.44 & 0.34 & 0.22 & 0.16 & 0.29 & 0.55 & 0.34 & 0.26 & 0.17 & 0.33\\
& 8 turns & -- & -- & 0.73 & 0.55 & 0.64 & 0.36 & 0.26 & 0.19 & 0.13 & 0.24 & 0.44 & 0.35 & 0.25 & 0.14 & 0.30\\
\bottomrule
\end{tabular}
\end{adjustbox}

\end{table*}

\begin{table*}[p]

\centering
\footnotesize
\setlength{\tabcolsep}{3.5pt}
\caption{\textbf{Qwen3.5-2B zero-shot results on structural test sets by start-goal distance on fixed $10\times10$ maps.}}
\label{tab:qwen35_structural_distance_full}
\begin{tabular}{ll||cccc|c||cccc|c}
\multirow{2}{*}{Test} & \multirow{2}{*}{Bucket} & \multicolumn{5}{c||}{\textbf{ID test maps ($d_\infty \leq 5$)}} & \multicolumn{5}{c}{\textbf{OOD maps ($d_\infty \geq 6$)}}\\
& & 2 & 3 & 4 & 5 & Avg & 6 & 7 & 8 & 9 & Avg\\
\toprule
\multirow{2}{*}{Direction} & 3 directions & 0.80 & 0.69 & 0.56 & 0.53 & 0.65 & 0.45 & 0.42 & 0.32 & 0.30 & 0.37\\
& 4 directions & 0.41 & 0.40 & 0.31 & 0.25 & 0.34 & 0.21 & 0.19 & 0.14 & 0.08 & 0.16\\
\midrule
\multirow{4}{*}{Turn}
& 5 turns & 0.37 & 0.29 & 0.32 & 0.31 & 0.32 & 0.31 & 0.34 & 0.35 & 0.30 & 0.33\\
& 6 turns & 0.18 & 0.19 & 0.22 & 0.27 & 0.22 & 0.27 & 0.27 & 0.26 & 0.25 & 0.26\\
& 7 turns & 0.17 & 0.14 & 0.13 & 0.23 & 0.17 & 0.17 & 0.18 & 0.18 & 0.23 & 0.19\\
& 8 turns & 0.10 & 0.10 & 0.07 & 0.08 & 0.09 & 0.20 & 0.12 & 0.14 & 0.17 & 0.16\\
\bottomrule
\end{tabular}

\end{table*}

\begin{figure*}[p]
\begin{minipage}{\textwidth}
\normalsize
\begin{multicols}{2}
\setlength{\parskip}{\appendixproseparskip}

\subsection{Additional examples}
\label{sec:additional_figures}

\textbf{Examples of generated CoT.} We show examples of A*, \textit{grid + description}, and \textit{coordinates + description} traces that our models generated for the same OOD map in Figs.~\ref{fig:astar_example}, \ref{fig:grid_example}, and~\ref{fig:coord_desc_example}, respectively.

\textbf{Examples of other input and trace formats.} Examples of the \textit{table} and \textit{sheet-style coordinate} input formats are shown in Fig.~\ref{fig:maze-repr-appendix}; their respective reasoning traces, including the sheet-coordinate variant of the A* CoT, are shown in Fig.~\ref{fig:reasoning-traces-appendix}.

\end{multicols}
\end{minipage}
\medskip

  \begingroup
  \renewenvironment{figure}[1][]{\begingroup}{\endgroup}%
  \renewenvironment{figure*}[1][]{\begingroup}{\endgroup}%
  \input{tables/example_cot_astar}%
  \endgroup

\end{figure*}

\begin{figure*}[p]
\noindent\begin{minipage}[t]{\dimexpr(\textwidth-\columnsep)/2\relax}\vspace{0pt}

  \begingroup
  \renewenvironment{figure}[1][]{\begingroup}{\endgroup}%
  \renewenvironment{figure*}[1][]{\begingroup}{\endgroup}%
  \input{tables/example_cot_grid}%
  \endgroup

\end{minipage}\hfill\begin{minipage}[t]{\dimexpr(\textwidth-\columnsep)/2\relax}\vspace{0pt}

  \begingroup
  \renewenvironment{figure}[1][]{\begingroup}{\endgroup}%
  \renewenvironment{figure*}[1][]{\begingroup}{\endgroup}%
  \input{tables/example_cot_coorddesc}%
  \endgroup

\end{minipage}

\end{figure*}

\clearpage

\begin{figure*}[p]

    \centering
    \begingroup
\newcommand{\RepFont}{\fontfamily{pcr}\fontseries{m}\selectfont\tiny}
\newcommand{\RepHeight}{3.20cm}

\newcommand{\RepBox}[2]{%
\begin{tcolorbox}[
    enhanced,
    colback=gray!12,
    colframe=gray!12,
    boxrule=0pt,
    arc=4pt,
    boxsep=0pt,
    left=4pt,
    right=4pt,
    top=3pt,
    bottom=3pt,
    width=#1,
    height=\RepHeight,
    valign=center,
]
{\RepFont #2}
\end{tcolorbox}%
}

\newcommand{\AsciiMapTable}{%
\begingroup
\RepFont
\def\Pipe{\textbar}%
\def\Blank{\makebox[3.5em][c]{}}%
\def\Slot##1{\makebox[3.5em][c]{##1}}%
\Pipe\Blank\Pipe\Slot{Col 1}\Pipe\Slot{Col 2}\Pipe\Slot{Col 3}\Pipe\\
\Pipe\Slot{Row 1}\Pipe\Slot{\_}\Pipe\Slot{*}\Pipe\Slot{\_}\Pipe\\
\Pipe\Slot{Row 2}\Pipe\Slot{\_}\Pipe\Slot{\#}\Pipe\Slot{\_}\Pipe\\
\Pipe\Slot{Row 3}\Pipe\Slot{\_}\Pipe\Slot{@}\Pipe\Slot{\#}\Pipe
\endgroup
}

\resizebox{\linewidth}{!}{%
\begin{tabular}{@{}c@{\hspace{4pt}}c@{\hspace{4pt}}c@{\hspace{4pt}}c@{}}
\multicolumn{1}{c}{\normalsize Image} &
\multicolumn{1}{c}{\normalsize Table} &
\multicolumn{1}{c}{\normalsize Coordinates (sheet)}
\\[-1pt]

\begin{tcolorbox}[
    enhanced,
    colback=white,
    colframe=white,
    boxrule=0pt,
    arc=0pt,
    boxsep=0pt,
    left=0pt,
    right=0pt,
    top=0pt,
    bottom=0pt,
    width=3.25cm,
    height=\RepHeight,
    valign=center,
]
\centering
\includegraphics[height=\RepHeight,keepaspectratio]{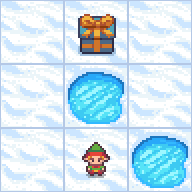}
\end{tcolorbox}
&
\RepBox{3.90cm}{\AsciiMapTable}
&
\RepBox{8.3cm}{%
Coordinates use sheet-style labels: a1 is the top-left corner,\\
letters move right, and numbers move down.\\
The map is a 3 by 3 grid.\\
The player starts at b3 and the goal is at b1.\\
Holes are located at \{b2,c3\}.\\
Every coordinate not listed as a hole is a safe cell.%
}
\end{tabular}%
}
\endgroup

    \caption{\textbf{Maze representations for tabular and spreadsheet-style coordinates.}
    We show the same maze as an \textit{image}, an ASCII \textit{table}, and a spreadsheet-style \textit{coordinates} representation.
    The sheet coordinate system names cells by column letter and row number, with a1 at the top-left corner. This makes the state of the table compact to list in text and removes the need to list empty cells.
    }
    \label{fig:maze-repr-appendix}

\end{figure*}

\begin{figure*}[p]

    \centering
    \input{tables/reasoning_traces_all_steps_appendix_table}
    \caption{\textbf{Reasoning traces with table and sheet-style coordinates.}
    \textbf{Top}: We show the same first two steps of the solution trace of the maze as ASCII tables, sheet-style coordinate states, and sheet-style A* reasoning.
    Sheet-style formats simply replace numerical notation with spreadsheet notation. For example, (0,0) is rewritten as a1. 
    \textbf{Bottom}: The beginning of the chain-of-thought for both numerical- and sheet-style coordinate variants of the A* format.
    }
    \label{fig:reasoning-traces-appendix}

\end{figure*}

\clearpage




\begin{thebibliography}{28}
\providecommand{\natexlab}[1]{#1}
\providecommand{\url}[1]{\texttt{#1}}
\expandafter\ifx\csname urlstyle\endcsname\relax
  \providecommand{\doi}[1]{doi: #1}\else
  \providecommand{\doi}{doi: \begingroup \urlstyle{rm}\Url}\fi

\bibitem[Anthropic(2024)]{claude3}
Anthropic.
\newblock Introducing the next generation of claude, 2024.
\newblock URL \url{https://www.anthropic.com/news/claude-3-family}.

\bibitem[Bai et~al.(2025)Bai, Chen, Liu, Wang, Ge, Song, Dang, Wang, Wang, Tang, et~al.]{bai2025qwen2}
Bai, S., Chen, K., Liu, X., Wang, J., Ge, W., Song, S., Dang, K., Wang, P., Wang, S., Tang, J., et~al.
\newblock Qwen2. 5-vl technical report.
\newblock \emph{arXiv preprint arXiv:2502.13923}, 2025.

\bibitem[Bai et~al.(2024)Bai, Geng, Mangalam, Bar, Yuille, Darrell, Malik, and Efros]{bai2024sequential}
Bai, Y., Geng, X., Mangalam, K., Bar, A., Yuille, A.~L., Darrell, T., Malik, J., and Efros, A.~A.
\newblock Sequential modeling enables scalable learning for large vision models.
\newblock In \emph{CVPR}, 2024.

\bibitem[Brockman et~al.(2016)Brockman, Cheung, Pettersson, Schneider, Schulman, Tang, and Zaremba]{brockman2016openaigym}
Brockman, G., Cheung, V., Pettersson, L., Schneider, J., Schulman, J., Tang, J., and Zaremba, W.
\newblock Openai gym, 2016.
\newblock URL \url{https://arxiv.org/abs/1606.01540}.

\bibitem[Chen et~al.(2025)Chen, Qin, Liu, Peng, Guan, Wang, Hu, Zhou, Gao, and Che]{chen2025towards}
Chen, Q., Qin, L., Liu, J., Peng, D., Guan, J., Wang, P., Hu, M., Zhou, Y., Gao, T., and Che, W.
\newblock Towards reasoning era: A survey of long chain-of-thought for reasoning large language models.
\newblock \emph{arXiv preprint arXiv:2503.09567}, 2025.

\bibitem[Chern et~al.(2024)Chern, Su, Ma, and Liu]{chern2024anole}
Chern, E., Su, J., Ma, Y., and Liu, P.
\newblock Anole: An open, autoregressive, native large multimodal models for interleaved image-text generation.
\newblock \emph{arXiv preprint arXiv:2407.06135}, 2024.

\bibitem[{Gemini Team}(2023)]{team2023gemini}
{Gemini Team}.
\newblock Gemini: a family of highly capable multimodal models.
\newblock \emph{arXiv preprint arXiv:2312.11805}, 2023.

\bibitem[Ho et~al.(2023)Ho, Schmid, and Yun]{ho2023large}
Ho, N., Schmid, L., and Yun, S.-Y.
\newblock Large language models are reasoning teachers.
\newblock In \emph{ACL}, 2023.

\bibitem[Huang et~al.(2025)Huang, Jia, Zhai, Cao, Ye, Zhao, Xu, Hu, and Lin]{huang2025vision}
Huang, W., Jia, B., Zhai, Z., Cao, S., Ye, Z., Zhao, F., Xu, Z., Hu, Y., and Lin, S.
\newblock Vision-r1: Incentivizing reasoning capability in multimodal large language models.
\newblock \emph{arXiv preprint arXiv:2503.06749}, 2025.

\bibitem[Hurst et~al.(2024)Hurst, Lerer, Goucher, Perelman, Ramesh, Clark, Ostrow, Welihinda, Hayes, Radford, et~al.]{hurst2024gpt}
Hurst, A., Lerer, A., Goucher, A.~P., Perelman, A., Ramesh, A., Clark, A., Ostrow, A., Welihinda, A., Hayes, A., Radford, A., et~al.
\newblock Gpt-4o system card.
\newblock \emph{arXiv preprint arXiv:2410.21276}, 2024.

\bibitem[Jin et~al.(2026)Jin, Zhou, Yang, Zhang, Liu, Zhu, and Deng]{jin2026latentum}
Jin, J., Zhou, Z., Yang, X., Zhang, H., Liu, P., Zhu, J., and Deng, Z.
\newblock {LatentUM}: Unleashing the potential of interleaved cross-modal reasoning via a latent-space unified model.
\newblock \emph{arXiv preprint arXiv:2604.02097}, 2026.

\bibitem[Kirkpatrick et~al.(2017)]{kirkpatrick2017overcoming}
Kirkpatrick, J. et~al.
\newblock Overcoming catastrophic forgetting in neural networks.
\newblock \emph{Proceedings of the National Academy of Sciences}, 114\penalty0 (13):\penalty0 3521--3526, 2017.

\bibitem[Kojima et~al.(2022)Kojima, Gu, Reid, Matsuo, and Iwasawa]{kojima2022large}
Kojima, T., Gu, S.~S., Reid, M., Matsuo, Y., and Iwasawa, Y.
\newblock Large language models are zero-shot reasoners.
\newblock In \emph{NeurIPS}, 2022.

\bibitem[Li et~al.(2025)Li, Wu, Zhang, Xia, Mao, Dong, Vuli{\'c}, and Wei]{li2025imagine}
Li, C., Wu, W., Zhang, H., Xia, Y., Mao, S., Dong, L., Vuli{\'c}, I., and Wei, F.
\newblock Imagine while reasoning in space: Multimodal visualization-of-thought.
\newblock \emph{arXiv preprint arXiv:2501.07542}, 2025.

\bibitem[Lu et~al.(2022)Lu, Mishra, Xia, Qiu, Chang, Zhu, Tafjord, Clark, and Kalyan]{lu2022learn}
Lu, P., Mishra, S., Xia, T., Qiu, L., Chang, K.-W., Zhu, S.-C., Tafjord, O., Clark, P., and Kalyan, A.
\newblock Learn to explain: Multimodal reasoning via thought chains for science question answering.
\newblock In \emph{NeurIPS}, 2022.

\bibitem[Mirzadeh et~al.(2025)Mirzadeh, Alizadeh, Shahrokhi, Tuzel, Bengio, and Farajtabar]{mirzadeh2025gsmsymbolic}
Mirzadeh, S.~I., Alizadeh, K., Shahrokhi, H., Tuzel, O., Bengio, S., and Farajtabar, M.
\newblock {GSM}-symbolic: Understanding the limitations of mathematical reasoning in large language models.
\newblock In \emph{ICLR}, 2025.

\bibitem[{Qwen Team}(2026)]{qwen35}
{Qwen Team}.
\newblock {Qwen3.5}: Towards native multimodal agents.
\newblock \url{https://qwen.ai/blog?id=qwen3.5}, February 2026.
\newblock Accessed: 2026-07-03.

\bibitem[Shojaee et~al.(2025)Shojaee, Mirzadeh, Alizadeh, Horton, Bengio, and Farajtabar]{shojaee2025illusion}
Shojaee, P., Mirzadeh, I., Alizadeh, K., Horton, M., Bengio, S., and Farajtabar, M.
\newblock The illusion of thinking: Understanding the strengths and limitations of reasoning models via the lens of problem complexity.
\newblock In \emph{NeurIPS}, 2025.

\bibitem[Stechly et~al.(2024)Stechly, Valmeekam, and Kambhampati]{stechly2024chain}
Stechly, K., Valmeekam, K., and Kambhampati, S.
\newblock Chain of thoughtlessness? an analysis of cot in planning.
\newblock In \emph{NeurIPS}, 2024.

\bibitem[von Werra et~al.(2020)von Werra, Belkada, Tunstall, Beeching, Thrush, Lambert, Huang, Rasul, and Gallouédec]{vonwerra2020trl}
von Werra, L., Belkada, Y., Tunstall, L., Beeching, E., Thrush, T., Lambert, N., Huang, S., Rasul, K., and Gallouédec, Q.
\newblock {TRL: Transformers Reinforcement Learning}, 2020.
\newblock URL \url{https://github.com/huggingface/trl}.

\bibitem[Wang et~al.(2025)Wang, Chang, and Wu]{wang2025theoretical}
Wang, Y., Chang, F.-C., and Wu, P.-Y.
\newblock A theoretical framework for ood robustness in transformers using gevrey classes.
\newblock \emph{arXiv preprint arXiv:2504.12991}, 2025.

\bibitem[Wei et~al.(2022)Wei, Wang, Schuurmans, Bosma, Xia, Chi, Le, Zhou, et~al.]{wei2022chain}
Wei, J., Wang, X., Schuurmans, D., Bosma, M., Xia, F., Chi, E., Le, Q.~V., Zhou, D., et~al.
\newblock Chain-of-thought prompting elicits reasoning in large language models.
\newblock \emph{NeurIPS}, 2022.

\bibitem[Wu et~al.(2024)Wu, Zhao, Saxon, Bui, Wang, Zhang, and Chang]{wu2024vsp}
Wu, Q., Zhao, H., Saxon, M., Bui, T., Wang, W.~Y., Zhang, Y., and Chang, S.
\newblock Vsp: Assessing the dual challenges of perception and reasoning in spatial planning tasks for vlms.
\newblock \emph{arXiv preprint arXiv:2407.01863}, 2024.

\bibitem[Xu et~al.(2025)Xu, Li, Zhou, Wan, Zhang, Korhonen, and Vuli{\'c}]{xu2025visual}
Xu, Y., Li, C., Zhou, H., Wan, X., Zhang, C., Korhonen, A., and Vuli{\'c}, I.
\newblock Visual planning: Let's think only with images.
\newblock \emph{arXiv preprint arXiv:2505.11409}, 2025.

\bibitem[Yang et~al.(2025)Yang, Yu, Chen, Shen, and Gan]{yang2025machine}
Yang, Z., Yu, X., Chen, D., Shen, M., and Gan, C.
\newblock Machine mental imagery: Empower multimodal reasoning with latent visual tokens.
\newblock \emph{arXiv preprint arXiv:2506.17218}, 2025.

\bibitem[Yue et~al.(2024)Yue, Ni, Zhang, Zheng, Liu, Zhang, Stevens, Jiang, Ren, Sun, et~al.]{yue2024mmmu}
Yue, X., Ni, Y., Zhang, K., Zheng, T., Liu, R., Zhang, G., Stevens, S., Jiang, D., Ren, W., Sun, Y., et~al.
\newblock Mmmu: A massive multi-discipline multimodal understanding and reasoning benchmark for expert agi.
\newblock In \emph{CVPR}, 2024.

\bibitem[Zhao et~al.(2025)Zhao, Tan, Ma, Li, Jiang, Wang, Yang, and Liu]{zhao2025chain}
Zhao, C., Tan, Z., Ma, P., Li, D., Jiang, B., Wang, Y., Yang, Y., and Liu, H.
\newblock Is chain-of-thought reasoning of llms a mirage? a data distribution lens.
\newblock \emph{arXiv preprint arXiv:2508.01191}, 2025.

\bibitem[Zhou et~al.(2025)Zhou, Tu, Wang, Wang, Muennighoff, Nie, Choi, Zou, Deng, Yan, et~al.]{zhou2025visualizing}
Zhou, Y., Tu, H., Wang, Z., Wang, Z., Muennighoff, N., Nie, F., Choi, Y., Zou, J., Deng, C., Yan, S., et~al.
\newblock When visualizing is the first step to reasoning: Mira, a benchmark for visual chain-of-thought.
\newblock \emph{arXiv preprint arXiv:2511.02779}, 2025.

\end{thebibliography}
\end{document}